\documentclass[10pt,twocolumn,letterpaper]{article}

\usepackage[pagenumbers]{cvpr} 

\usepackage{graphicx}
\usepackage{amsmath}
\usepackage{amssymb}
\usepackage{booktabs}

\usepackage{epsfig}
\usepackage{graphicx}

\usepackage{threeparttable}
\usepackage{multirow}
\usepackage{color}

\usepackage{enumerate}
\usepackage{stfloats}%
\usepackage{cite}%
\usepackage{caption}%
\usepackage{animate}%
\usepackage{bbm}
\usepackage{bm}
\usepackage{algorithm}
\usepackage{algorithmic}
\usepackage{bbding}

\usepackage[title]{appendix}

\newlength{\itemwidth}

\usepackage[pagebackref,breaklinks,colorlinks]{hyperref}

\usepackage[capitalize]{cleveref}
\crefname{section}{Sec.}{Secs.}
\Crefname{section}{Section}{Sections}
\Crefname{table}{Table}{Tables}
\crefname{table}{Tab.}{Tabs.}

\def\eg{\emph{e.g}\onedot} 

\def\ie{\emph{i.e}\onedot} 

\def\cf{\emph{cf}\onedot} 

\def\etc{\emph{etc}\onedot}

\def\aka{\emph{a.k.a}\onedot}
\def\etal{\emph{et al}\onedot}

\def\cvprPaperID{2141} 
\def\confName{CVPR}
\def\confYear{2022}

\begin{document}

\title{Context-Aware Video Reconstruction for Rolling Shutter Cameras}

\author{Bin Fan ~~~~~~~ Yuchao Dai\thanks{Y. Dai is the corresponding author (daiyuchao@gmail.com).} ~~~~~~~ Zhiyuan Zhang ~~~~~~~ Qi Liu ~~~~~~~ Mingyi He\\
	School of Electronics and Information, Northwestern Polytechnical University, Xi'an, China\\
}
\maketitle

\begin{abstract}
With the ubiquity of rolling shutter (RS) cameras, it is becoming increasingly attractive to recover the latent global shutter (GS) video from two consecutive RS frames, which also places a higher demand on realism. Existing solutions, using deep neural networks or optimization, achieve promising performance. However, these methods generate intermediate GS frames through image warping based on the RS model, which inevitably result in black holes and noticeable motion artifacts. In this paper, we alleviate these issues by proposing a context-aware GS video reconstruction architecture. It facilitates the advantages such as occlusion reasoning, motion compensation, and temporal abstraction. Specifically, we first estimate the bilateral motion field so that the pixels of the two RS frames are warped to a common GS frame accordingly. Then, a refinement scheme is proposed to guide the GS frame synthesis along with bilateral occlusion masks to produce high-fidelity GS video frames at arbitrary times. Furthermore, we derive an approximated bilateral motion field model, which can serve as an alternative to provide a simple but effective GS frame initialization for related tasks. Experiments on synthetic and real data show that our approach achieves superior performance over state-of-the-art methods in terms of objective metrics and subjective visual quality. 
Code is available at \url{https://github.com/GitCVfb/CVR}.
\end{abstract}

\begin{figure}[!t]
	\centering
	\includegraphics[width=0.475\textwidth]{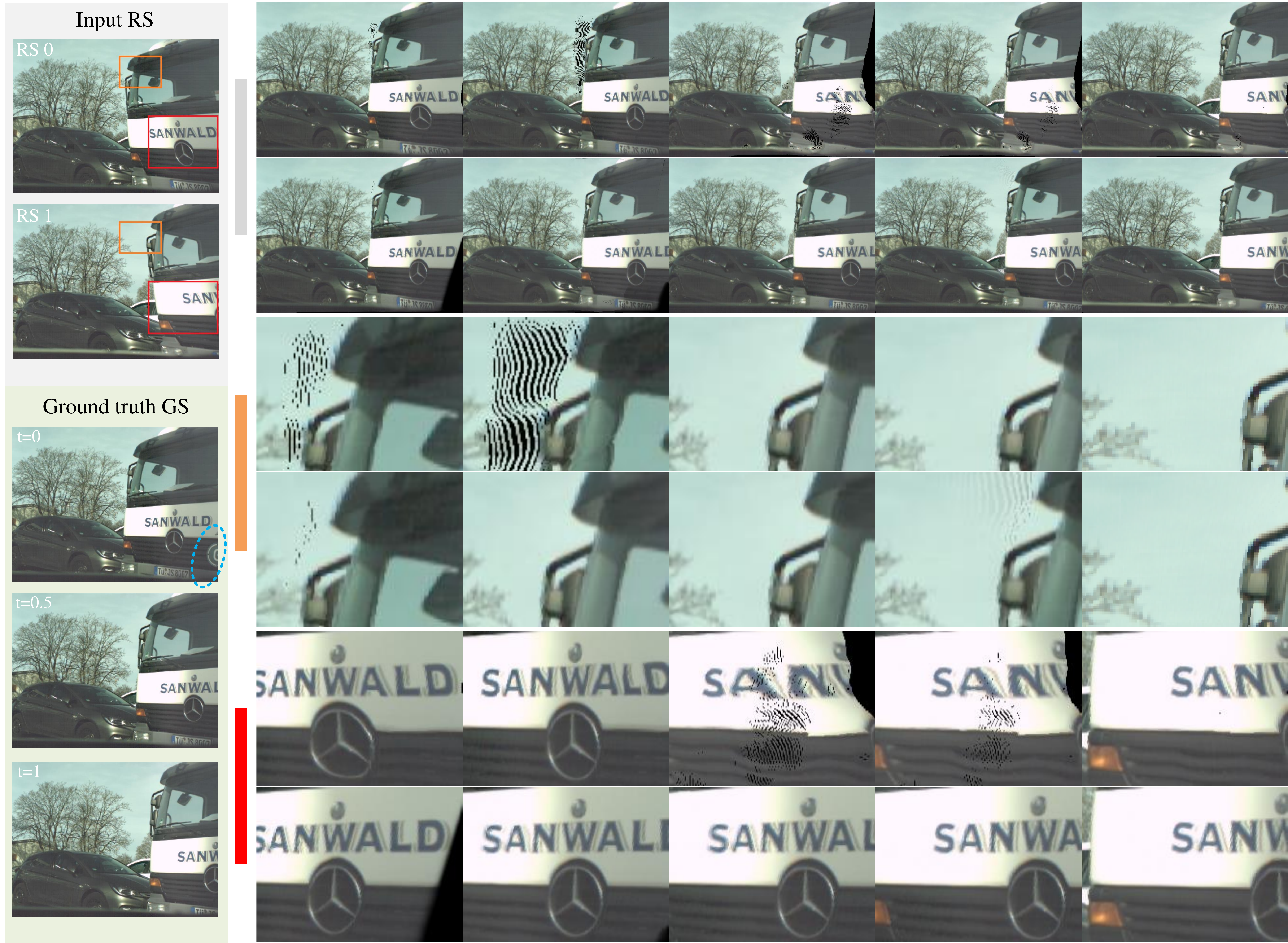} 
	\vspace{-5.8mm} 
	\caption{\textbf{GS video reconstruction example.} The left column shows two input consecutive RS images, and three ground-truth GS images at time 0, 0.5, and 1, respectively. Rows to the right show five GS frames (at times 0, 0.25, 0.5, 0.75, 1) extracted by \cite{fan2021rssr} (top) and our method (below), followed by two corresponding zoom-in regions. The orange box represents occluded black holes and the red box indicates motion artifacts specific to moving objects. Our method recovers higher fidelity GS images due to contextual aggregation and motion enhancement. Note that the black image edges by our method are because they are not available in both RS frames (\cf blue circle). Best viewed on Screen.\label{fig:teaser_img}}
	\vspace{-4.2mm}
\end{figure}

\section{Introduction} \label{sec:Introduction}
Many modern CMOS cameras equipped with rolling shutter (RS) dominate the consumer photography market due to their low cost and simplicity in design, and are also prevalent in the automotive sector and motion picture industry \cite{schubert2019rolling,hedborg2012rolling,vasu2018occlusion,zhuang2020homography}.
Within this acquisition mode, pixels on the rolling shutter CMOS sensor plane are exposed from top to bottom in a row-by-row fashion with a constant inter-row delay. 
This leads to undesirable visual distortions called the RS effect (\eg wobble, skew) in the presence of fast motion, which is a hindrance to scene understanding and a nuisance in photography.
With the increased demand for high quality and high framerate video of consumer-grade devices (\eg tablets, smartphones), video frame interpolation (VFI) has attracted increasing attention in the computer vision community.
Unfortunately, despite the remarkable success, the currently existing VFI methods \cite{bao2019depth,park2020bmbc,park2021asymmetric,jiang2018super,xu2019quadratic} implicitly assume that the camera employs a global shutter (GS) mechanism, \ie all pixels are exposed simultaneously. They are therefore unable to produce satisfying in-between frames with rolling shutter video acquired by \eg these devices in dynamic scenes or fast camera movements, resulting in RS artifacts remaining \cite{fan2021rssr}.

To address this problem, many RS correction methods \cite{im2018accurate,wu2021simultaneous,liu2020deep,zhuang2019learning,rengarajan2017unrolling,fan2021rsdpsnet} have been actively studied to eliminate the RS effect. 
In analogy to VFI generating non-existent intermediate GS frames from two consecutive GS frames, recovering the latent intermediate GS frames from two consecutive RS frames, \eg \cite{liu2020deep,fan2021sunet,zhuang2017rolling,zhuang2020homography}, serves as a tractable goal that overcomes the limited acquisition framerate and RS artifacts of commercial RS cameras.
This is significantly challenging because the output GS frames must follow coherence both temporally and spatially.
To this end, traditional methods \cite{zhuang2017rolling,zhuang2020homography} are often based on the assumption of constant velocity or constant acceleration camera motion, which struggle to accurately reflect the real camera motion and scene geometry, resulting in the persistence of ghosting and unsmooth artifacts \cite{fan2021rssr,liu2020deep}.
Recent deep learning-based solutions have achieved impressive performance, but they typically can only recover one GS image corresponding to a particular scanline, such as the first \cite{fan2021sunet} or central \cite{liu2020deep,zhong2021rscd} scanline, limiting their potentials for view transitions from RS to multiple-GS.

In this paper, we tackle the task of reviving and reliving all latent views of a scene as beheld by a virtual GS camera in the imaging interval of two consecutive RS frames.
Therefore, we must jointly deal with VFI and RS correction tasks, \ie interpolating smooth and trustworthy distortion-free video sequences.
It is worth mentioning that the most relevant work to our task is \cite{fan2021rssr}, which is dedicated to the geometry-aware RS inversion by warping each RS frame to its corresponding virtual GS counterpart.
Nevertheless, as illustrated in Fig.~\ref{fig:teaser_img}, the GS images recovered by \cite{fan2021rssr} still suffers from two limitations:
\begin{itemize}
	\vspace{-1.2mm}
	\item \textbf{Masses of black holes} (\cf orange box). This is a common issue for warping-based methods (\eg \cite{zhuang2017rolling,zhuang2020homography,zhuang2019learning,fan2021rssr,rengarajan2016bows}) due to the occlusion between the RS and GS images, leading to the possibility of permanent loss of some valuable image contents. To maintain visual consistency, a cropping operation is used to discard the resulting holes, but may degrade the visual experience.
	\vspace{-6.0mm}
	\item \textbf{Noticeable object-specific motion artifacts} (\cf red box). When recording dynamic scenes, the moving object violates the constant velocity motion assumption of RS cameras used in \cite{fan2021rssr}, resulting in its inability to accurately capture motion boundaries specific to moving objects. Thus severe motion artifacts are generated.
	\vspace{-5.5mm}
\end{itemize}

In contrast, we investigate contextual aggregation and motion enhancement based on the bilateral motion field (BMF) to alleviate these issues, which aims to synthesize crisp and pleasing GS video frames by occlusion reasoning and temporal abstraction.
Specifically, we propose \emph{CVR} (\textbf{C}ontext-aware \textbf{V}ideo \textbf{R}econstruction architecture), which consists of two stages to recover a faithful and coherent GS video sequence from two input consecutive RS images. 
In the first initialization stage, we adopt a motion interpretation module to estimate the initial bilateral motion field, which warps the two RS frames to a common GS version. 
We design two schemes to achieve this goal.
One is based on \cite{fan2021rssr} which requires a pre-trained encoder-decoder network; the other is our proposed approximation of \cite{fan2021rssr}, without resorting to a deep network.
Also, we show that this simple approximation is able to provide a feasible solution for the initial prediction.
Afterward, a second refinement stage is introduced to handle black holes and ambiguous misalignments caused by occlusions and object-specific motion patterns.
As a result of exploiting bilateral motion residuals and occlusion masks, it can guide the subsequent GS frame synthesis to reason about complex motion profiles and occlusions.
Furthermore, inspired by \cite{fan2021sunet}, we propose a contextual consistency constraint to effectively aggregate the contextual information, such that the unsmooth areas can be enhanced in an adaptive manner.
Extensive experimental results demonstrate that our method surpasses the state-of-the-art (SOTA) methods by a large margin in removing RS artifacts. Meanwhile, our method is capable of generating high-fidelity GS videos.

The main contributions of this paper are three-fold:
\begin{enumerate}[1)]
	\vspace{-2.0mm}
	\item We propose a simple yet effective bilateral motion field approximation model, which serves as a reliable initialization for GS frame refinement.
	\vspace{-2.0mm}
	\item We develop a stable and efficient context-aware GS video reconstruction framework, which can reason about complex occlusions, motion patterns specific to objects, and temporal abstractions.
	\vspace{-2.0mm}
	\item Experiments show that our method achieves SOTA results while maintaining an efficient network design.
\end{enumerate}

\section{Related Work} \label{sec:RelatedWork}
\vspace{0.5mm}
\noindent\textbf{Video frame interpolation} has been widely studied in recent years, which can be categorized into phase-based \cite{meyer2015phase,meyer2018phasenet}, kernel-based \cite{choi2020channel,liu2017video,niklaus2017video}, and flow-based \cite{bao2019depth,park2020bmbc,jiang2018super,siyao2021deep} methods.
With the latest advances in optical flow estimation \cite{dosovitskiy2015flownet,sun2018pwc,teed2020raft}, the flow-based VFI methods have been actively studied to explicitly exploit motion information.
After the seminal work \cite{jiang2018super}, subsequent improvements are dedicated to better intermediate flow estimation on one hand, such as quadratic \cite{xu2019quadratic}, rectified quadratic \cite{liu2020enhanced}, and cubic \cite{chi2020all} flow interpolations. 
Moreover, Bao \etal \cite{bao2019depth} strengthened the initial flow field using the predicted depth map via a depth-aware flow projection layer.
Park \etal estimated a symmetric bilateral motion \cite{park2020bmbc} to produce the intermediate flows directly, and they have recently developed an asymmetric bilateral motion model \cite{park2021asymmetric} to refine the intermediate frame.
On the other hand, better refinement and fusion of details were focused on, including contextual warping \cite{niklaus2018context,bao2019depth,moing2021ccvs}, occlusion inference \cite{xue2019video,bao2019memc}, cycle constraints \cite{reda2019unsupervised,liu2019deep} for more accurate frame synthesis, and softmax splatting \cite{niklaus2020softmax} for more efficient forward warping, \etc

All of these VFI approaches work with a common assumption that the camera employs a GS mechanism. Hence, they are incapable of correctly synthesizing the in-between frames in the case of RS images. In this paper, we integrate an effective motion interpretation module to boost the reliable estimation of the initial flow field, yielding high-quality results without aliasing.

\vspace{0.5mm}
\noindent\textbf{Rolling shutter correction} 
advocates the mitigation or elimination of RS distortion, \ie recovering the latent GS image, from a single frame \cite{rengarajan2017unrolling,zhuang2019learning,rengarajan2016bows,lao2018robust} or multiple frames \cite{liu2020deep,zhuang2017rolling,grundmann2012calibration,saurer2013rolling,albl2020two,wang2020relative}.
Dai \etal \cite{dai2016rolling} derived the discrete two-view RS epipolar geometry. Zhuang \etal \cite{zhuang2017rolling} proposed a differential RS epipolar constraint to undistort two consecutive RS images, whose stereo version was further explored in \cite{fan2021rsstereo}.
Likewise, Lao \etal \cite{lao2020rolling} developed a discrete RS homography model to perform the plane-based RS correction. Zhuang and Tran \cite{zhuang2020homography} presented a differential RS homography to account for the scanline-varying poses of RS cameras.
In addition, some additional assumptions are often taken into account, such as pure rotational motion \cite{rengarajan2016bows,lao2018robust,ringaby2012efficient,forssen2010rectifying}, Ackermann motion \cite{Purkait2018Minimal}, and Manhattan world \cite{purkait2017rolling}.
With the rise of deep learning, many appealing RS correction results have been achieved.
For two input consecutive RS frames, Liu \etal \cite{liu2020deep} put forward a deep shutter unrolling network to estimate the latent GS frame, and Fan \etal \cite{fan2021sunet} proposed a symmetric network architecture to efficiently aggregate the contextual cues.
Zhong \etal \cite{zhong2021rscd} used a deformable attention module to jointly solve the RS correction and deblurring problem.
Unfortunately, they can only hallucinate one GS image at a specific moment, \eg corresponding to the first \cite{fan2021sunet} or central \cite{liu2020deep,zhong2021rscd} scanline time, and thus fall short of reconstructing a smooth and coherent GS video.

Very recently, Fan and Dai \cite{fan2021rssr} developed the first rolling shutter temporal super-resolution network to extract a high framerate GS video from two consecutive RS images. 
It warps each RS frame to a latent GS frame corresponding to any of its scanlines through geometry-aware propagation. As a result, undesirable holes (\eg black edges) appear due to the occlusion between the RS and GS images.
Furthermore, it leverages a constant velocity motion assumption, which does not accurately capture the motion boundaries and produces artifacts around the moving objects.
Two examples are shown in Figs.~\ref{fig:teaser_img} and \ref{fig:multiple_frames}.
In contrast, we propose a GS frame synthesis module, which is composed of contextual aggregation and motion enhancement layers, to reason about complex occlusions and motion patterns specific to moving objects, resulting in a significantly improved performance of GS video reconstruction.

\begin{figure}[!t]
	\centering
	\includegraphics[width=0.47\textwidth]{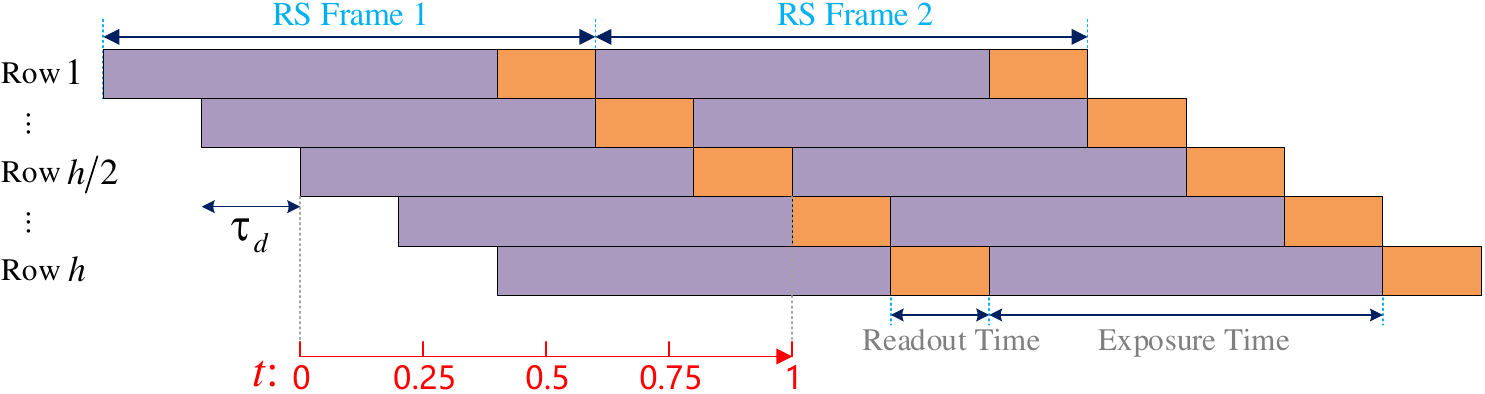} \vspace{-2.5mm}
	\caption{RS mechanism over two consecutive frames. We aim at recovering the latent GS images at time $t\in[0,1]$. \label{fig:RS_mechanism}}
	\vspace{-4.7mm}
\end{figure}

\section{RS-aware Frame Warping} \label{sec:RSWarping}
\noindent\textbf{RS image formation model.}
When an RS camera is in motion during the image acquisition, all its scanlines are exposed sequentially at different timestamps. Hence each scanline possesses a different local frame, as illustrated in Fig.~\ref{fig:RS_mechanism}.
Without loss of generality, we assume that all pixels in the same row are exposed instantaneously at the same time. The number of rows in the image is $h$, and the constant inter-row delay time is $\tau_d$.
Therefore, the RS image formation model can be obtained as follows:
\begin{equation}\label{eq:1}
\left\lfloor\mathbf{I}^{r}(\mathbf{x})\right\rfloor_{s}=\left\lfloor\mathbf{I}^{g}_{s}(\mathbf{x})\right\rfloor_{s},
\end{equation}
where $\mathbf{I}^{g}_{s}$ is virtual GS images captured at time $\tau_d(s-h/2)$, and $\lfloor\cdot\rfloor_s$ denotes the extraction of pixel $\mathbf{x}$ in scanline $s$.

\vspace{0.5mm}
\noindent\textbf{RS effect removal by forward warping.}
Since an RS image can be viewed as the result of successive row-by-row combinations of virtual GS image sequences within the imaging duration, one can invert the above RS imaging mechanism to remove RS distortions by
\begin{equation}\label{eq:2}
\mathbf{I}^{r}(\mathbf{x})=\mathbf{I}^{g}_{s}(\mathbf{x}+\mathbf{u}_{r \to s}),
\end{equation}
where $\mathbf{u}_{r \to s}$ is the displacement vector of pixel $\mathbf{x}$ from the RS image $\mathbf{I}^{r}$ to the virtual GS image $\mathbf{I}^{g}_{s}$.
Stacking $\mathbf{u}_{r \to s}$ of all pixels yields a pixel-wise motion field, \aka \emph{undistortion flow} ${\mathbf U}_{r \to s}$, which can be used to RS-aware forward warping analogous to \cite{liu2020deep,fan2021rssr,fan2021sunet,zhong2021rscd}.
However, when multiple pixels are mapped to the same location, forward warping is prone to suffer from conflicts, inevitably leading to overlapped pixels and holes. Softmax splatting \cite{niklaus2020softmax} alleviates these problems by adaptively combining overlapping pixel information.
Thus, the target GS frame corresponding to scanline $s$ can be generated by 
\begin{equation}\label{eq:3}
\hat{\mathbf{I}}^{g}_{s} = \mathcal{W}_F(\mathbf{I}^{r}, {\mathbf U}_{r \to s}),
\end{equation}
where $\mathcal{W}_F$ represents the forward warping operator. 
We use softmax splatting in our implementation.

\vspace{0.5mm}
\noindent\textbf{Problem setup.}
As depicted in Fig.~\ref{fig:RS_mechanism}, time $t$ and scanline $s$ correspond to each other. 
For compactness, in the following we will discard the symbol $s$ and use the subscript $t$ to denote the GS image ${\mathbf{I}}^{g}_{t}$ corresponding to time $t$.
Following \cite{liu2020deep,zhuang2020homography,fan2021rsstereo}, we further assume that the readout time ratio \cite{zhuang2017rolling}, \ie the ratio between the total scanline readout time (\ie $h\tau_d$) and the inter-frame delay time, is equal to one. 
That is to say, the idle time between two adjacent RS frames is ignored in a short period of imaging time (\eg $<50$ ms).
This is proved to be effective to account for the scanline-varying camera poses, avoiding non-trivial readout calibration \cite{meingast2005geometric}.
Moreover, this also ensures temporally tractable frame interpolation for RS images.
See Appendix~\ref{sec:instructions_setup} for further instructions.
Consequently, the central scanlines of the two consecutive RS images are recorded at time instances 0 and 1, respectively.

Given two RS frames ${\mathbf I}^r_0$ and ${\mathbf I}^r_1$ at adjacent times 0 and 1, we aim to synthesize an intermediate GS frame $\hat{\mathbf I}_t^g$, $t \in [0,1]$.
This time interval is chosen because, as observed in \cite{fan2021sunet}, many details of the recovered GS images corresponding to time $t\in[-0.5,0) \cup (1,1.5]$ are more likely to be missing due to too much deviation from the temporal consistency.

\subsection{Bilateral Motion Field Initialization}
\begin{figure}[!t]
	\centering
	\includegraphics[width=0.47\textwidth]{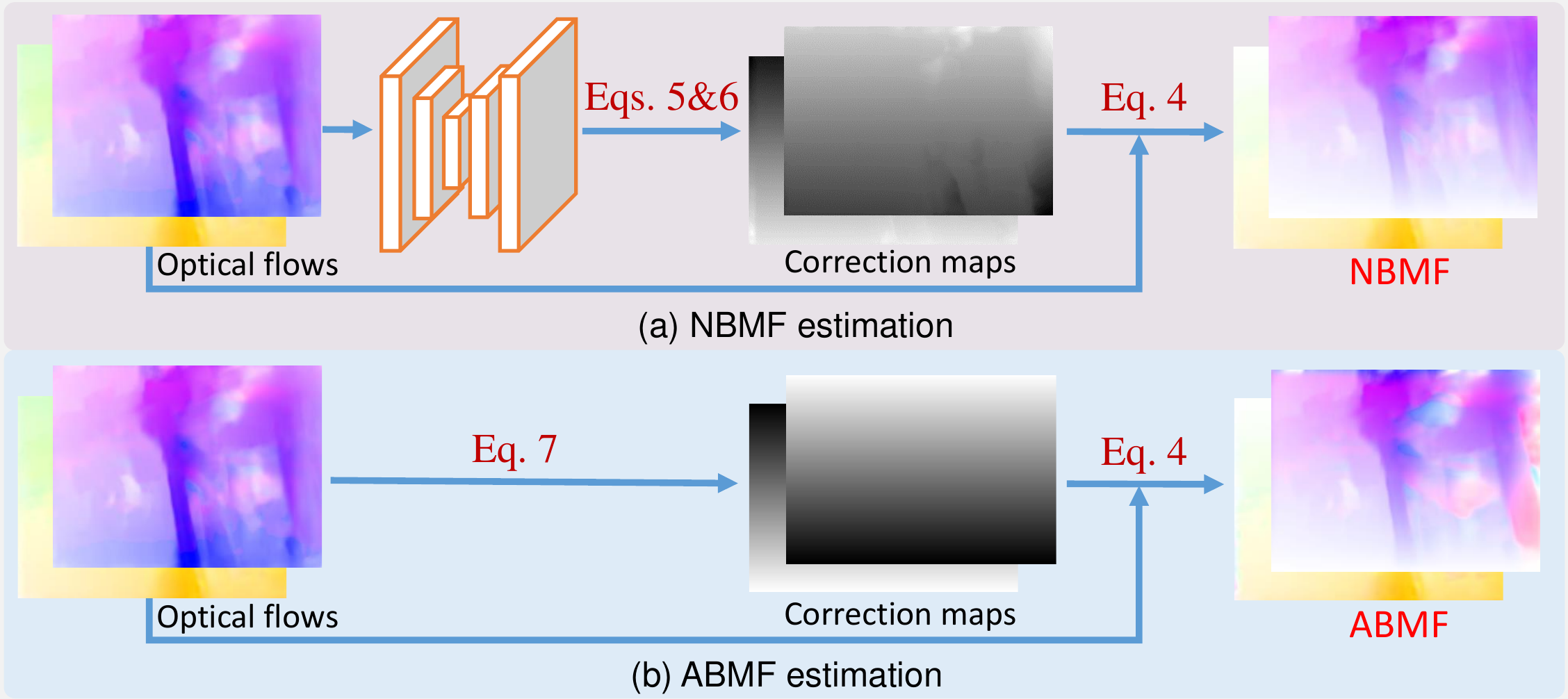} \vspace{-3.0mm}
	\caption{Illustration of the initial BMF estimation, including (a) NBMF and its approximation (b) ABMF. \label{fig:nbmf_abmf}}
	\vspace{-5.4mm}
\end{figure}

\vspace{-0.1mm}
\noindent\textbf{Network-based bilateral motion field (NBMF).}
To deliver each RS pixel $\mathbf{x}$ exposed at time $\tau$ (\ie $\tau_0 \in [-0.5,0.5]$ or $\tau_1 \in [0.5,1.5]$, with subscripts indicating the image index) to the GS canvas corresponding to the camera pose at time $t \in [0,1]$, we need to estimate the motion field ${\mathbf U}_{0 \to t}$ or ${\mathbf U}_{1 \to t}$ (\cf Eq.~\eqref{eq:3}) to constrain each pixel's displacement.
Note that the subscripts $0 \to t$ and $1 \to t$ indicate the RS-aware forward warping from RS images ${\mathbf I}^r_0$ and ${\mathbf I}^r_1$ to $\hat{\mathbf{I}}^{g}_{t}$, respectively.
According to \cite{fan2021rssr}, we extend to the time dimension to model the BMF ${\mathbf U}_{0 \to t}$ and ${\mathbf U}_{1 \to t}$ by a scaling operation on the corresponding optical flow fields ${\mathbf F}_{0 \to 1}$ and ${\mathbf F}_{1 \to 0}$ between two consecutive RS frames, \ie 
\begin{equation}\label{eq:4}
\begin{aligned}
{\mathbf U}_{0 \to t}(\mathbf{x}) &= {\mathbf C}_{0 \to t}(\mathbf{x}) \cdot {\mathbf F}_{0 \to 1}(\mathbf{x}),\\
{\mathbf U}_{1 \to t}(\mathbf{x}) &= {\mathbf C}_{1 \to t}(\mathbf{x}) \cdot {\mathbf F}_{1 \to 0}(\mathbf{x}),
\end{aligned}
\end{equation}
where 
\begin{equation}\label{eq:5}
\begin{aligned}
{\mathbf C}_{0 \to t}(\mathbf{x}) &= \frac{(t-\tau_0) (h-\pi_v)}{h},\\
{\mathbf C}_{1 \to t}(\mathbf{x}) &= \frac{(\tau_1-t) (h+\pi'_v)}{h},
\end{aligned}
\end{equation}
represent the bilateral correction maps. $\pi_v$ and $\pi'_v$ encapsulate the underlying RS geometry \cite{fan2021rssr} to reveal the inter-RS-frame vertical optical flow, depending on the camera parameters, the camera motion, and the depth and position of pixel $\mathbf{x}$. Furthermore, the BMF corresponding to different time steps $t_1$ and $t_2$ can be directly interconverted by
\begin{equation}\label{eq:6}
{\mathbf U}_{i \to t_2}(\mathbf{x}) = \frac{t_2-\tau}{t_1-\tau} \cdot {\mathbf U}_{i \to t_1}(\mathbf{x}), \,\,\, i=0,1.
\end{equation}
Note that the motion field for RS removal has a significant time dependence (\aka scanline dependence \cite{fan2021rssr}).
To capture the correction map in Eq.~\eqref{eq:5}, a geometric optimization problem was posed in \cite{zhuang2017rolling,zhuang2020homography} based on the differential formulation \cite{ma2000linear,fan2021fast}.
Recently, as shown in Fig.~\ref{fig:nbmf_abmf}~(a), an encoder-decoder network was proposed in \cite{fan2021rssr} to essentially learn the underlying RS geometry, such that the BMF can be computed by Eq.~\eqref{eq:4} coupled with the estimated bidirectional optical flows, termed as NBMF. The arbitrary-time GS images are then generated by image warping based on explicit intra-frame propagation in Eq.~\eqref{eq:6}.
However, since the occlusion view is not available during warping, the resulting holes are visually unsatisfactory. Also, \cite{fan2021rssr} is not adaptive to dynamic objects due to the reliance on a constant velocity motion assumption of the RS camera.

\vspace{0.5mm}
\noindent\textbf{Approximated bilateral motion field (ABMF).}
We observe that $\pi_v$ and $\pi'_v$ in Eq.~\eqref{eq:5} characterize the latent inter-GS-frame vertical optical flow, which are usually much smaller than the number of image rows $h$ (\cf Appendix~\ref{sec:ABMF_analyses} for in-depth analysis). Hence, we propose an approximated constraint $h-\pi_v \approx h \approx h+\pi'_v$ to rewrite Eq.~\eqref{eq:5} as:
\begin{equation}\label{eq:7}
\begin{aligned}
{\mathbf C}_{0 \to t}(\mathbf{x}) &= t-\tau_0,\\
{\mathbf C}_{1 \to t}(\mathbf{x}) &= \tau_1-t,
\end{aligned}
\end{equation}
where the time dependence is retained while the parallax effects (\ie depth variation and camera motion) are neglected. 
That is, it is independent of the image content and can be pre-defined for a given image resolution. 
As depicted in Fig.~\ref{fig:nbmf_abmf}~(b), such approximation is able to reach the correction map and then the ABMF via Eq.~\eqref{eq:4} in a simple and straightforward manner instead of relying on specialized deep neural networks.
Note that the interconversion between varying ABMF satisfies Eq.~\eqref{eq:6} as well.
The experimental results in Sec.~\ref{sec:Comparison_sota} show that our ABMF, coupled with the contextual aggregation and motion enhancement, can serve as a strong and tractable baseline for GS frame synthesis. 

\begin{figure*}[!t]
	\centering
	\includegraphics[width=0.88\textwidth]{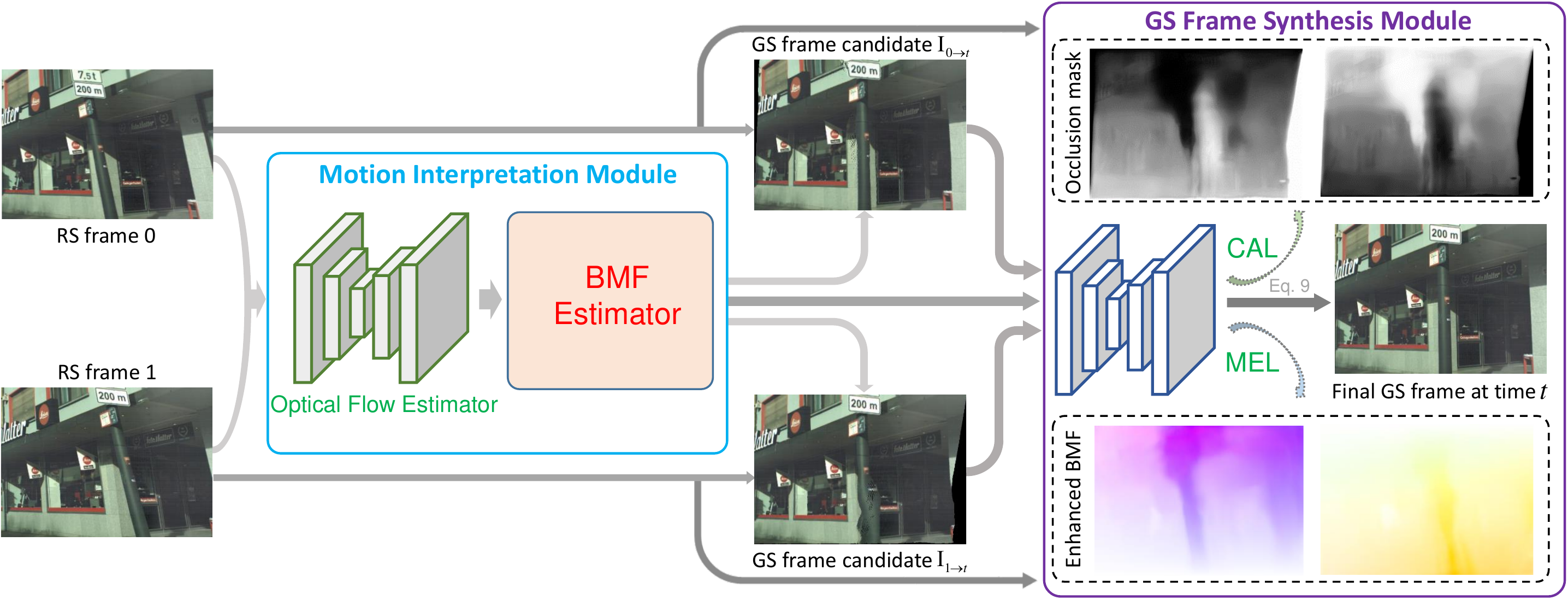} \vspace{-2.2mm}
	\caption{\textbf{Overall architecture.} It has two main processes. 
	First, two initial GS frame candidates are obtained by the motion interpretation module. The details of BMF estimator (\ie NBMF or ABMF) are elaborated in Fig.~\ref{fig:nbmf_abmf}. Then, a GS frame synthesis module is proposed to reason about complex occlusions, motion profiles, and temporal abstractions to generate the final high-fidelity GS image at time $t\in[0,1]$.
	\label{fig:architecture}}
	\vspace{-9.3mm}
\end{figure*}

\vspace{-1.5mm}
\section{Context-aware Video Reconstruction} \label{sec:CVR}
\vspace{-1.0mm}
We advocate recovering the intermediate global shutter image $\hat{\mathbf I}_t^g, t \in [0,1]$ from two input consecutive rolling shutter images ${\mathbf I}_0^r$ and ${\mathbf I}_1^r$. In this section, we will explain how to design a deep network to reason about time-aware motion profiles and occlusions, such that the photorealistic time-arbitrary GS image can be recovered faithfully.

\vspace{-1.0mm}
\subsection{Architecture Overview} \label{subsec:Architecture}
\vspace{-1.0mm}
As shown in Fig.~\ref{fig:architecture}, the proposed network consists of two modules, \ie an NBMF-based or ABMF-based motion interpretation module, and a context (\ie occlusions and partial dynamics) aware GS frame synthesis module.
Firstly, we estimate the bidirectional optical flow fields ${\mathbf F}_{0 \to 1}$ and ${\mathbf F}_{1 \to 0}$ between ${\mathbf I}_0^r$ and ${\mathbf I}_1^r$, followed by the BMF estimation ${\mathbf U}_{0 \to t}$ and ${\mathbf U}_{1 \to t}$ via Eq.~\eqref{eq:4}, which is based on NBMF (\ie Eq.~\eqref{eq:5}) or ABMF (\ie Eq.~\eqref{eq:7}), as illustrated in Fig.~\ref{fig:nbmf_abmf}. 
Then, the input RS frames are forward warped using the initial bilateral motions, resulting in two initial intermediate GS frame candidates at time $t$. 
Finally, the GS frame synthesis module takes the input RS frames, bidirectional optical flows, bilateral motion fields, and the initial intermediate GS frame candidates to synthesize the final GS reconstruction by aggregating the context information and compensating for the motion boundaries adaptively. 
Note that we empirically find that our ABMF-based CVR approach (called \emph{CVR*}) performs well despite its simplicity, while our NBMF-based CVR approach (called \emph{CVR}) can further improve the quality of the final GS images.

\vspace{0.5mm}
\textbf{Motion interpretation module $\mathcal{M}$}
is composed of two submodules: an optical flow estimator and a bilateral motion field estimator.
We first utilize the widely used PWC-Net \cite{sun2018pwc} as the optical flow estimator to predict the bidirectional optical flow.
To obtain an effective initial BMF, we follow \cite{fan2021rssr} and use a dedicated encoder-decoder U-Net architecture \cite{ronneberger2015unet,paliwal2020deep}, as shown in Fig.~\ref{fig:nbmf_abmf}~(a), to estimate NBMF for forward warping, which is termed as $\mathcal{M}_N$. Particularly, $\mathcal{M}_N$ needs to be pre-trained by using the ground-truth (GT) central-scanline GS images for supervision. 
Alternatively, we propose to exploit its approximate version as shown in Fig.~\ref{fig:nbmf_abmf}~(b), \ie an ABMF-based motion interpretation module $\mathcal{M}_A$, to yield a simpler and faster prediction of the initial BMF.
Finally, two initial intermediate GS frame candidates ${\mathbf I}_{0 \to t}^g$ and ${\mathbf I}_{1 \to t}^g$ can be generated by Eq.~\eqref{eq:3} based on the initial BMF estimations ${\mathbf U}_{0 \to t}$ and ${\mathbf U}_{1 \to t}$, respectively. 

\vspace{0.5mm}
\textbf{GS frame synthesis module}
$\mathcal{G}$ can be boiled down to two main layers: a motion enhancement layer (MEL) and a contextual aggregation layer (CAL). 
Note that some black holes and ambiguous misalignments may exist in the initial intermediate GS frame candidates due to heavy occlusions and partial moving objects, degrading the visual experience.
Therefore, we aim at alleviating artifacts at the boundaries of dynamic objects and filling the occluded holes.
Towards this goal, ${\mathbf I}_0^r$, ${\mathbf I}_1^r$, ${\mathbf F}_{0 \to 1}$, ${\mathbf F}_{1 \to 0}$, ${\mathbf U}_{0 \to t}$, ${\mathbf U}_{1 \to t}$, ${\mathbf I}_{0 \to t}^g$, and ${\mathbf I}_{1 \to t}^g$ are concatenated and fed into $\mathcal{G}$ to estimate the BMF residuals $\Delta{\mathbf U}_{0 \to t}$ and $\Delta{\mathbf U}_{1 \to t}$ and the bilateral occlusion masks ${\mathbf O}_{0 \to t}$ and ${\mathbf O}_{1 \to t}$. 
This time-aware occlusion mask is essential to guide GS frame synthesis to handle occlusions.
We employ an encoder-decoder U-Net network \cite{ronneberger2015unet,paliwal2020deep} as the backbone of $\mathcal{G}$, which has the same structure but different channels as the network in $\mathcal{M}_N$.
The network is fully convolutional with skip connections and leaky ReLu activation functions. Besides, we leverage a sigmoid activation function on the output channels corresponding to the bilateral occlusion mask to limit its value between 0 and 1.
Because $\mathcal{G}$ accepts cascades at different time instances, it can implicitly model the temporal abstraction to recover GS frames corresponding to arbitrary time step $t \in [0,1]$.

Specifically, the final enhanced BMF can be obtained as:
\begin{equation}\label{eq:8}
\begin{aligned}
\hat{\mathbf U}_{0 \to t} &= {\mathbf U}_{0 \to t} + \Delta{\mathbf U}_{0 \to t},\\
\hat{\mathbf U}_{1 \to t} &= {\mathbf U}_{1 \to t} + \Delta{\mathbf U}_{1 \to t},
\end{aligned}
\end{equation}
which can improve the quality of BMF by combining it with the proposed contextual consistency constraint, especially in motion boundaries and unsmooth regions.
Subsequently, we can produce two refined intermediate GS frame candidates $\hat{\mathbf I}_{0 \to t}^g$ and $\hat{\mathbf I}_{1 \to t}^g$ by RS-aware forward warping in Eq.~\eqref{eq:3}.
Further, we assume that the content of the target GS image corresponding to $t \in [0,1]$ can be recovered by at least one of the input RS images, which is promising as discussed in \cite{fan2021sunet}. We therefore impose the constraint that ${\mathbf O}_{1 \to t} = {\mathbf 1} - {\mathbf O}_{0 \to t}$. Intuitively, ${\mathbf O}_{0 \to t}(\mathbf{x}) = 0$ implies ${\mathbf O}_{1 \to t}(\mathbf{x}) = 1$, \ie target pixels can be faithfully rendered by fully trusting ${\mathbf I}_1^r$, and vice versa. Similar to \cite{paliwal2020deep,xu2019quadratic,jiang2018super}, we also take advantage of the temporal distances $1-t$ and $t$ for the input RS frames ${\mathbf I}_0^r$ and ${\mathbf I}_1^r$, such that the temporally-closer pixels can be assigned a higher confidence.
At last, the final intermediate GS frame $\hat{\mathbf I}_t^g$ can be synthesized by 
\begin{equation}\label{eq:9}
\hat{\mathbf I}_t^g = \frac{(1-t){\mathbf O}_{0 \to t}\hat{\mathbf I}_{0 \to t}^g + t{\mathbf O}_{1 \to t}\hat{\mathbf I}_{1 \to t}^g} {(1-t){\mathbf O}_{0 \to t} + t{\mathbf O}_{1 \to t}}.
\end{equation}

\begin{figure*}\centering
	\setlength{\tabcolsep}{0.015cm}
	\setlength{\itemwidth}{3.687cm}
	\hspace*{-\tabcolsep}\begin{tabular}{cccccc}
		\includegraphics[width=0.78\itemwidth]{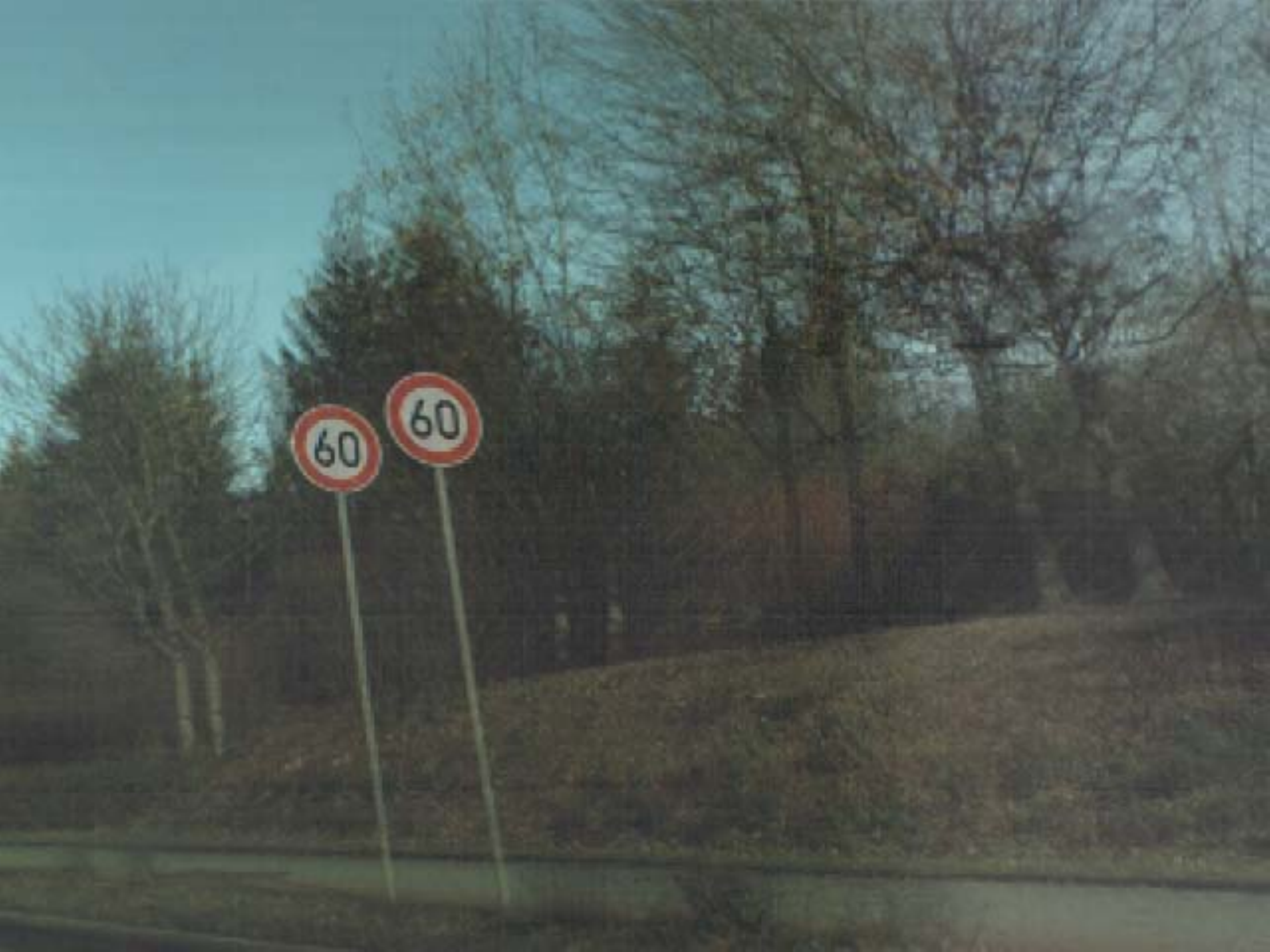}
		&
		\includegraphics[width=0.78\itemwidth]{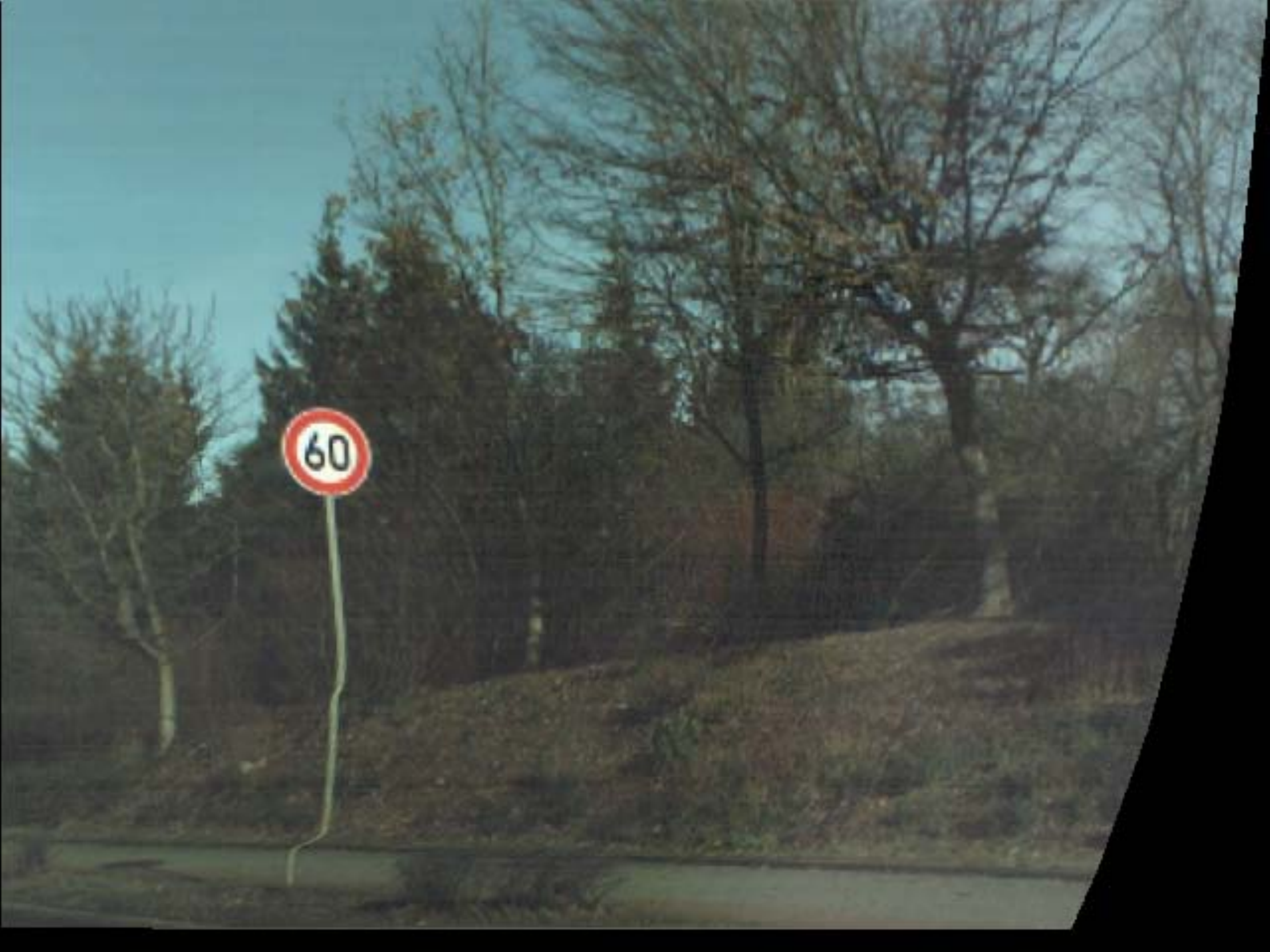}
		&
		\includegraphics[width=0.78\itemwidth]{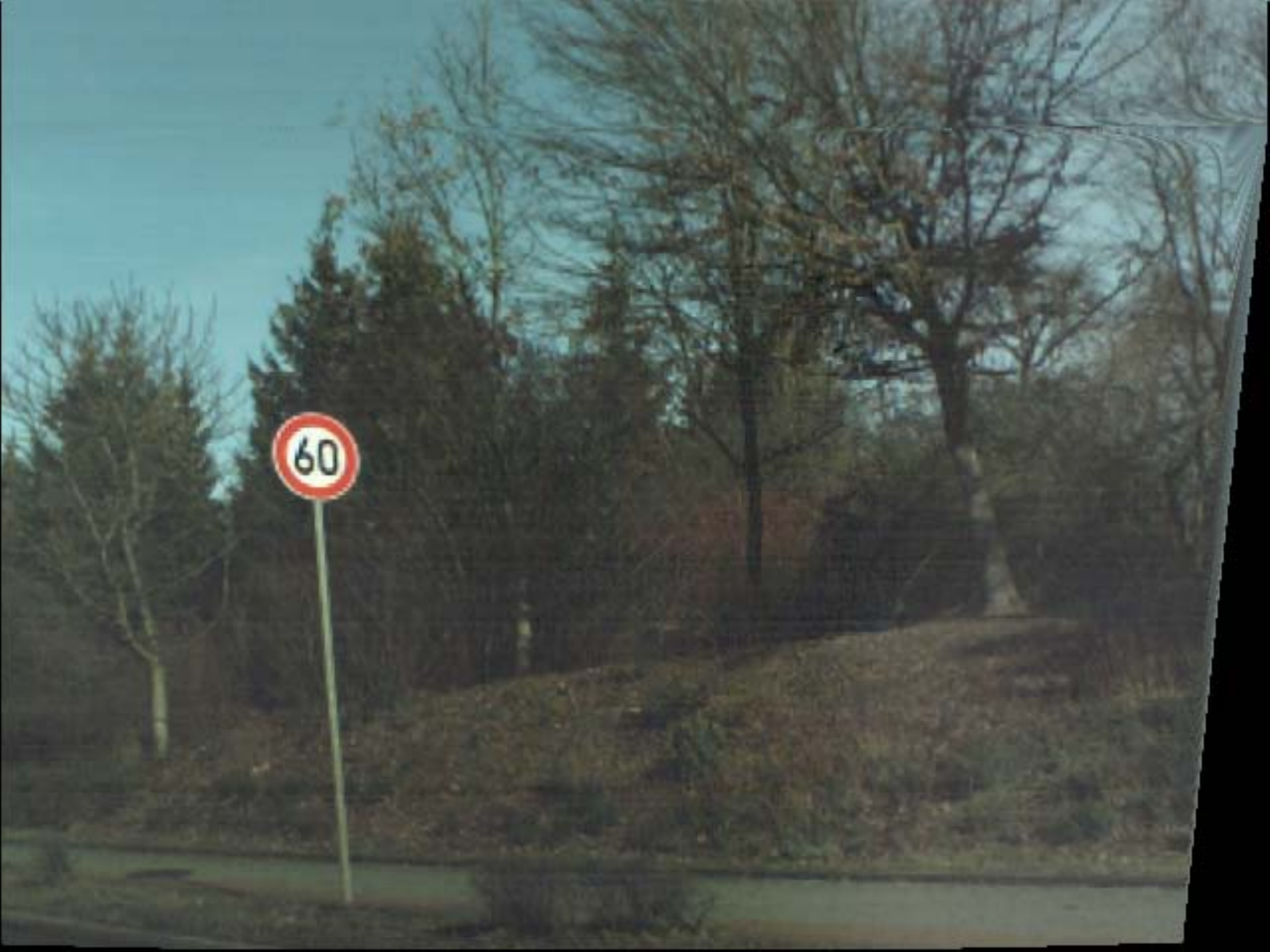}
		&
		\includegraphics[width=0.78\itemwidth]{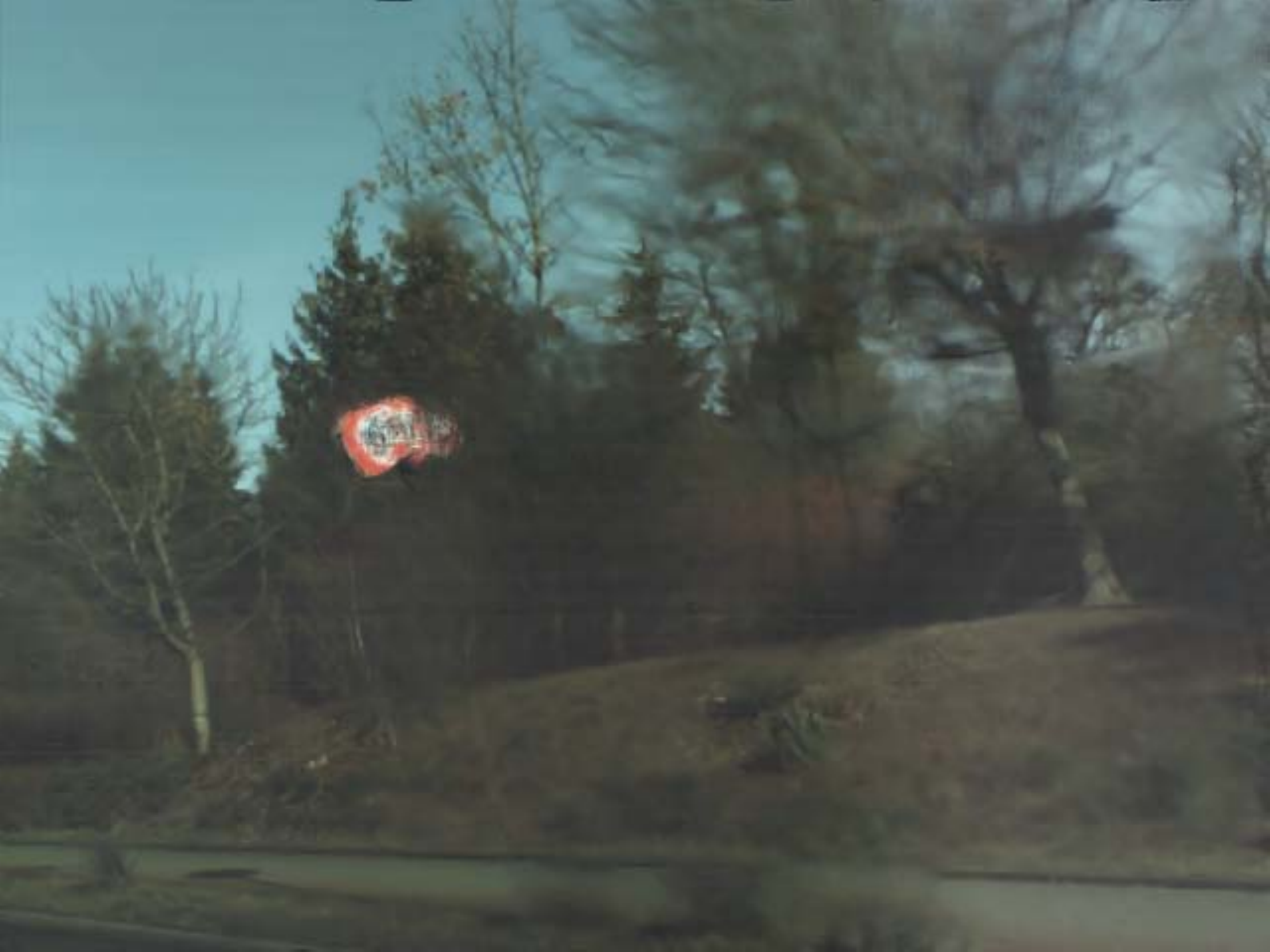}
		&
		\includegraphics[width=0.78\itemwidth]{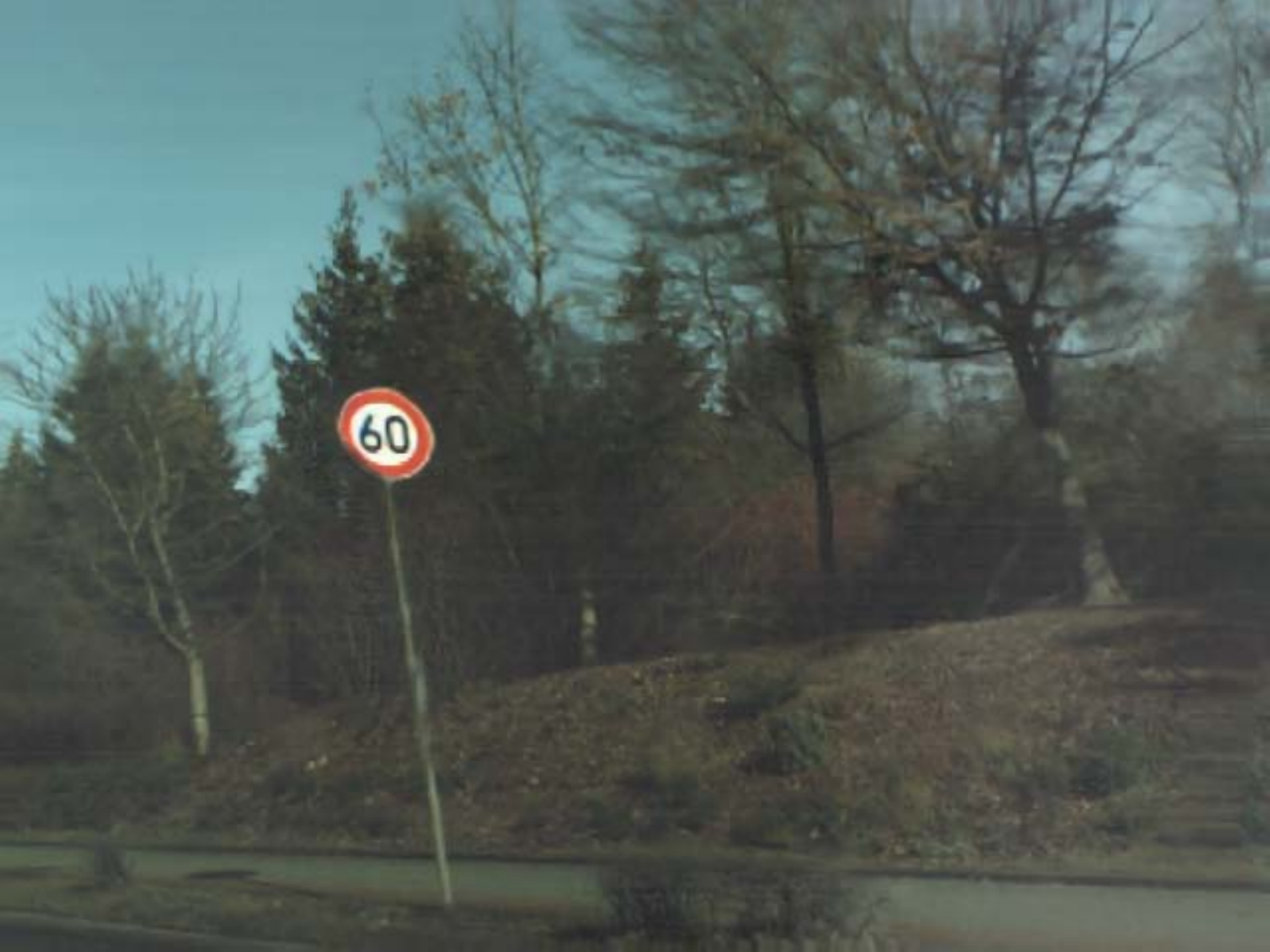}
		&
		\includegraphics[width=0.78\itemwidth]{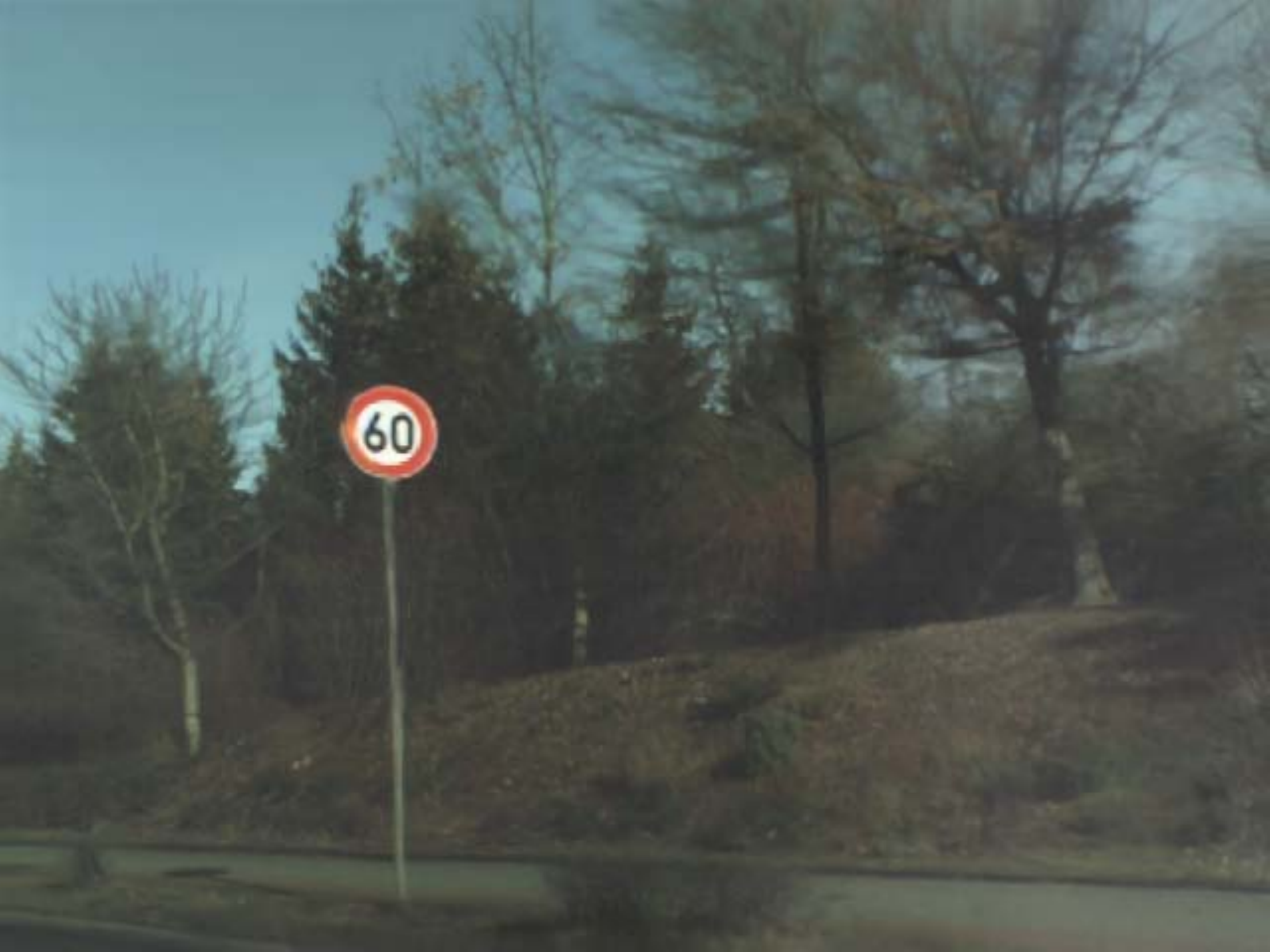}
		\vspace{-0.155cm} \\
		\scriptsize Input RS~(Overlayed)
		&
		\scriptsize DiffSfM~\cite{zhuang2017rolling}
		&
		\scriptsize DiffHomo~\cite{zhuang2020homography}
		&
		\scriptsize BMBC~\cite{park2020bmbc}
		&
		\scriptsize DAIN~\cite{bao2019depth}
		&
		\scriptsize Cascaded method
		\vspace{-0.01cm} \\
		\includegraphics[width=0.78\itemwidth]{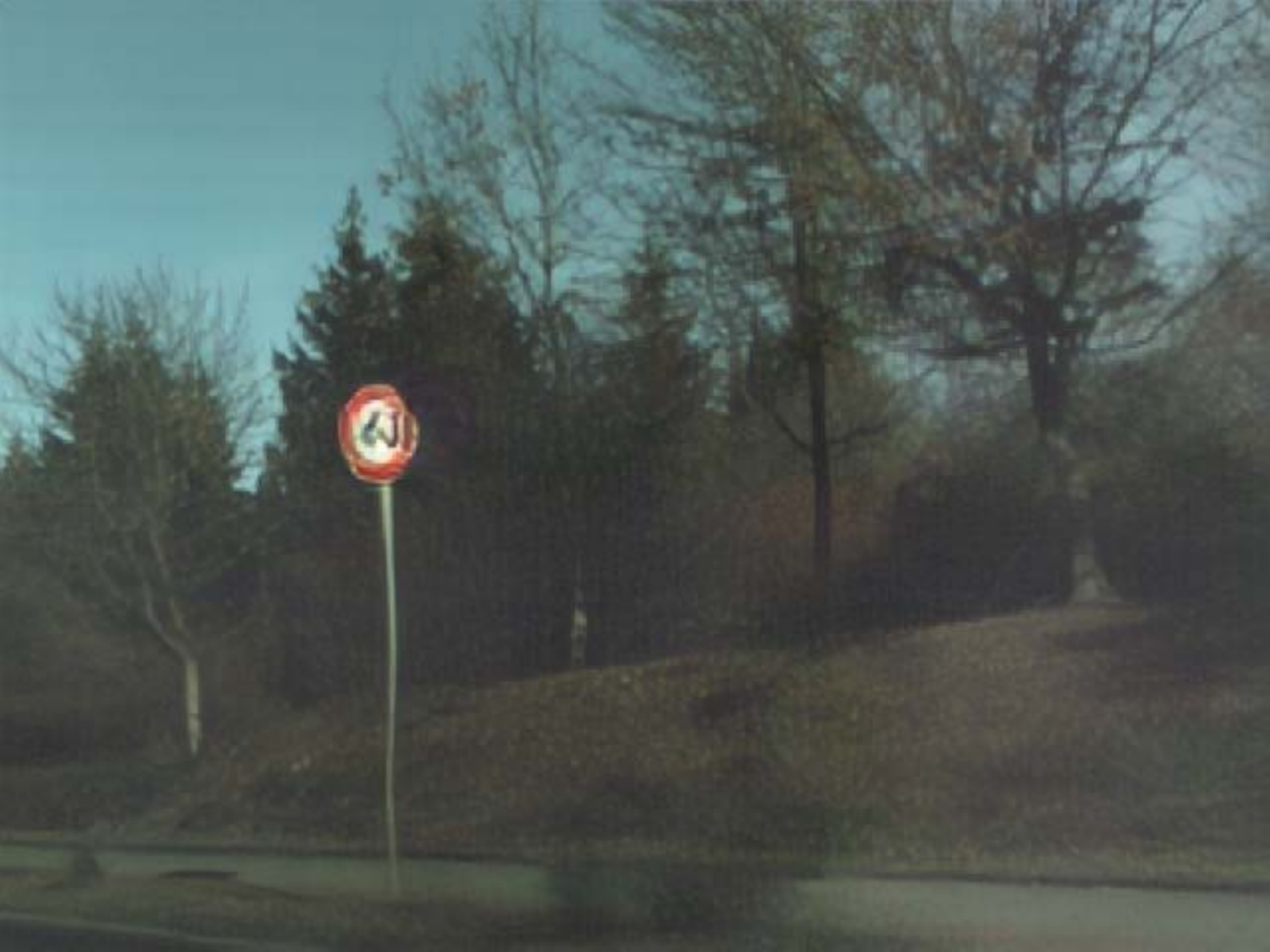}
		&
		\includegraphics[width=0.78\itemwidth]{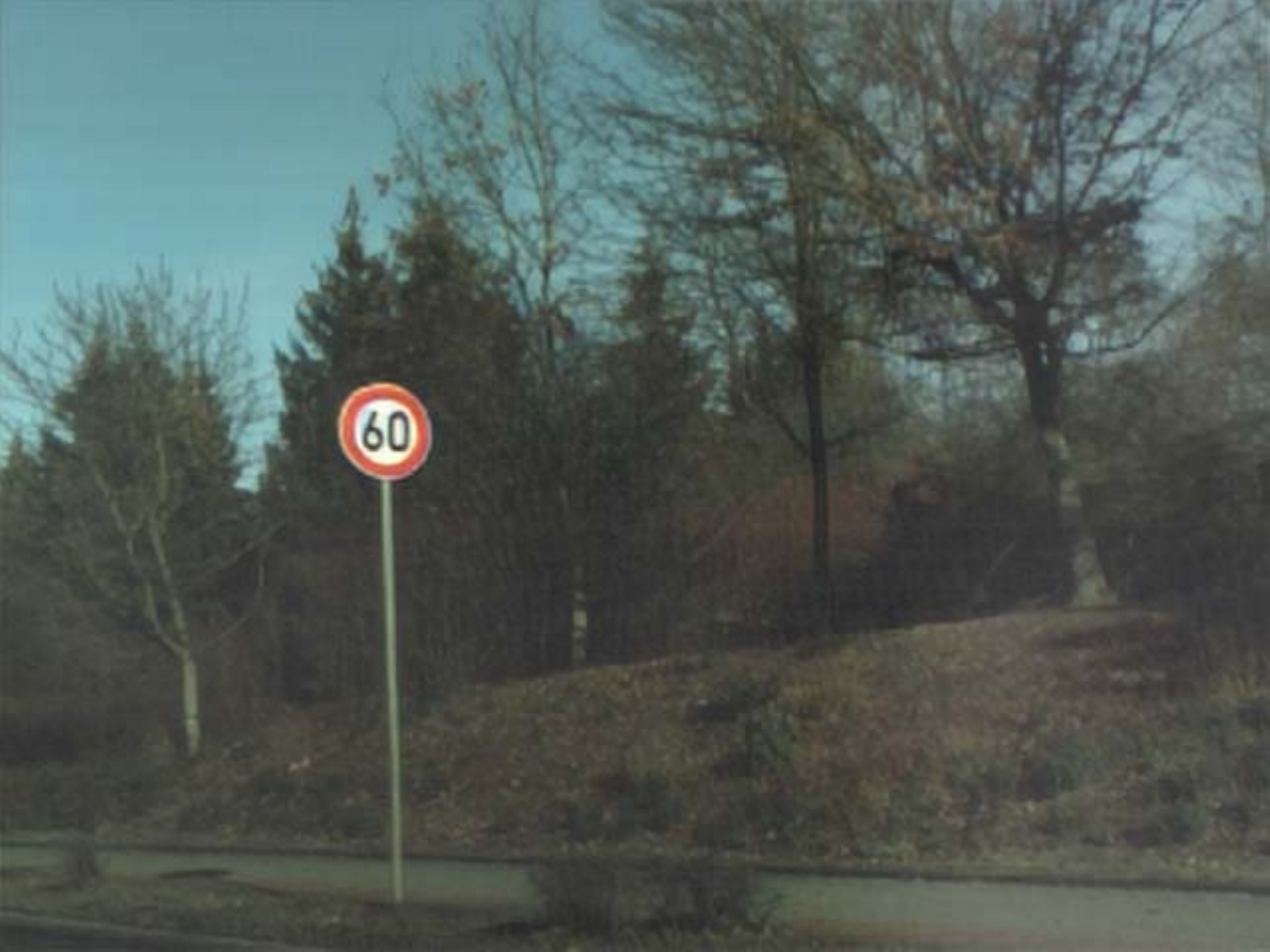}
		&
		\includegraphics[width=0.78\itemwidth]{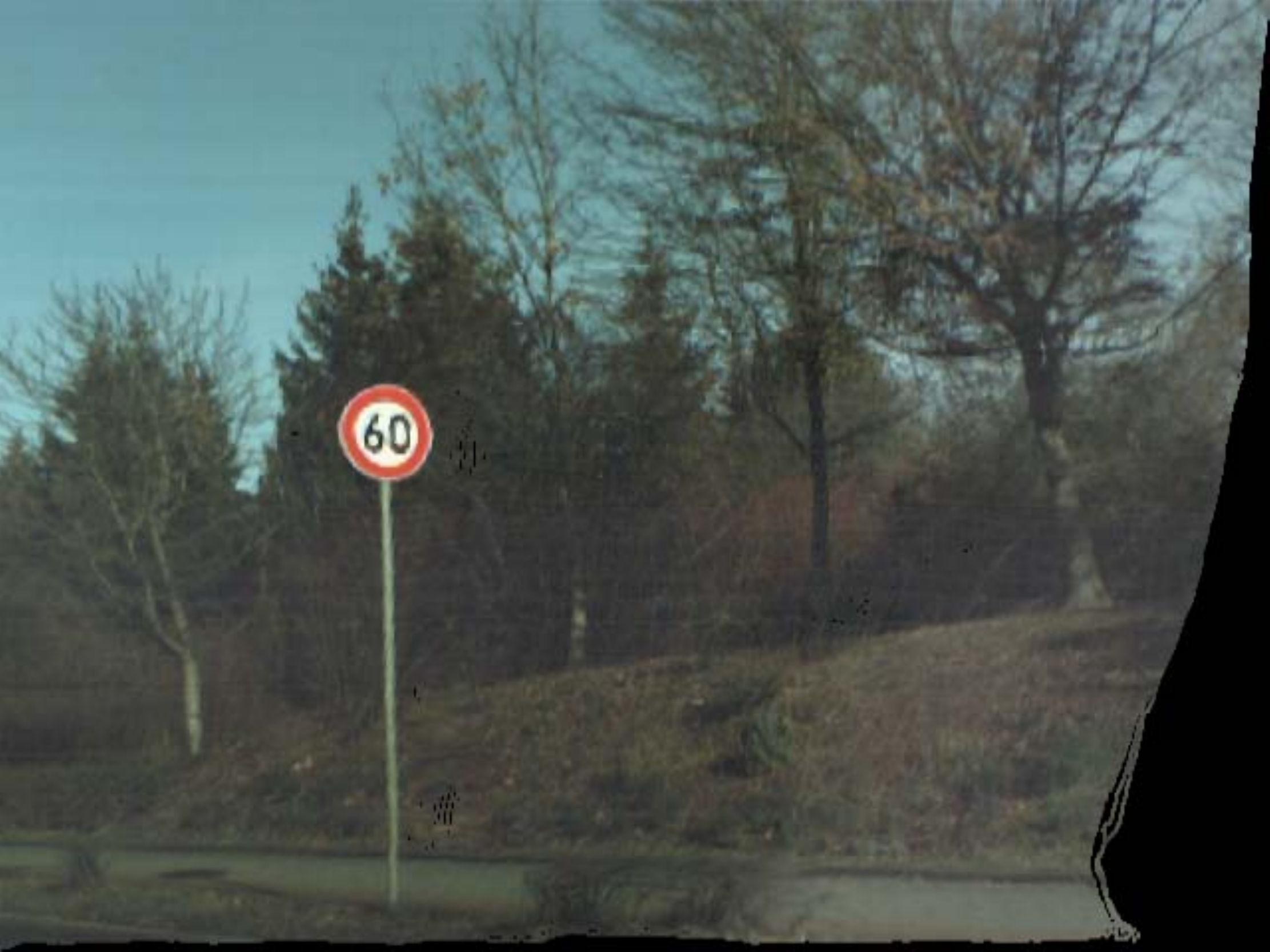}
		&
		\includegraphics[width=0.78\itemwidth]{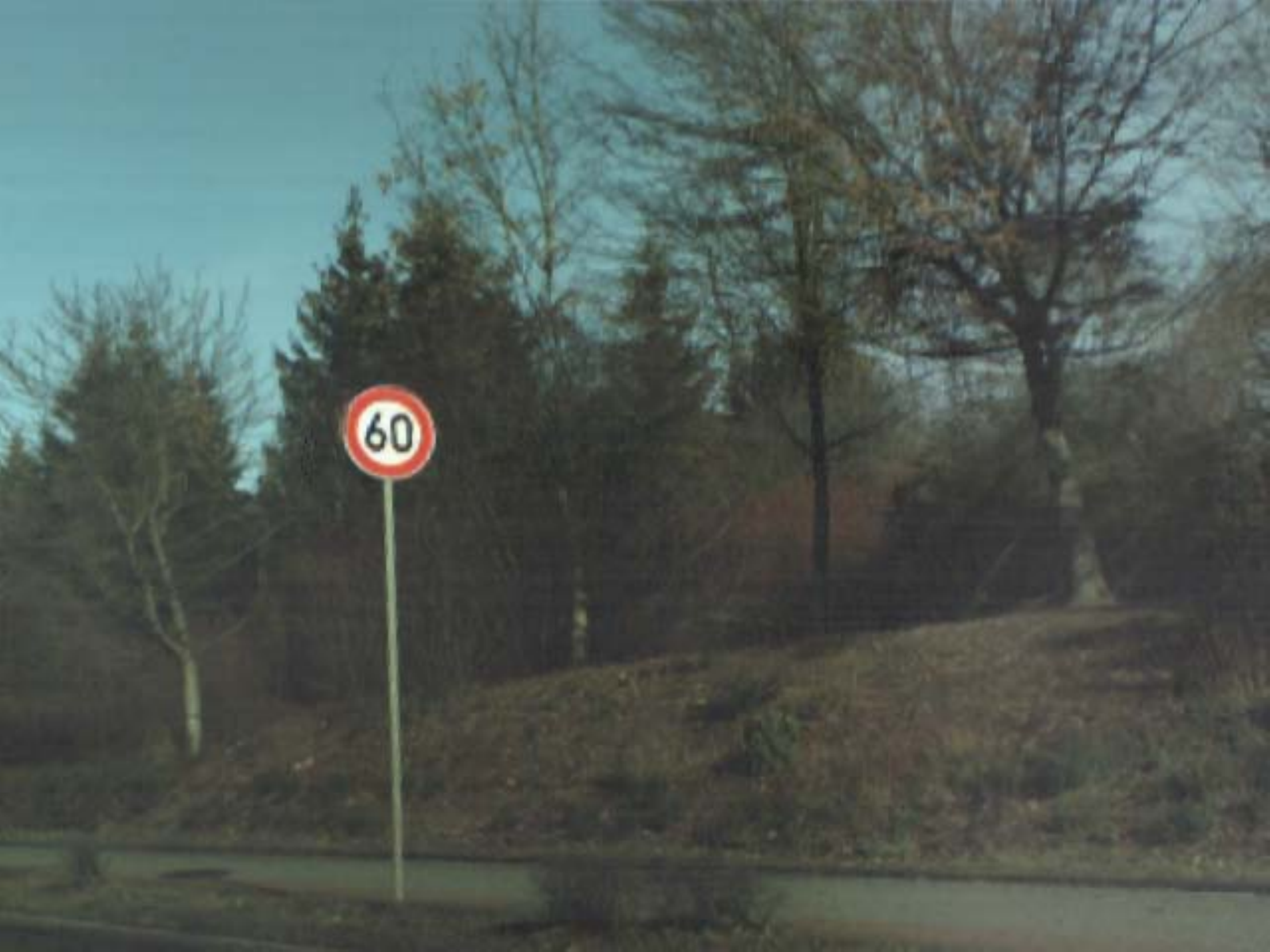}
		&
		\includegraphics[width=0.78\itemwidth]{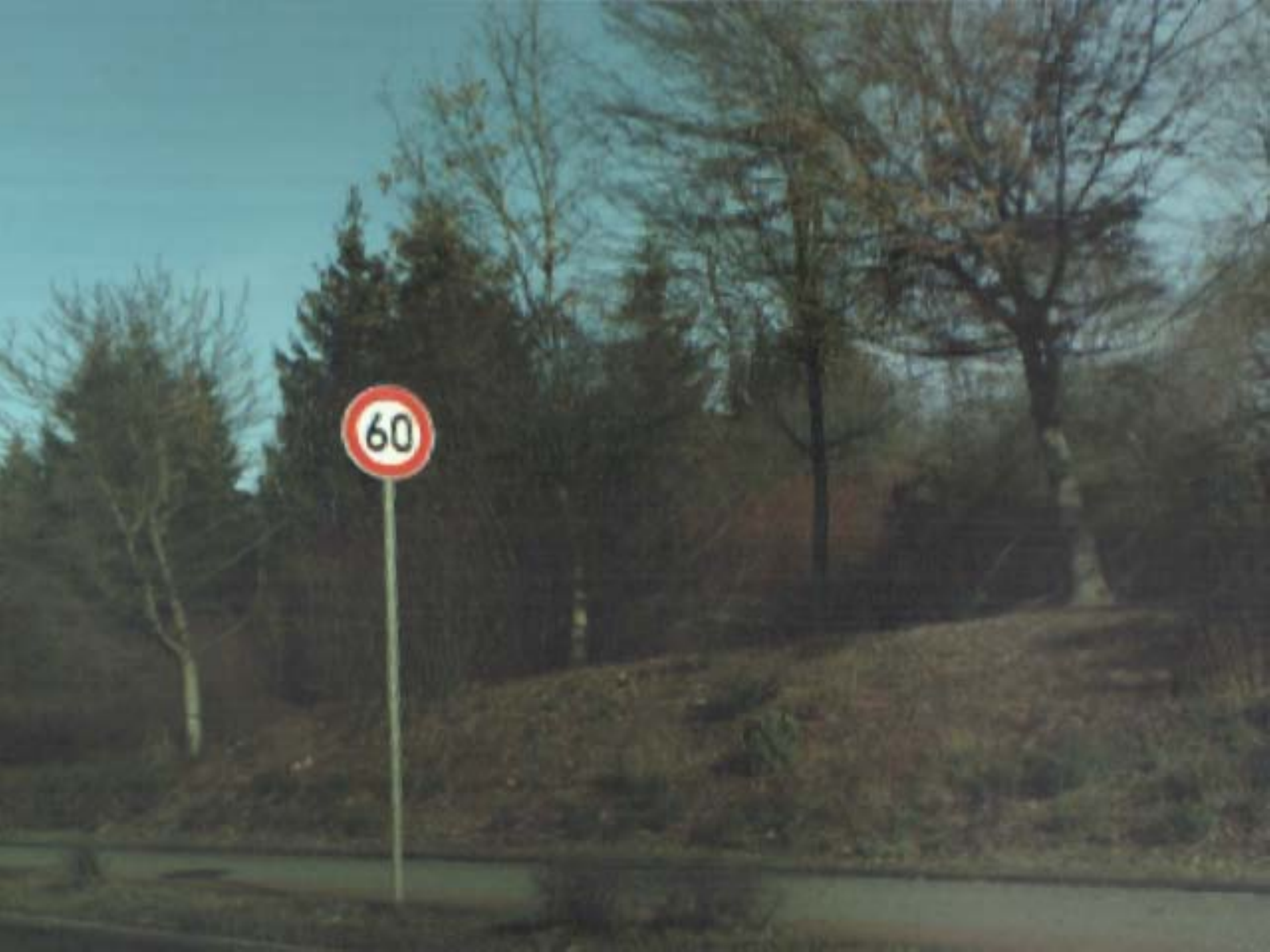}
		&
		\includegraphics[width=0.78\itemwidth]{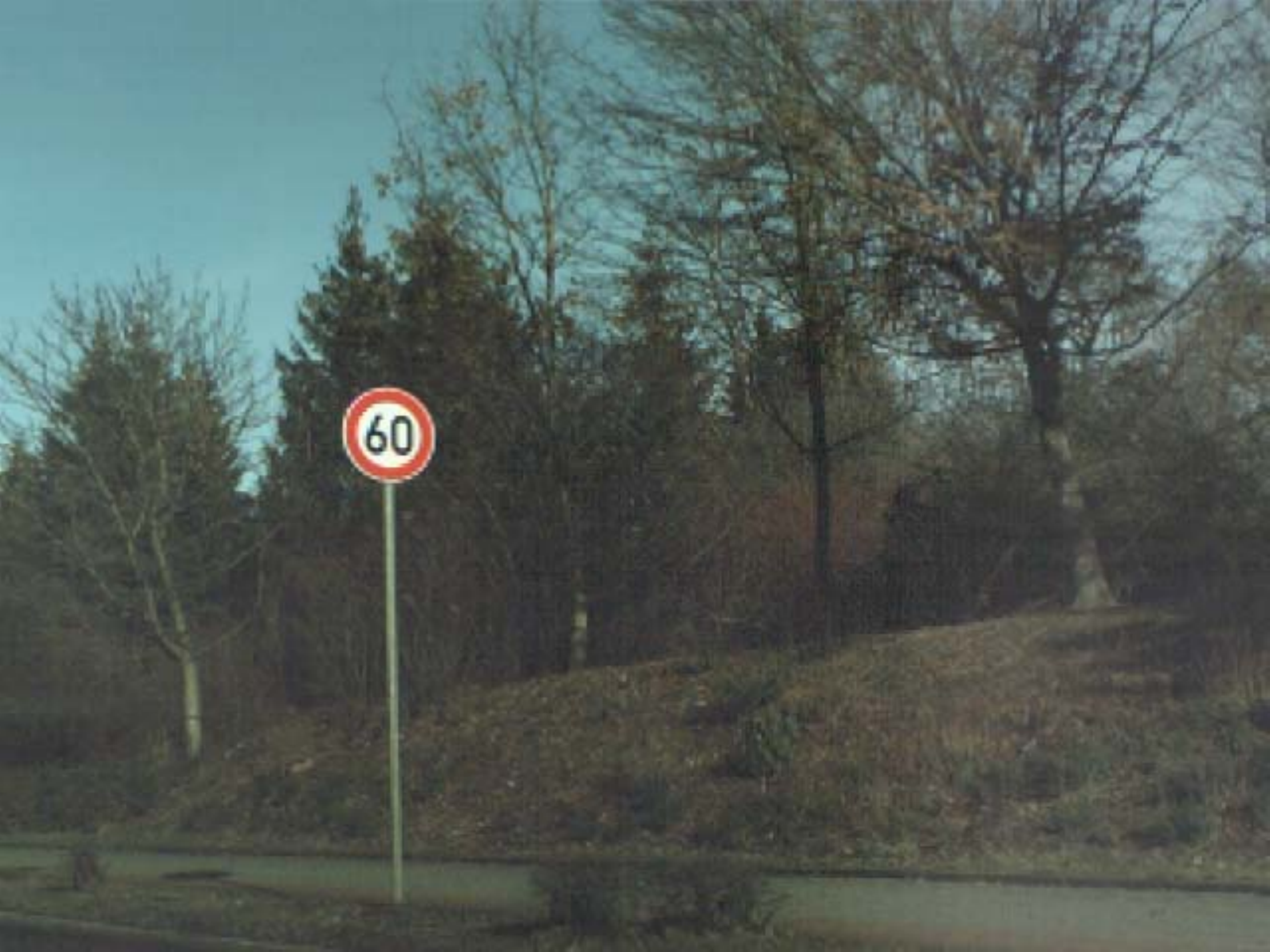}
		\vspace{-0.155cm} \\
		\scriptsize DeepUnrollNet~\cite{liu2020deep}
		&
		\scriptsize SUNet~\cite{fan2021sunet}
		&
		\scriptsize RSSR~\cite{fan2021rssr}
		&
		\scriptsize CVR* (Ours)
		&
		\scriptsize CVR (Ours)
		&
		\scriptsize Ground-truth
		\\
	\end{tabular}\vspace{-0.335cm}
	\caption{Qualitative results against baselines. Our method can successfully remove RS artifacts, yielding higher fidelity GS images.}
	\label{fig:exp_first_row}
	\vspace{-2.2mm}
\end{figure*}

\begin{table*}[!t]
	\footnotesize
	\caption{Quantitative comparisons on recovering GS images at time step $t=0.5$. The numbers in {\color{red}\textbf{red}} and {\color{blue}\underline{blue}} represent the best and second-best performance. Our method is far superior to baseline methods and the proposed ABMF model is effective as an initialization.} \label{tab:first_comparison}
	\vspace{-2.6mm}
	\centering
	\setlength{\tabcolsep}{3mm}{
		\begin{tabular}{lcccccccccc}
			\hline
			\multirow{2}{*}{Method}  &Runtime &\multicolumn{3}{c}{PSNR$\uparrow$ (dB)}      &   & \multicolumn{2}{c}{SSIM$\uparrow$} &   & \multicolumn{2}{c}{LPIPS$\downarrow$}     \\ \cline{3-5} \cline{7-8} \cline{10-11}
			&(seconds)  & CRM            & CR             & FR            & & CR            & FR   & & CR            & FR         \\ \hline
			DiffSfM \cite{zhuang2017rolling} &467  & 24.20    & 21.28     & 20.14   & & 0.775     & 0.701  & & 0.1322     & 0.1789        \\ 
			DiffHomo \cite{zhuang2020homography}  &424  & 19.60    & 18.94     & 18.68   & & 0.606     & 0.609      & & 0.1798     & 0.2229    \\ \hline
			DeepUnrollNet \cite{liu2020deep}        &0.34  & 26.90    & 26.46     & 26.52   & & 0.807  & 0.792   & & 0.0703    & 0.1222    \\
			SUNet \cite{fan2021sunet}        &0.21  & 29.28    & {29.18}     & {28.34}   & & 0.850  & {0.837}    & & {0.0658}    & {0.1205}    \\ \hline
			RSSR*  &0.09 & 28.20 & 23.86 & 21.02 & & 0.839 & 0.768 & & 0.0764 & 0.1866 \\
			RSSR \cite{fan2021rssr} &0.12  & {30.17} & 24.78 & 21.23 & & {0.867} & 0.776 & & 0.0695     & 0.1659\\ \hline
			\emph{CVR*} (Ours) &0.12  & \color{blue}\underline{31.82} & \color{blue}\underline{31.60} & \color{blue}\underline{28.62} & & \color{blue}\underline{0.927} & \color{blue}\underline{0.845} & & \color{blue}\underline{0.0372}     & \color{blue}\underline{0.1117}\\ 
			\emph{CVR} (Ours) &0.14  & \color{red}\textbf{32.02} & \color{red}\textbf{31.74} & \color{red}\textbf{28.72} & & \color{red}\textbf{0.929} & \color{red}\textbf{0.847} & & \color{red}\textbf{0.0368}     & \color{red}\textbf{0.1107}\\ \hline
		\end{tabular}
		\normalsize
		\vspace{-0.55mm}
		\begin{tablenotes}
			\raggedleft
			\item{
				\small{*: \emph{applying our proposed approximated bilateral motion field (ABMF) model.}}
			}
	\end{tablenotes}}
	\vspace{-4.0mm}
\end{table*}

\subsection{Loss Function} \label{subsec:Loss}
\vspace{-1.0mm}
Similar to \cite{fan2021rssr,liu2020deep,zhong2021rscd}, we use the reconstruction loss $\mathcal{L}_r$, the perceptual loss $\mathcal{L}_p$ \cite{johnson2016perceptual}, and the total variation loss $\mathcal{L}_{tv}$ to improve the quality of final GS and BMF predictions. Moreover, inspired by \cite{fan2021sunet}, we propose a contextual consistency constraint loss $\mathcal{L}_c$ to enforce the alignment of refined intermediate GS frame candidates with ground-truth, which is crucial to facilitate occlusion inference and motion compensation. In short, our loss function $\mathcal{L}$ is defined as: 
\begin{equation}\label{eq:10}
\mathcal{L} = \lambda_r\mathcal{L}_r + \mathcal{L}_p + \lambda_c\mathcal{L}_c + \lambda_{tv}\mathcal{L}_{tv},
\end{equation}
where $\lambda_r$, $\lambda_c$ and $\lambda_{tv}$ are hyper-parameters.
More details can be found in Appendix~\ref{sec:details_loss}.

\begin{figure*}[!t]
	\centering
	\includegraphics[width=0.988\textwidth]{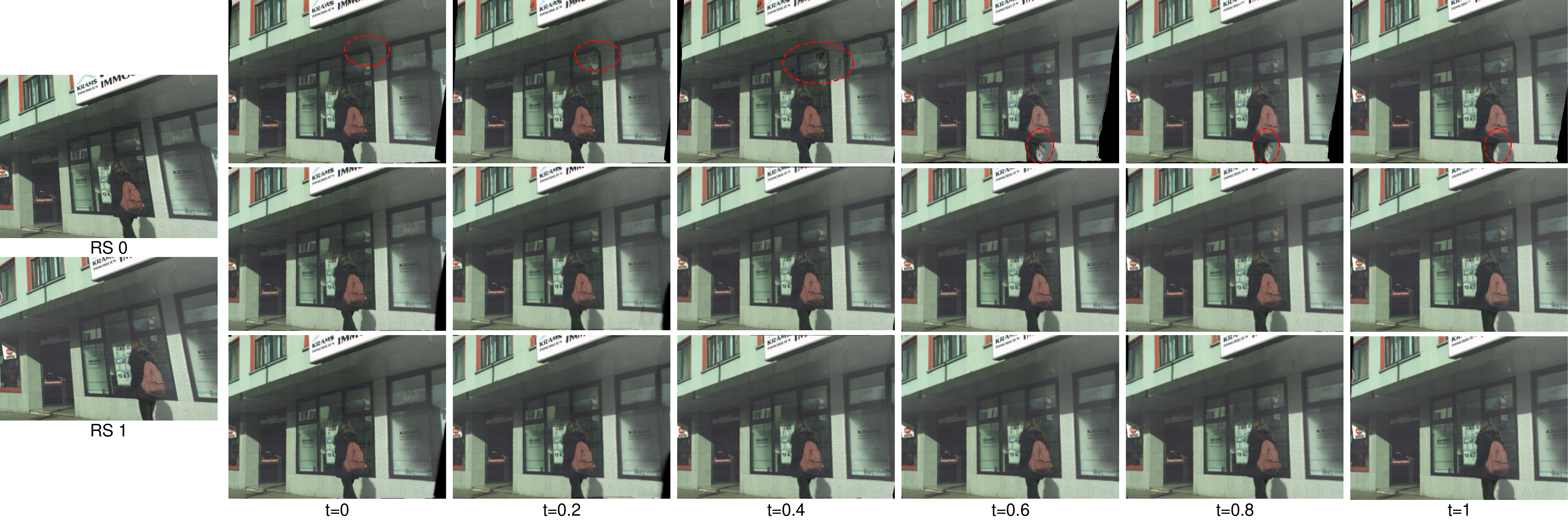}\vspace{-2.8mm}
	\caption{Example results of recovering six GS video images from the two input RS images (left column) by using RSSR \cite{fan2021rssr}, \emph{CVR*}, and \emph{CVR} (three rows from top to bottom), respectively. Apart from many unfriendly black holes at the GS image edges, RSSR generates local errors and motion artifacts as shown in red circles. Our method can produce temporally consistent GS sequences with richer details. }
	\label{fig:multiple_frames}
	\vspace{-5.60mm}
\end{figure*}

\vspace{-1.0mm}
\section{Experimental Setup} \label{sec:experiment_setup}
\vspace{-1.0mm}
\noindent\textbf{Datasets.}
We use the standard RS correction benchmark datasets \cite{liu2020deep} including Carla-RS and Fastec-RS, and divide the training and test sets as in \cite{liu2020deep}.
The Carla-RS dataset is synthesized based on the Carla simulator \cite{dosovitskiy2017carla}, involving general 6-DOF camera motions.
The Fastec-RS dataset records real-world RS images synthesized by a high-FPS GS camera mounted on a ground vehicle.
Since they provide the first- and central-scanline GT supervisory signals, \ie $t=0$, $0.5$, and $1$, we utilize this triplet as GT to train our network. Note that we add a small perturbation to make Eq.~\eqref{eq:9} work properly, for example, transforming them to $t=0.01$, $0.5$, and $0.99$, respectively.
At the test phase, our method is capable of recovering GS video frames at any time $t\in[0,1]$.

\vspace{0.25mm}
\noindent\textbf{Training details.}
Our method is trained end-to-end using the Adam optimizer \cite{Kingma_Adam_ICLR_2015} with $\beta_1=0.9$ and $\beta_2=0.999$. 
We empirically set $\lambda_r=10$, $\lambda_c=5$, and $\lambda_{tv}=0.1$. 
The experiments are performed on an NVIDIA GeForce RTX 2080Ti GPU with a batch size of 4.
We propose to train our network in two stages. 
Firstly, we solely train $\mathcal{M}$.
To train the ABMF-based $\mathcal{M}_A$, we fine-tune PWC-Net \cite{sun2018pwc} for 100 epochs from its pre-trained model on the RS benchmark in a self-supervised way \cite{jonschkowski2020matters,wang2018occlusion,liu2020self,fan2021rssr}, and then ABMF can be computed directly and explicitly.
Note that the training details of the NBMF-based $\mathcal{M}_N$ can be found in \cite{fan2021rssr} with the supervision of central-scanline GT GS images.
Secondly, we jointly train the entire model (\ie $\mathcal{M}$ and $\mathcal{G}$) by $\mathcal{L}$ for another 50 epochs.
At this time, the learning rate of $\mathcal{G}$ is set to $10^{-4}$ for training from scratch, and that of $\mathcal{M}$ is set to $10^{-5}$ for fine-tuning. 
We keep the vertical resolution constant and adopt a uniform random crop with a horizontal resolution of 256 pixels to augment the training data, similar to \cite{fan2021rssr,fan2021sunet} for better contextual exploration.

\vspace{0.25mm}
\noindent\textbf{Evaluation strategies.} As the Carla-RS dataset has the GT occlusion mask, we perform quantitative evaluations as follows: Carla-RS dataset with occlusion mask (\emph{CRM}), Carla-RS dataset without occlusion mask (\emph{CR}), and Fastec-RS dataset (\emph{FR}).
Standard metrics PSNR and SSIM, and learned perceptual metric LPIPS \cite{zhang2018unreasonable} are applied.
Higher PSNR/SSIM or lower LPIPS score indicates better quality.
Note that unless otherwise stated, \emph{we refer to the GS images at time $t=0.5$ for consistent comparisons.}

\vspace{0.38mm}
\noindent\textbf{Baselines.}
We perform comparisons with the following baselines. 
(i) \textbf{DiffSfM} \cite{zhuang2017rolling} and \textbf{DiffHomo} \cite{zhuang2020homography} are traditional two-image based RS correction methods that require sophisticated optimization using RS models.
(ii) \textbf{SUNet} \cite{fan2021sunet} and \textbf{DeepUnrollNet} \cite{liu2020deep} recover only one GS frame from two consecutive RS frames by designing specialized CNNs.
While \textbf{RSCD} \cite{zhong2021rscd} achieves this goal from three adjacent RS images.
(iii) \textbf{RSSR} \cite{fan2021rssr} generates a GS video from two consecutive RS images using deep learning, but suffers from black holes and motion artifacts. Moreover, we integrate the proposed ABMF model into RSSR to yield \textbf{RSSR*}.
(iv) \textbf{DAIN} \cite{bao2019depth} and \textbf{BMBC} \cite{park2020bmbc} are SOTA VFI methods that are tailored for GS cameras.
(v) \textbf{Cascaded method} generates two GS images sequentially from three consecutive RS inputs using DeepUnrollNet, and then interpolates in-between GS ones using DAIN.
(vi) \emph{\textbf{CVR}} and \emph{\textbf{CVR*}} are our proposed methods based on NBMF and ABMF, respectively. 
Note that our \emph{RSSR*}, RSSR, our \emph{CVR*}, and our \emph{CVR} form a clear hierarchy of RS-based video reconstruction methods.

\vspace{-1.0mm}
\section{Results and Analysis} \label{sec:experiment_results}
\vspace{-0.9mm}
In this section, we compare with the baseline approaches and provide analysis and insight into our method.

\vspace{-1.0mm}
\subsection{Comparison with SOTA Methods} \label{sec:Comparison_sota}
\vspace{-0.8mm}
We report the quantitative and qualitative results in Table~\ref{tab:first_comparison} and Fig.~\ref{fig:exp_first_row}, respectively.
Our proposed method achieves overwhelming dominance in RS effect removal, which is mainly attributed to context aggregation and motion pattern inference.
Furthermore, although our proposed ABMF model is inferior to RSSR \cite{fan2021rssr} when used to remove the RS effect (\ie RSSR*), it can serve as a strong baseline for GS video frame reconstruction when combined with GS frame refinement.
We believe that our hierarchical pipeline can provide a fresh perspective for the video reconstruction task with RS cameras. 
More results and analysis are shown in Appendix~\ref{sec:additional_results}.

\begin{figure*}[!t]
    \centering
	\setlength{\tabcolsep}{0.015cm}
	\setlength{\itemwidth}{3.532cm}
	\hspace*{-\tabcolsep}\begin{tabular}{cccccc}
		\includegraphics[width=0.78\itemwidth]{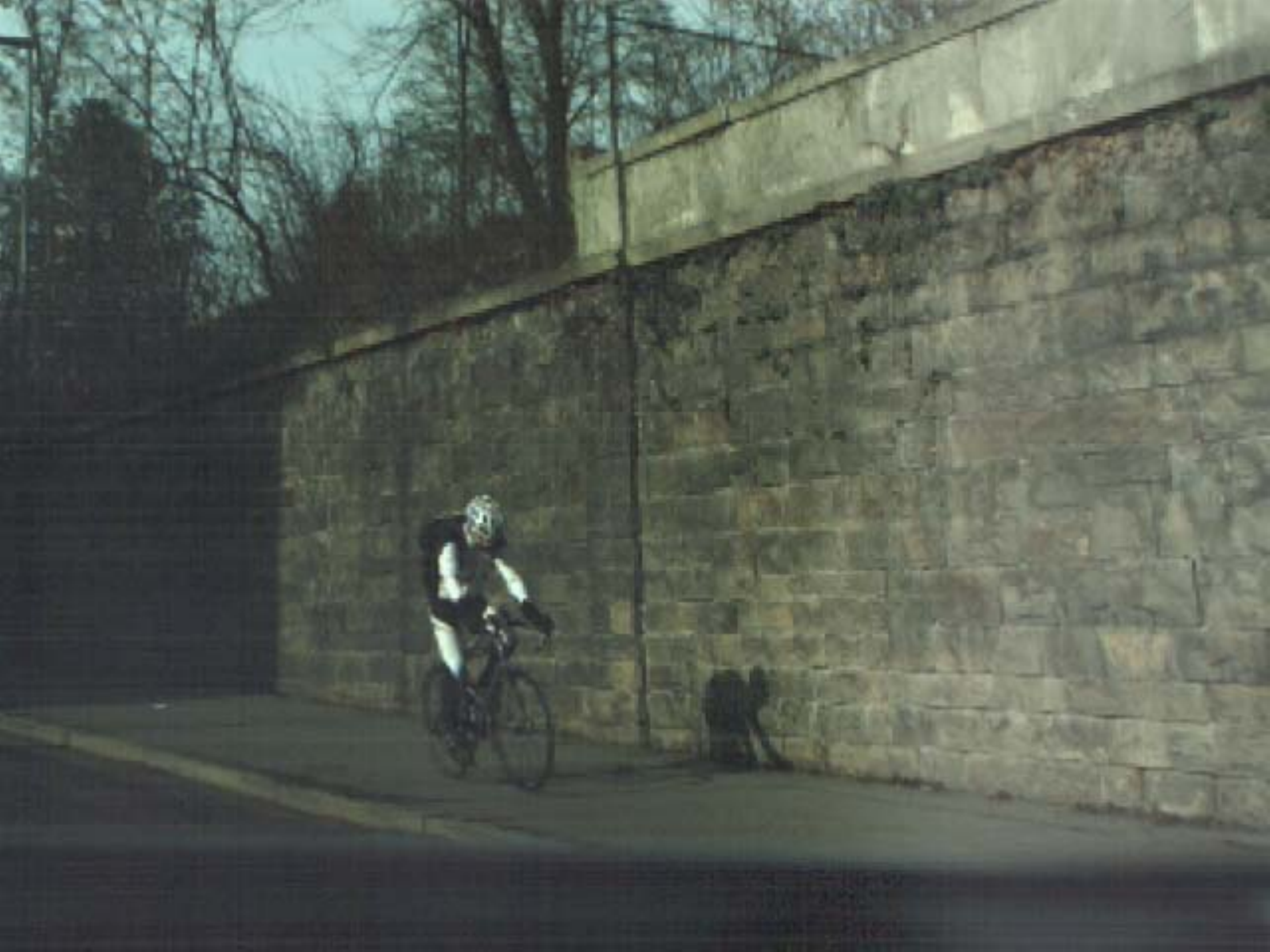}
		&
		\includegraphics[width=0.78\itemwidth]{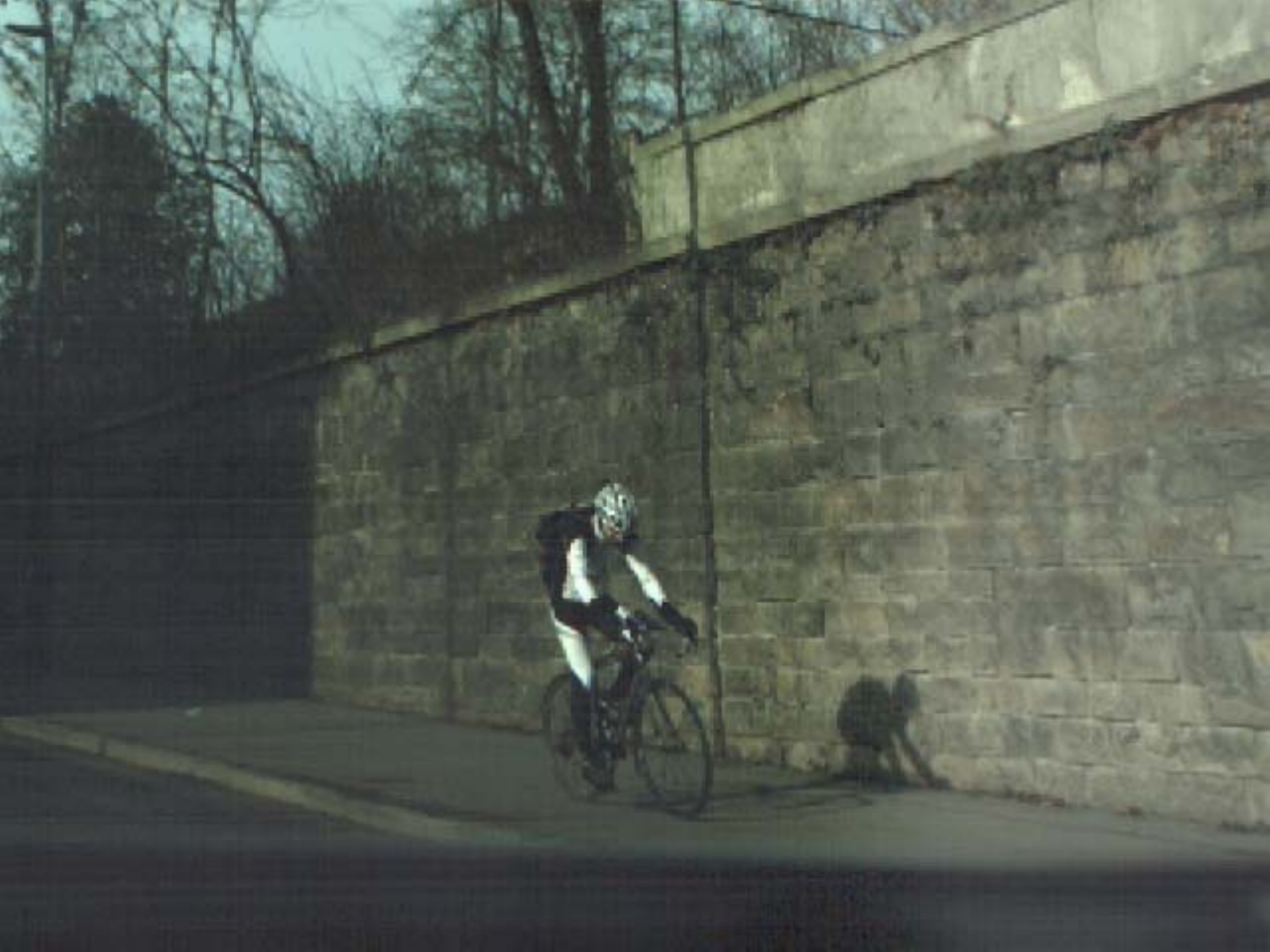}
		&
		\includegraphics[width=0.78\itemwidth]{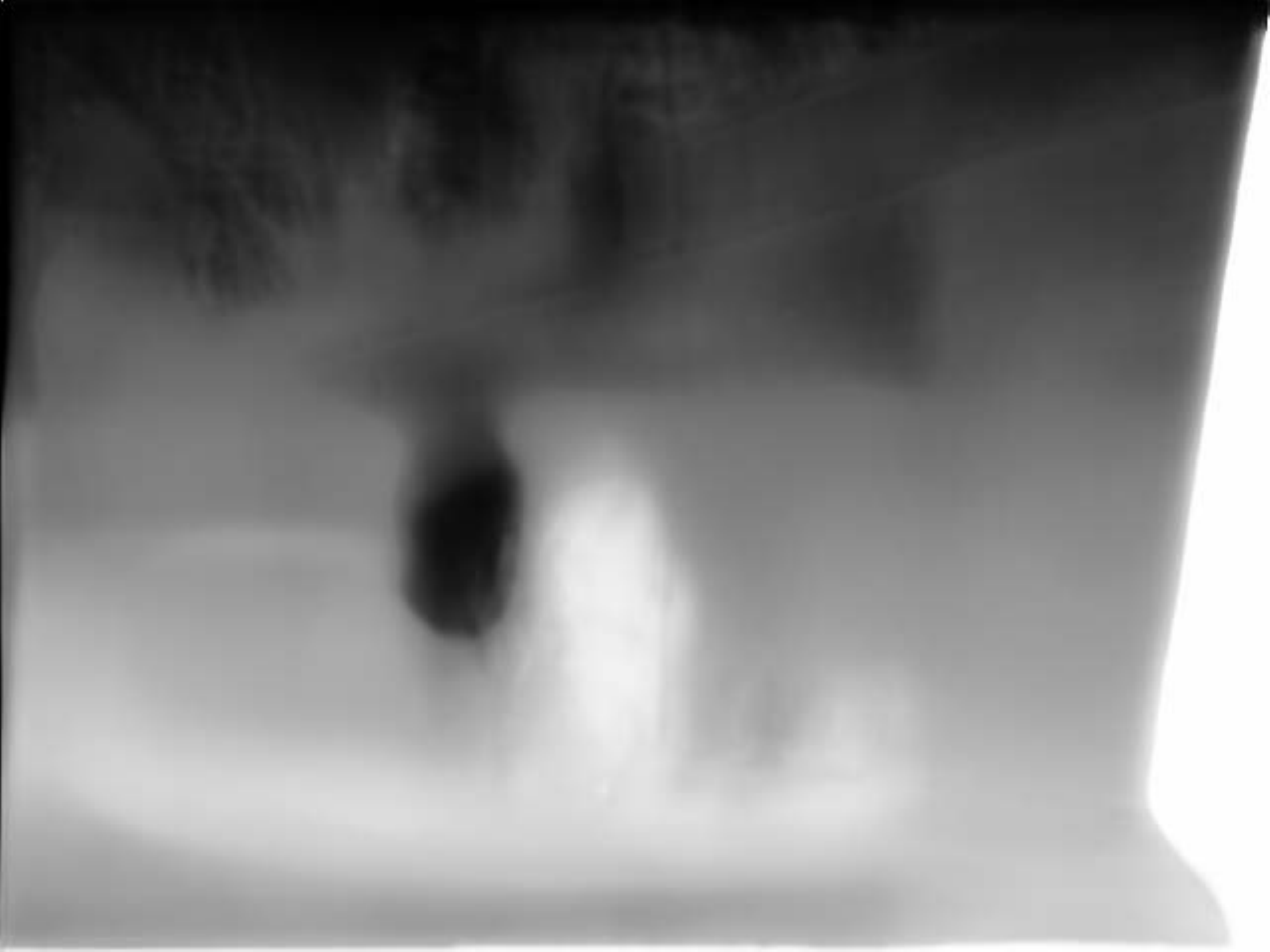}
		&
		\includegraphics[width=0.78\itemwidth]{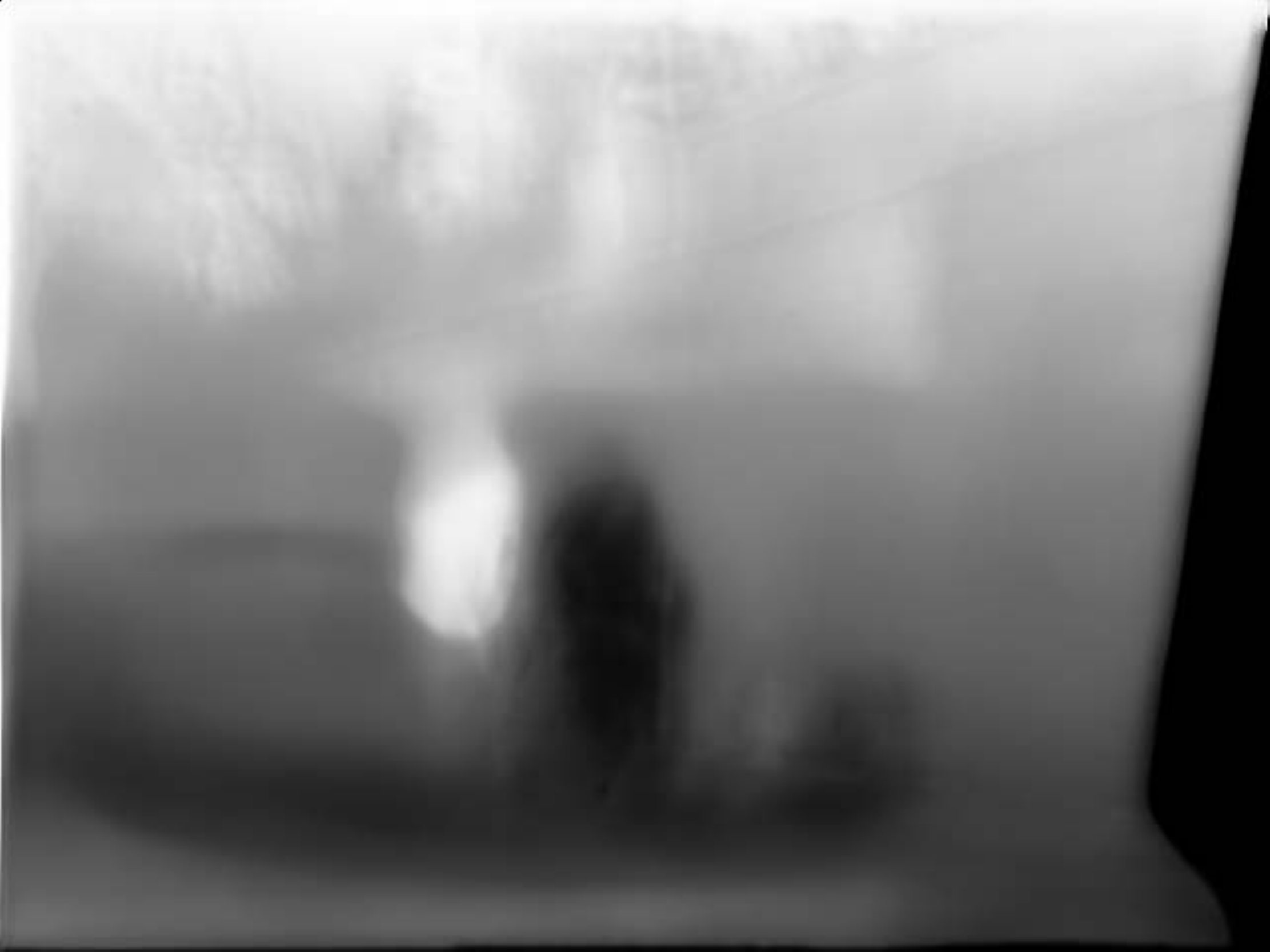}
		&
		\includegraphics[width=0.78\itemwidth]{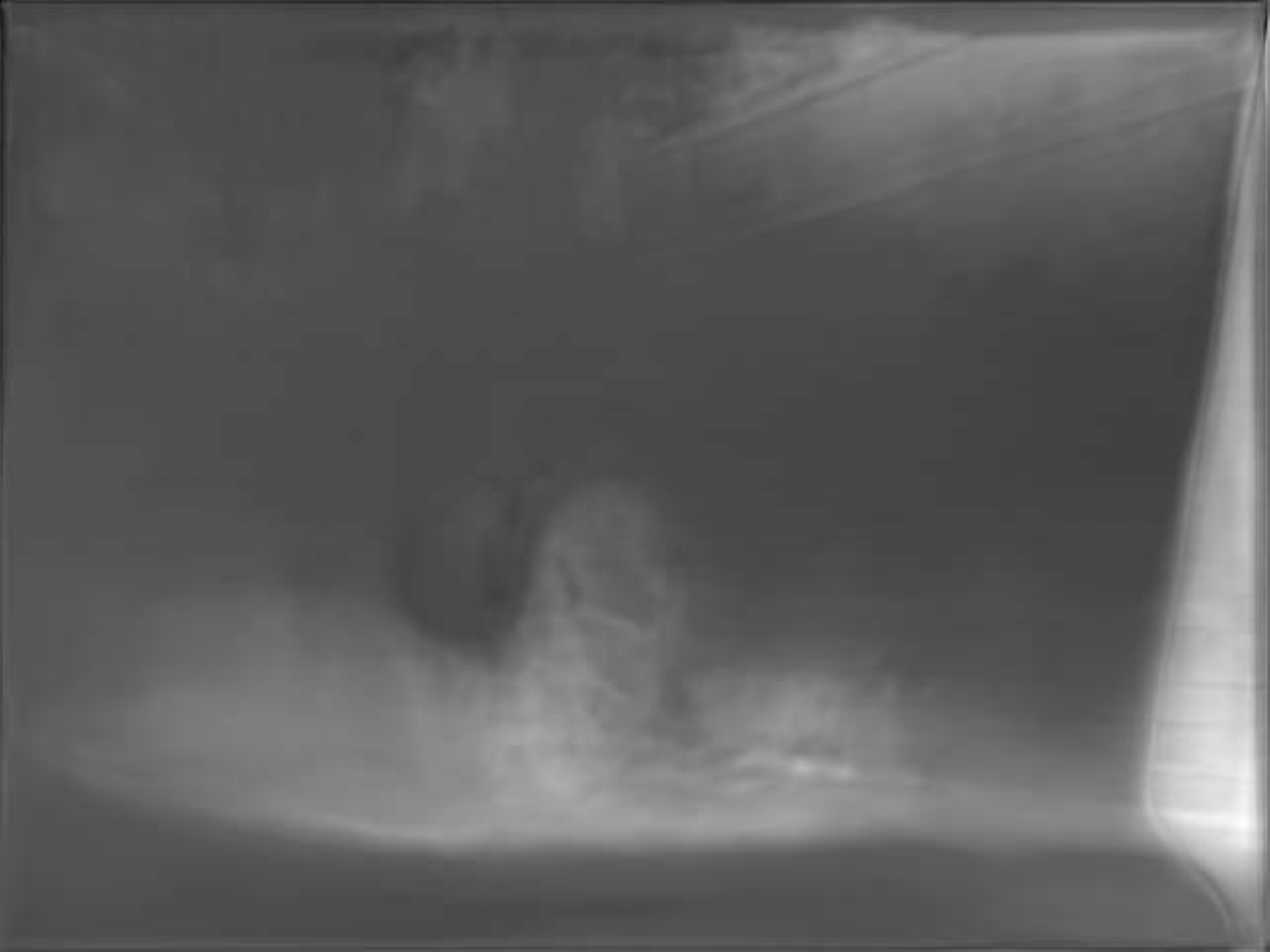}
		&
		\includegraphics[width=0.78\itemwidth]{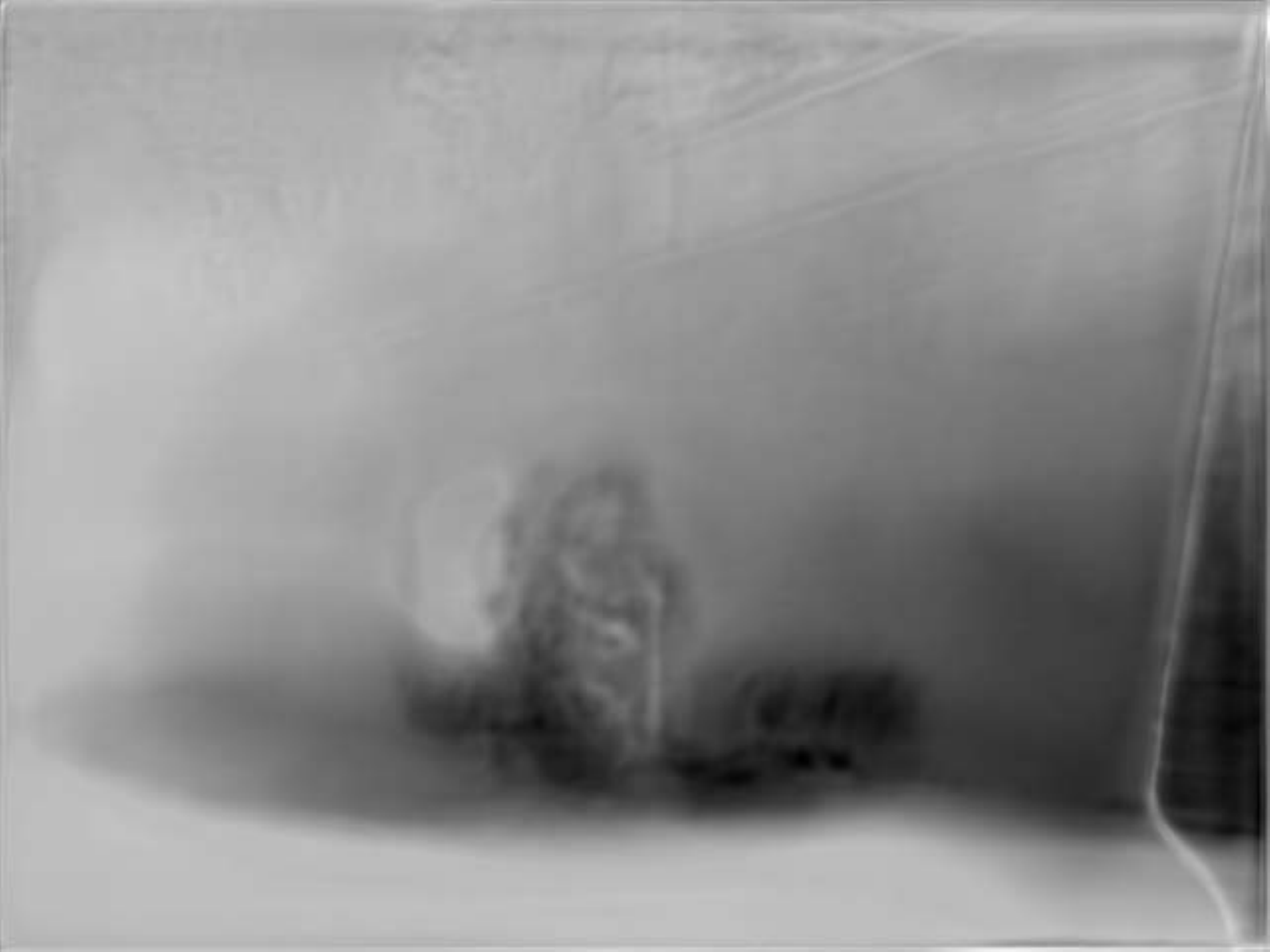}
		\vspace{-0.13cm} \\
		\scriptsize Input RS: ${\mathbf I}_{0}^r$
		&
		\scriptsize Input RS: ${\mathbf I}_{1}^r$
		&
		\scriptsize ${\mathbf O}_{0\to0.5}$
		&
		\scriptsize ${\mathbf O}_{1\to0.5}$
		&
		\scriptsize $\left\|\Delta{\mathbf U}_{0\to0.5}\right\|_{2}$
		&
		\scriptsize $\left\|\Delta{\mathbf U}_{1\to0.5}\right\|_{2}$
		\vspace{-0.017cm} \\
		\includegraphics[width=0.78\itemwidth]{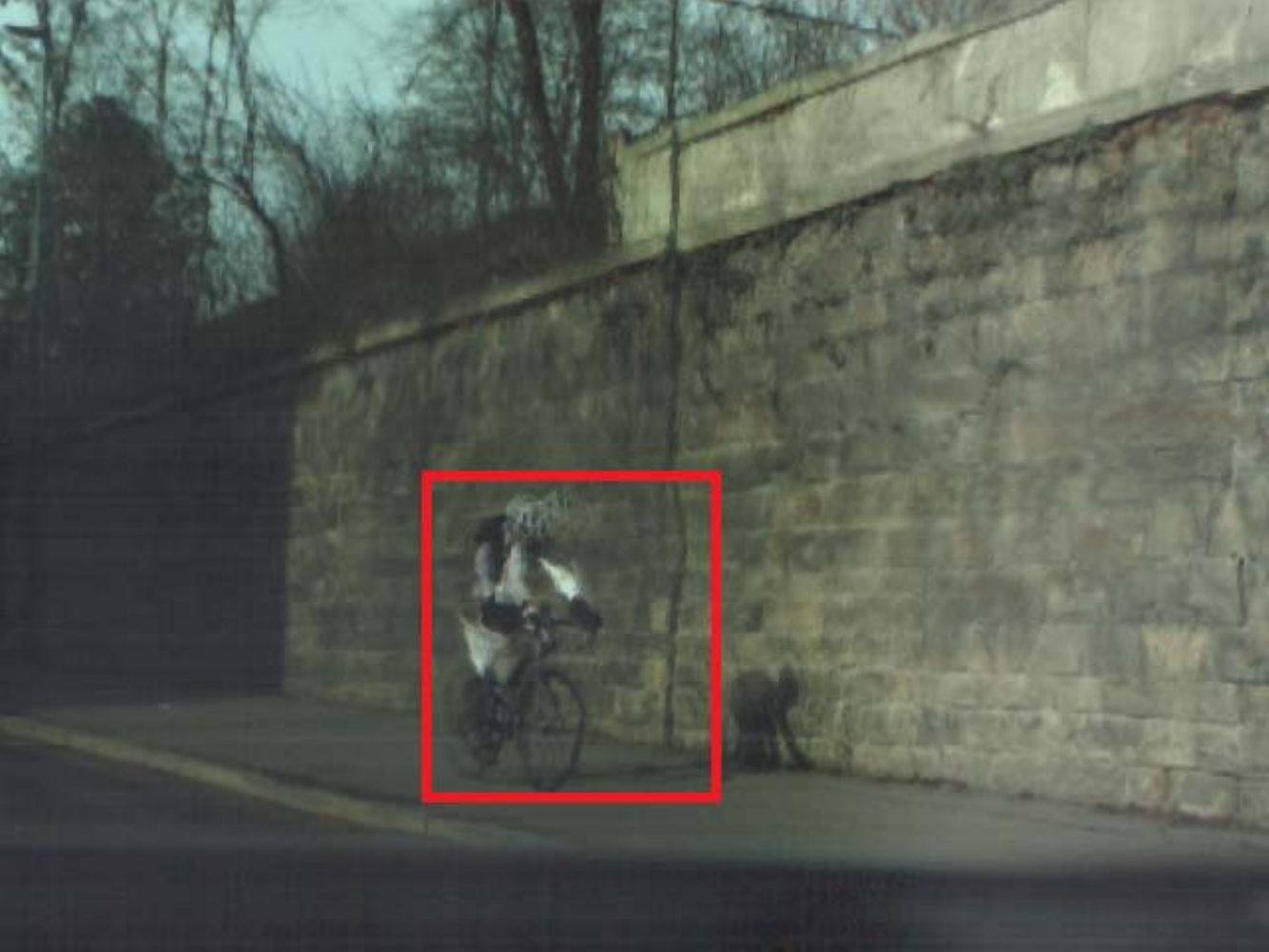}
		&
		\includegraphics[width=0.78\itemwidth]{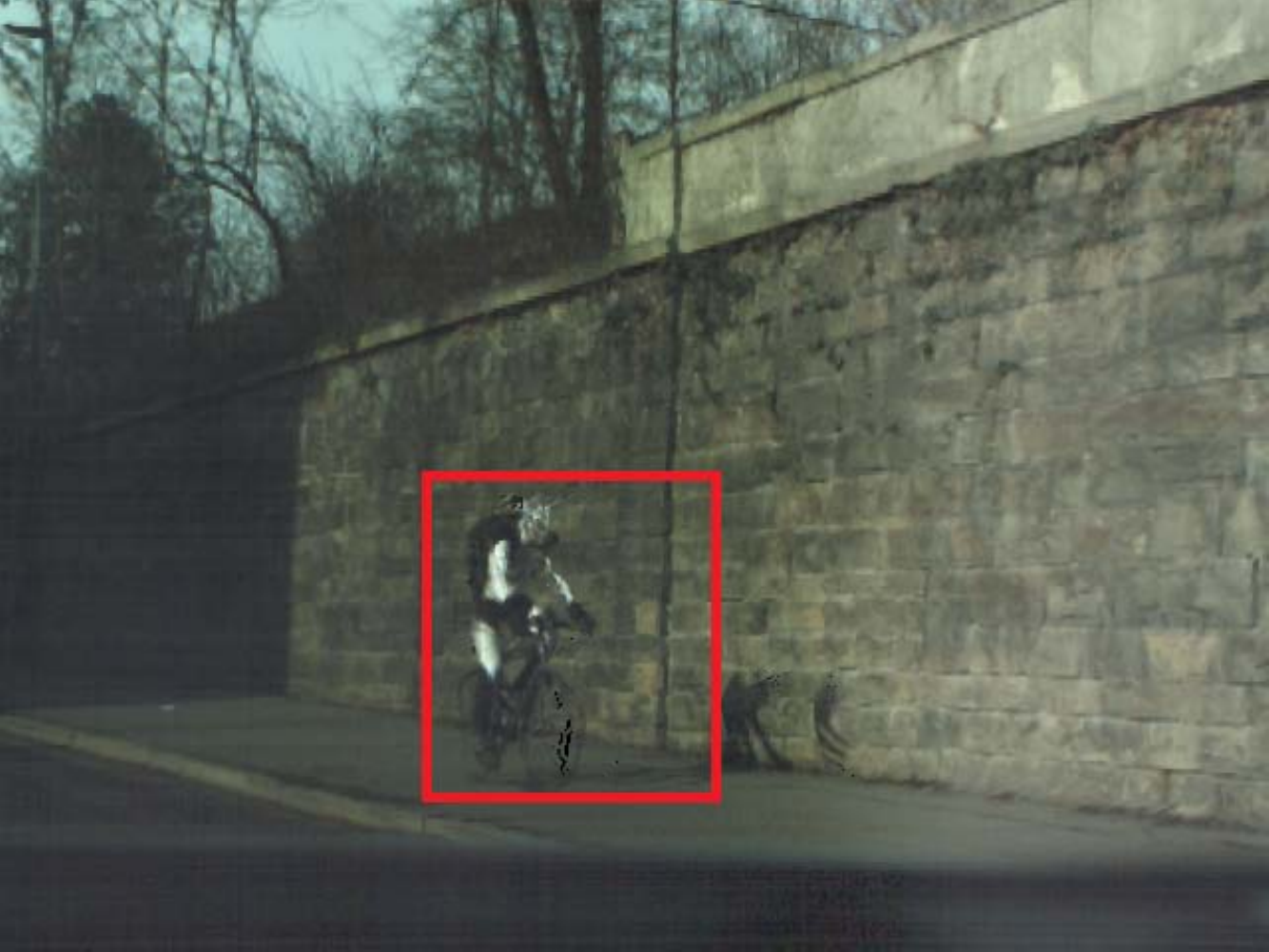}
		&
		\includegraphics[width=0.78\itemwidth]{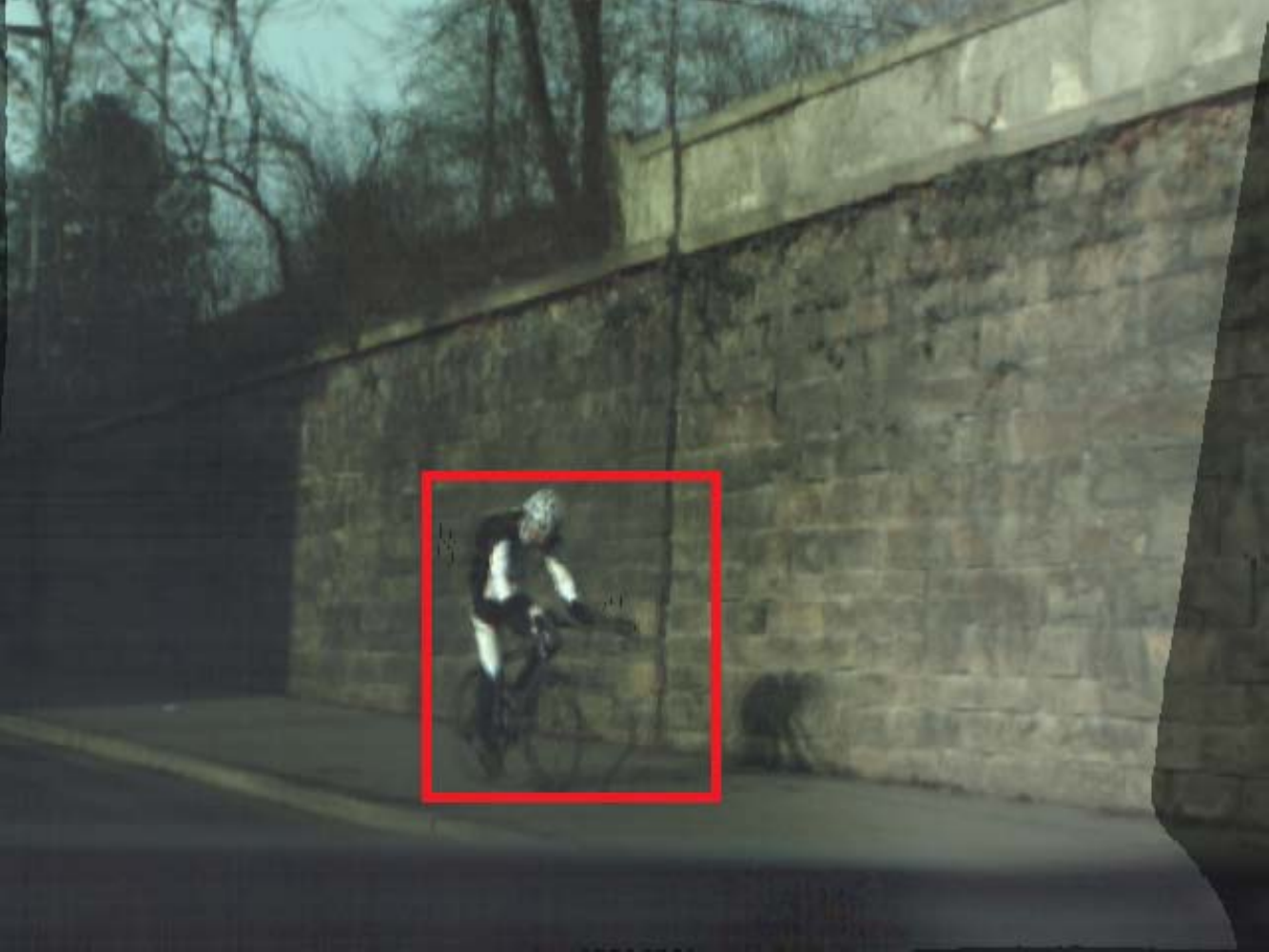}
		&
		\includegraphics[width=0.78\itemwidth]{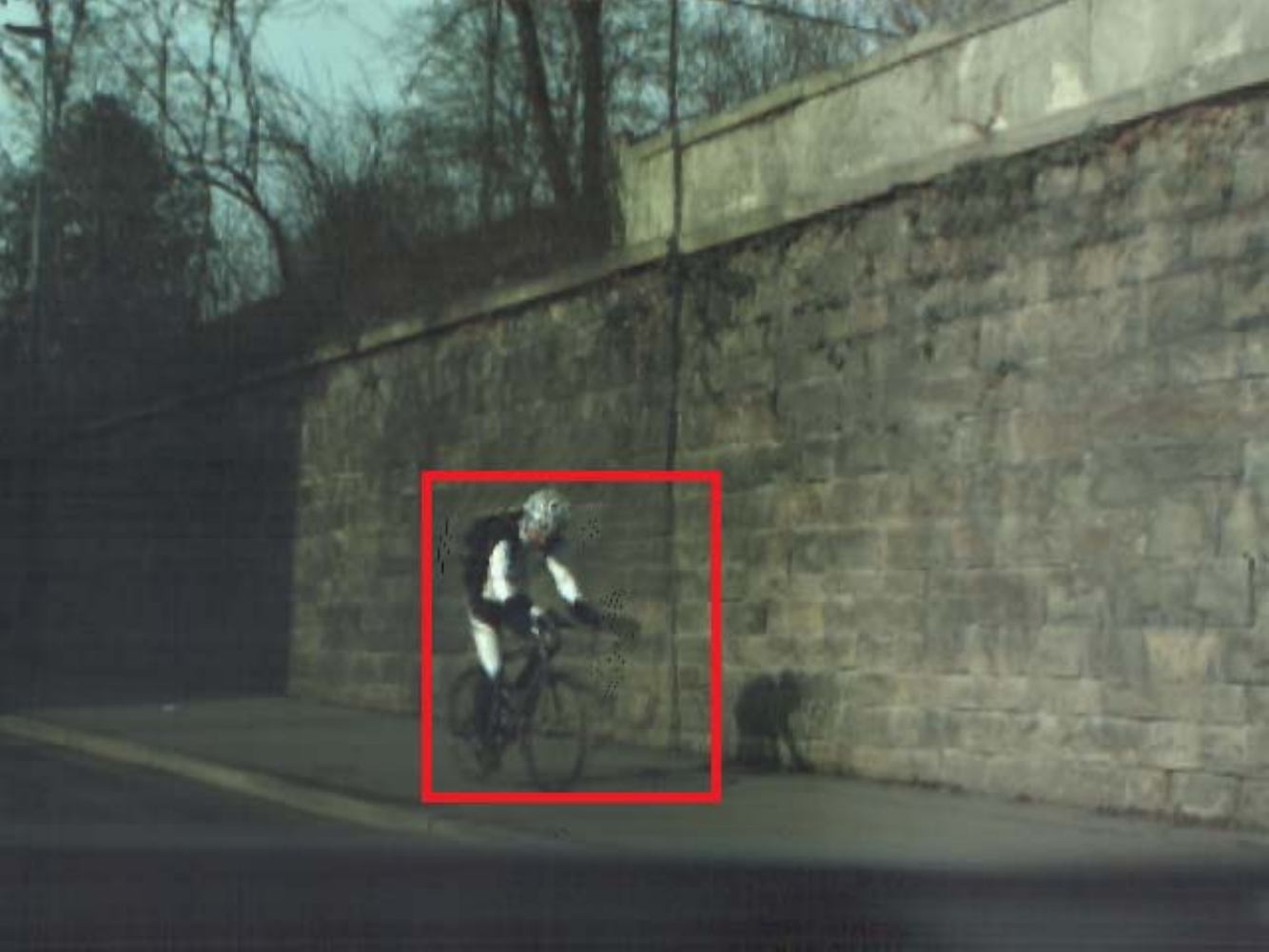}
		&
		\includegraphics[width=0.78\itemwidth]{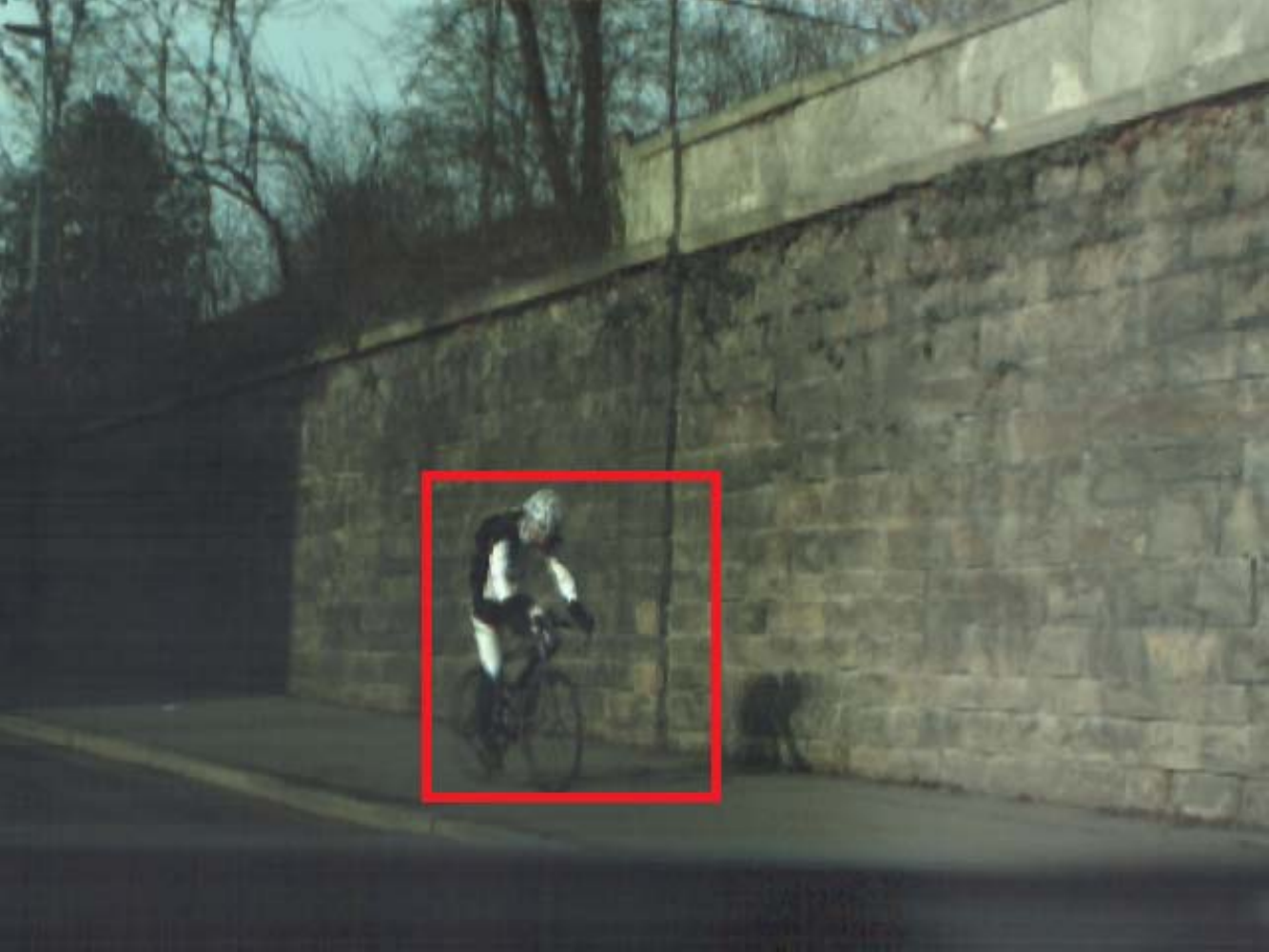}
		&
		\includegraphics[width=0.78\itemwidth]{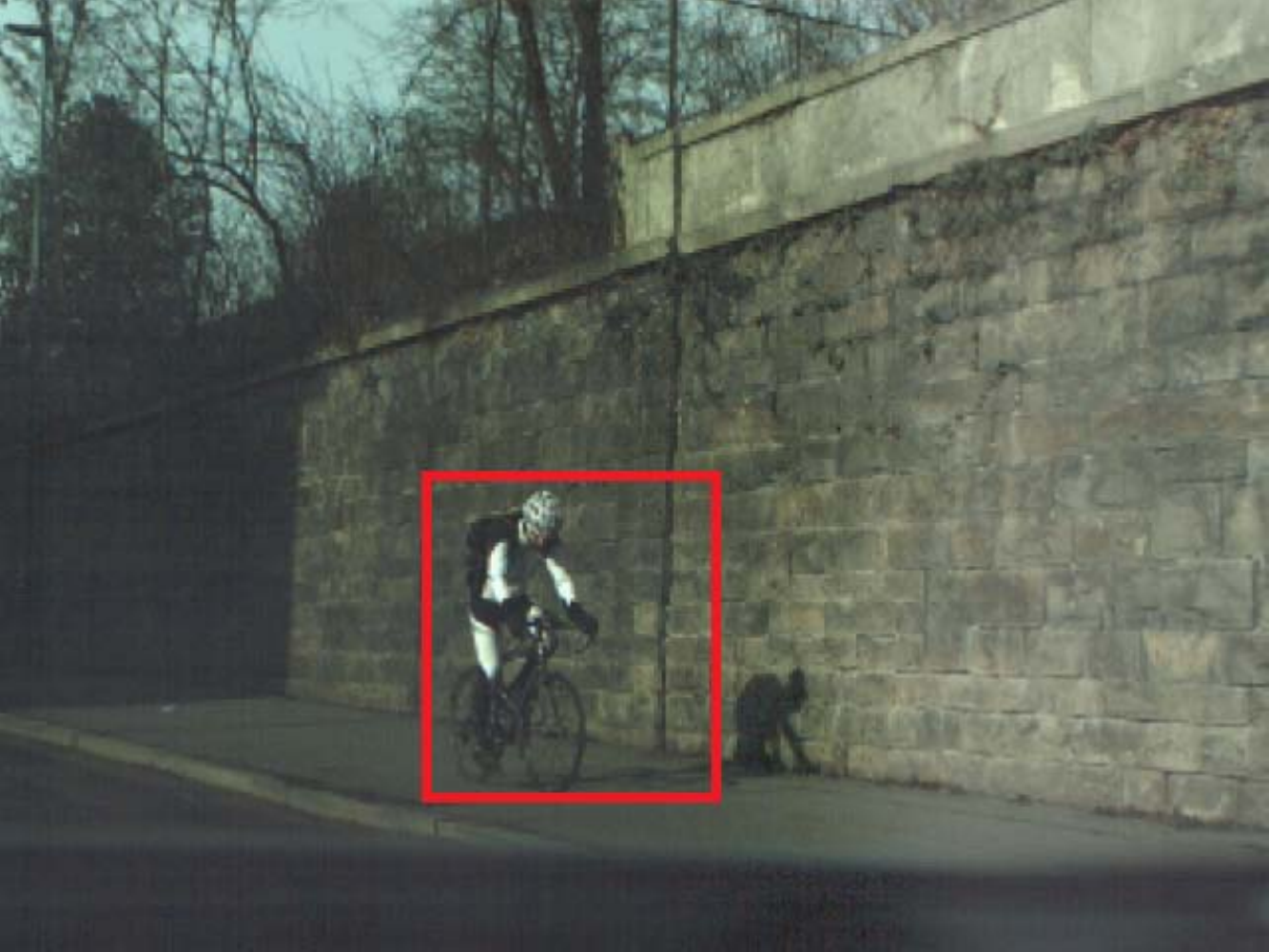}
		\vspace{-0.13cm} \\
		\scriptsize LBMF
		&
		\scriptsize RAFT-based
		&
		\scriptsize w/o ${\mathbf O}$ 
		&
		\scriptsize w/o $\Delta{\mathbf U}$
		&
		\scriptsize CVR (Ours)
		&
		\scriptsize Ground-truth
		\vspace{-0.017cm} \\
		\includegraphics[width=0.52\itemwidth]{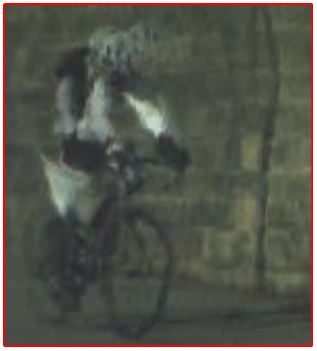}
		&
		\includegraphics[width=0.52\itemwidth]{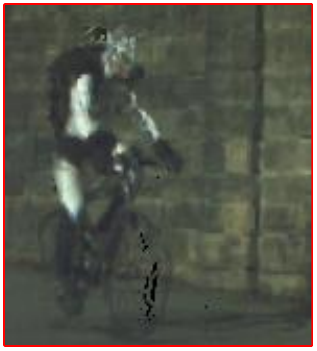}
		&
		\includegraphics[width=0.52\itemwidth]{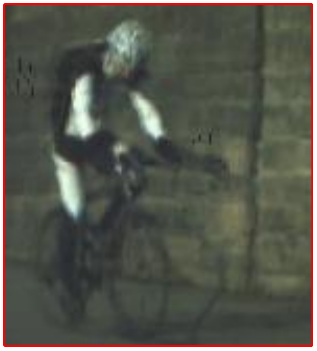}
		&
		\includegraphics[width=0.52\itemwidth]{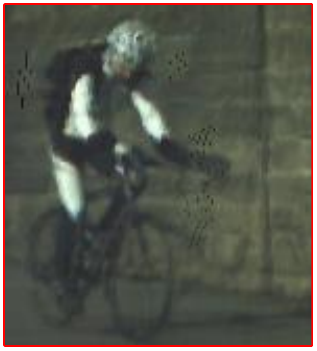}
		&
		\includegraphics[width=0.52\itemwidth]{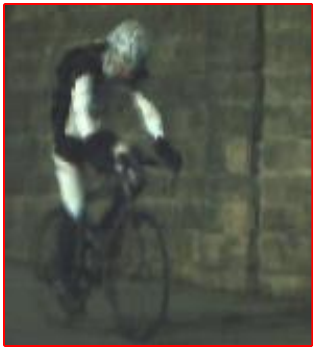}
		&
		\includegraphics[width=0.52\itemwidth]{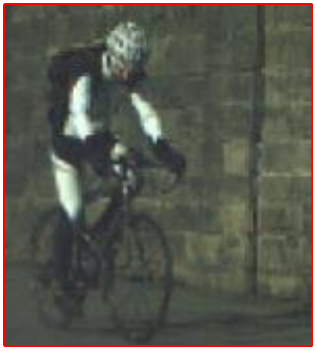}
		\vspace{-0.13cm} \\
		\scriptsize LBMF (Crop)
		&
		\scriptsize RAFT-based (Crop)
		&
		\scriptsize w/o ${\mathbf O}$ (Crop)
		&
		\scriptsize w/o $\Delta{\mathbf U}$ (Crop)
		&
		\scriptsize CVR (Crop)
		&
		\scriptsize Ground-truth (Crop)
		\\
	\end{tabular}\vspace{-0.33cm}
	\caption{Visual results of ablation study. Our context-aware method is also adaptable to motion artifacts specific to moving objects.}
	\label{fig:ablation_coms}
	\vspace{-5.00mm}
\end{figure*}

Note that our method is able to produce a continuous GS sequence, which is far beyond \cite{liu2020deep,fan2021sunet,zhong2021rscd}, although \cite{fan2021sunet} can decode the plausible details of the GS image at a specific time.
Traditional methods \cite{zhuang2017rolling,zhuang2020homography} cannot estimate the underlying RS geometry robustly and accurately, resulting in ghosting artifacts. They are also computationally inefficient due to the complicated handling. 
Due to inherent flaws in the network architectures, the VFI methods \cite{bao2019depth,park2020bmbc} fail to remove the RS effect.
An intuitive cascade of RS correction and VFI methods tends to accumulate errors and is prone to blurring artifacts and local inaccuracies. Such cascades also have large models and thus be relatively time-consuming.
In contrast, our end-to-end pipeline performs favorably against the SOTA methods in terms of both RS correction and inference efficiency.
Note also that obnoxious black holes and object-specific motion artifacts appear in \cite{fan2021rssr}, degrading the visual experience, as outlined in Sec.~\ref{sec:Introduction}.
In general, our \emph{CVR} improves RSSR and therefore recovers higher realism results, and our \emph{CVR*} also develops a new concise and efficient framework for related tasks.

\vspace{-0.4mm}
\subsection{GS Video Reconstruction Results}
\vspace{-0.3mm}
We apply our method to generate multiple in-between GS frames at arbitrary time $t \in [0, 1]$. 
The visual results for $5 \times$ temporal upsampling are shown in Fig.~\ref{fig:multiple_frames}.
More results are provided in Appendix~\ref{sec:video_demo}.
Our method can not only successfully remove the RS effect, but also can robustly reconstruct smooth and continuous GS videos.

\begin{table}[!t]
	\footnotesize
	\caption{Ablation results for \emph{CVR} architecture on $\mathcal{M}$, $\mathcal{G}$ and $\mathcal{L}$.} \label{tab:ablation}
	\vspace{-2.5mm}
	\centering
	\begin{tabular}{lcccccc}
		\hline
		\multirow{2}{*}{Settings} & \multicolumn{3}{c}{PSNR$\uparrow$ (dB)}                      &   & \multicolumn{2}{c}{SSIM$\uparrow$}      \\ \cline{2-4} \cline{6-7}
		& CRM       & CR         & FR        & & CR       & FR     \\ \hline
		LBMF                    & 26.10     & 25.97      & 25.78     & & 0.806     & 0.771     \\ 
		RAFT-based              & 30.50     & 29.89      & 27.99     & & 0.917     & 0.840     \\ 
		Freeze $\mathcal{M}$    & 31.94     & 31.65      & 28.11     & & 0.928     & 0.837     \\ \hline
		${\mathbf T}\cdot\Delta{\mathbf U}$ & 32.00     & 31.63      & 28.56     & & \textbf{0.929}     & 0.845  \\
		w/o $\Delta{\mathbf U}$ & 31.90     & 31.65      & 28.32     & & 0.928     & 0.841     \\ 
		w/o ${\mathbf O}$       & 28.22     & 26.31      & 24.04     & & 0.902     & 0.813     \\ \hline
		w/o $\mathcal{L}_r$     & 31.80     & 31.53      & 28.31     & & 0.927     & 0.840     \\
		w/o $\mathcal{L}_p$     & 31.60     & 31.34      & 28.49     & & \textbf{0.929}     & 0.842     \\
		w/o $\mathcal{L}_c$     & 31.88     & 31.64      & 28.44     & & 0.928     & 0.842     \\
		w/o $\mathcal{L}_{tv}$  & 31.93     & 31.71      & 28.45     & & 0.928     & 0.844     \\ 
 		Add $\mathcal{L}'_c$    & 31.97     & 31.58      & 28.55     & & \textbf{0.929}     & 0.844    \\ \hline
		full model & \textbf{32.02} & \textbf{31.74} & \textbf{28.72} & & \textbf{0.929} & \textbf{0.847} \\ \hline
	\end{tabular}
	\vspace{-6.4mm}
\end{table}

\vspace{-0.4mm}
\subsection{Ablation Studies} \label{sec:Ablations}
\vspace{-0.3mm}
\noindent\textbf{Ablation on motion interpretation module $\mathcal{M}$.}
We first replace NBMF and ABMF with linear BMF (\ie LBMF), which is a widely used BMF initialization scheme in popular VFI methods, \eg \cite{park2020bmbc,jiang2018super,niklaus2018context,niklaus2020softmax,siyao2021deep}.
Then, we replace PWC-Net with the SOTA optical flow estimation pipeline RAFT \cite{teed2020raft}. 
Finally, we freeze $\mathcal{M}$ and solely train $\mathcal{G}$ in the training phase.
As can be seen from Table~\ref{tab:ablation} and Fig.~\ref{fig:ablation_coms}, LBMF is extremely ineffective for the RS-based video construction task, which reveals the superiority of our proposed NBMF as well as ABMF. This could facilitate further research in related fields, especially the simpler ABMF.
Since the RAFT-based full baseline is not easily optimized jointly end-to-end, it is prone to unsmoothness at local motion boundaries.
Additionally, training the entire network together with $\mathcal{M}$ can improve model performance.

\vspace{0.25mm}
\noindent\textbf{Ablation on GS frame synthesis module $\mathcal{G}$.}
We analyze the role of each component of $\mathcal{G}$ in Table~\ref{tab:ablation}, including \textbf{1}) multiplying $\Delta{\mathbf U}$ by a normalized scanline offset ${\mathbf T}$ to explicitly model its scanline dependence like \cite{liu2020deep,fan2021rssr,zhong2022bringing}, and \textbf{2}) removing MEL (\ie w/o $\Delta{\mathbf U}$) and CAL (\ie w/o ${\mathbf O}$), separately.
Combined with Fig.~\ref{fig:ablation_coms}, one can observe that they both lead to performance degradation, especially removing CAL, which causes aliasing effects during context aggregation, \eg misaligned wheels and black edges. Moreover, removing MEL will reduce the adaptability of our method to object-specific motion artifacts, especially for the more challenging Fastec-RS dataset. In summary, our method can adaptively infer occlusions and enhance motion boundaries.

 \vspace{0.25mm}
 \noindent\textbf{Ablation on loss function $\mathcal{L}$.}
 We remove the loss terms one by one to analyze their respective roles.
 We also directly use the $\mathcal{L}_1$ difference between the two refined intermediate GS frame candidates $\hat{\mathbf I}_{0 \to t}^g$ and $\hat{\mathbf I}_{1 \to t}^g$ to measure the context consistency, which constitutes a self-supervised loss term $\mathcal{L}'_c$. Adding $\mathcal{L}'_c$ to $\mathcal{L}_c$ creates a loop loss.
 In Table~\ref{tab:ablation}, using $\mathcal{L}'_c$ does not lead to performance gains as supervising the final GS prediction (\eg by $\mathcal{L}_r$) has significant dominance in context alignment. Overall, our loss function $\mathcal{L}$ is effective because it performs best when all loss terms are used.

\vspace{-1.2mm}
\subsection{Limitation and Discussion} \label{sec:Limitation}
\vspace{-1.2mm}
Our method relies on optical flow estimation, so there may be aliasing artifacts in areas such as low/weak textures.
Besides, although we have assumed that the pixels of the target GS image at time $t \in [0,1]$ are visible in one of the RS images, some of them at the edges of the GS image may not be available, \eg the lower right corner of GS images at $t=0$ in Figs.~\ref{fig:teaser_img}~and~\ref{fig:multiple_frames}, due to severe occlusions from fast camera motion or object motion.
Future use of more frames may be able to fill in these possible invisible regions.

\vspace{-1.2mm}
\section{Conclusion} \label{sec:Conclusion}
\vspace{-1.4mm}
In this paper, we have presented a context-aware architecture \emph{CVR} for end-to-end video reconstruction of RS cameras, which incorporates temporal smoothness to recover high-fidelity GS video frames with fewer artifacts and better details. Moreover, we have developed a simple yet efficient pipeline \emph{CVR*} based on the proposed ABMF model which works robustly with RS cameras. Our proposed framework exploits the spatio-temporal coherence embedded in the latent GS video via motion interpretation and occlusion reasoning, significantly outperforming the SOTA methods. We hope this study can shed light for future research on video frame reconstruction of RS cameras.


{\small
\bibliographystyle{ieee_fullname}
\bibliography{References}
}


\newcommand{\beginsupplement}{%
	\setcounter{table}{0}
	\renewcommand{\thetable}{A\arabic{table}}%
	\setcounter{figure}{0}
	\renewcommand{\thefigure}{A\arabic{figure}}%
}
\beginsupplement

\clearpage

\begin{appendices}

\emph{
In this appendix, we first derive a general parametric form of bilateral motion field (BMF) by considering the readout time ratio, and then justify our problem setup.
Next, we provide thorough analyses to our ABMF model, occlusion reasoning, and motion enhancement. 
Afterward, we show additional experimental results on RS correction, intermediate flow, generalization, and GS video recovery, which fully demonstrates the superiority of our pipeline.
We also include a video demo to present dynamic GS video reconstruction results.
Additional details of the loss function are then added.
Furthermore, we report a partial ablation study of CVR*.
At last, several failure cases are given to look forward to possible future research.
}

\section{Instructions on Problem Setup} \label{sec:instructions_setup}
\vspace{-0.8mm}
In this section, we show a detailed derivation of the general parameterization of BMF in the time dimension, followed by an explanation of our problem setup.

\vspace{-0.5mm}
\subsection{General formulation of BMF} \label{sec:general_BMF}
\vspace{-0.4mm}
We first give a brief description of the connection between the motion field ${\mathbf U}_{0 \to s}$ and the optical flow ${\mathbf F}_{0 \to 1}$ by accounting for the first RS frame $\mathbf{I}^{r}_0$ as an example.
Since this does not contain our contribution, we only give the necessary details to follow the derivation below. More details of this connection can be found in \cite{fan2021rssr}.
Suppose that to estimate ${\mathbf U}_{0 \to s}$ that warps each pixel (\eg ${\mathbf x}$ in scanline $\kappa$) of $\mathbf{I}^{r}_0$ to the GS counterpart corresponding to its scanline $s$, this connection under the constant velocity motion model can be formulated as:
\begin{equation}\label{eq:11}
{\mathbf U}_{0 \to s}(\mathbf{x}) = {\mathbf C}_{0 \to s}(\mathbf{x}) \cdot {\mathbf F}_{0 \to 1}(\mathbf{x}),
\end{equation}
where
\begin{equation}\label{eq:12}
{\mathbf C}_{0 \to s}(\mathbf{x}) = \frac{\gamma (s-\kappa) (h-\gamma\pi_v)}{h^2},
\end{equation}
denotes the forward correction map. Here, $\gamma\in(0,1]$ is the readout time ratio \cite{zhuang2017rolling}, $h$ is the number of scanlines, and $\pi_v$ represents the latent inter-GS-frame vertical optical flow.

Next, we extend it to the time domain and derive a more general formulation than that in the main paper, \ie $\gamma$ will be taken into account. 
According to the definition in the main paper, the one-to-one correspondence between time $t$ and scanline $s$ will satisfy
\begin{equation}\label{eq:13}
t = \frac{\gamma}{h} \left(s-\frac{h}{2}\right).
\end{equation}
It is easy to verify that the central scanlines of $\mathbf{I}^{r}_0$ and $\mathbf{I}^{r}_1$ correspond to time instances 0 and 1, respectively. Note that the first scanline of $\mathbf{I}^{r}_0$ will coincide with time $-\frac{\gamma}{2}$.

Assume that $t\in[0,1]$ corresponds to the scanline $s$ to be restored, and $\tau_0\in[-\frac{\gamma}{2},\frac{\gamma}{2}]$ corresponds to the exposure time of scanline $\kappa$ of $\mathbf{I}^{r}_0$, we can obtain $t-\tau_0 \triangleq \frac{\gamma (s-\kappa)}{h}$.
Further combining with Eq.~\eqref{eq:12} yields 
\begin{equation}\label{eq:14}
{\mathbf C}_{0 \to t}(\mathbf{x}) = \frac{(t-\tau_0) (h-\gamma\pi_v)}{h}.
\end{equation}
Similarly, the backward correction map that accounts for the second RS frame $\mathbf{I}^{r}_1$ with $\tau_1\in[1-\frac{\gamma}{2},1+\frac{\gamma}{2}]$ can be defined as:
\begin{equation}\label{eq:15}
{\mathbf C}_{1 \to t}(\mathbf{x}) = \frac{(\tau_1-t) (h+\gamma\pi'_v)}{h}.
\end{equation}

Note that Eqs.~\eqref{eq:14} and \eqref{eq:15} model the bilateral correction map through the time paradigm in a general sense. In this way, the general parametric form of BMF is modeled. This generality is reflected by the fact that, unlike Eq.~(5) in the main paper, we get $\tau_0\in[-\frac{\gamma}{2},\frac{\gamma}{2}]$ and $\tau_1\in[1-\frac{\gamma}{2},1+\frac{\gamma}{2}]$ instead of $\tau_0\in[-0.5,0.5]$ and $\tau_1\in[0.5,1.5]$.
This is because we assume $\gamma=1$ in our problem setup. In the following, we will explain the feasibility of this setup.

\subsection{Feasibility analysis of our problem setup} \label{sec:feasibility_setup}
Our problem setup with $\gamma=1$ is based on three main reasons. 
Firstly, the Carla-RS and Fastec-RS datasets proposed by \cite{liu2020deep} are the only available RS correction benchmark datasets. And they are constructed by using the assumption of $\gamma=1$. Since we leverage these two datasets to train our network, this setup can also be considered as we use $\gamma=1$ in advance.
Secondly, as manifested by \cite{liu2020deep,fan2021sunet,fan2021rssr} and our experiments, the trained deep learning model based on these two datasets can be successfully generalized to RS data acquired by real cameras (\ie $\gamma$ may not be equal to 1). Also, $\gamma$ is directly set to 1 in \cite{zhuang2020homography,fan2021rsstereo} to correct real RS images, which can avoid the non-trivial readout calibration. These studies demonstrate that assuming $\gamma=1$ is generally feasible for modeling RS correction problems. 
Thirdly, this setup can facilitate temporally tractable frame interpolation between two consecutive RS images. Note that we derive the general parameterization of BMF in Appendix~\ref{sec:general_BMF}, which will help the future exploration of more general RS-based video reconstruction tasks (\eg with extremely small $\gamma$).

\section{Additional Architecture Analyses} \label{sec:architecture_analyses}
In this section, we provide more in-depth analyses of our method in terms of the proposed ABMF model, time-aware occlusion reasoning and motion enhancement architectures.

\subsection{Further analysis on ABMF model} \label{sec:ABMF_analyses}
\begin{figure}[!t]
	\centering
	\includegraphics[width=0.38\textwidth]{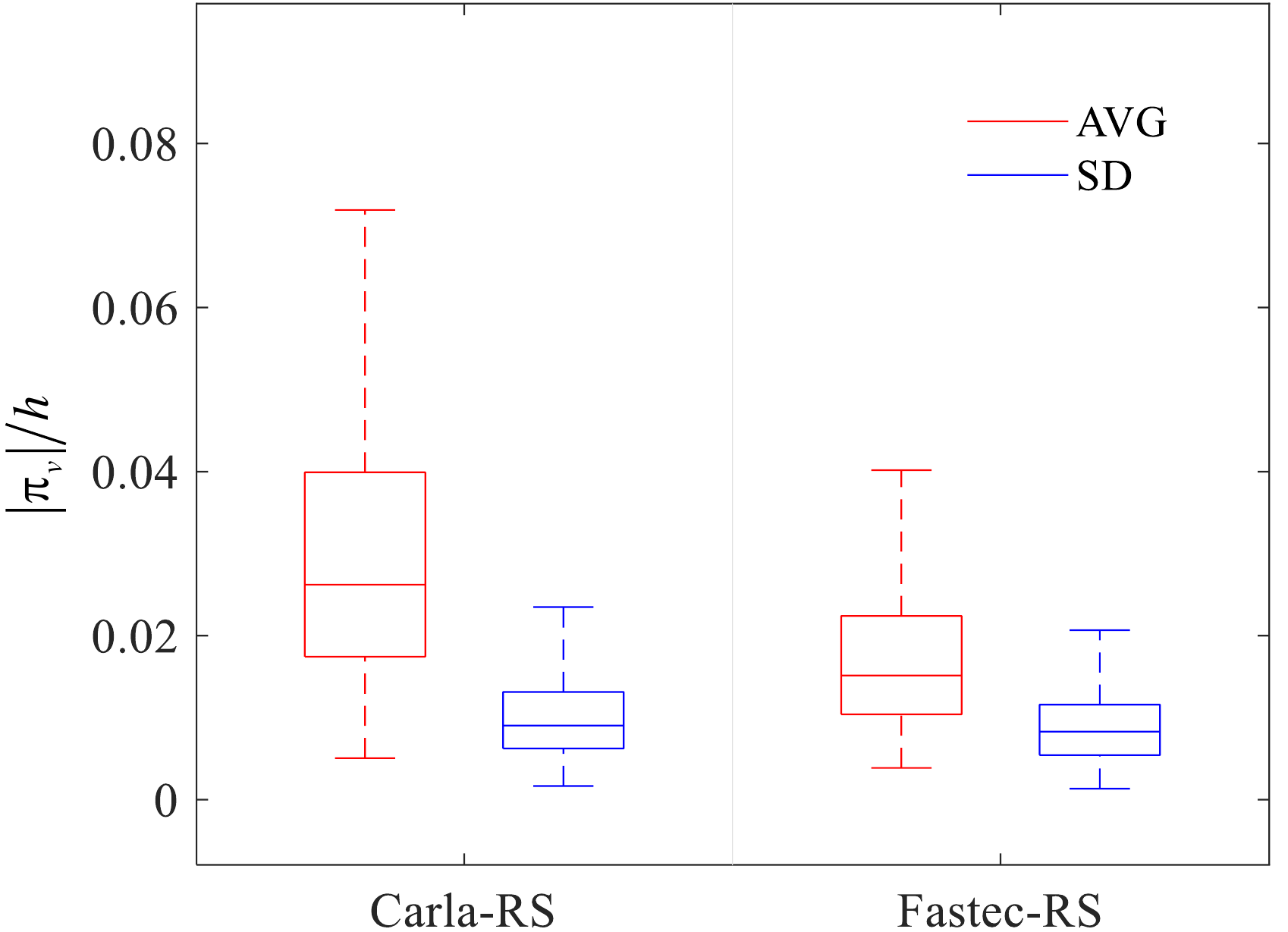} 
	\vspace{-0.7mm} 
	\caption{Statistical results of the vertical pixel displacement ratio $\frac{|\pi_v|} {h}$ under the Carla-RS and Fastec-RS datasets. Red is the average value (AVG) and blue is the standard deviation (SD). \label{fig:abmf_analysis}}
	\vspace{-4.7mm} 
\end{figure}

\begin{figure*}[!t]
	\centering
	\setlength{\tabcolsep}{0.010cm}
	\setlength{\itemwidth}{3.570cm}
	\hspace*{-\tabcolsep}\begin{tabular}{cccccc}
		\includegraphics[width=0.78\itemwidth]{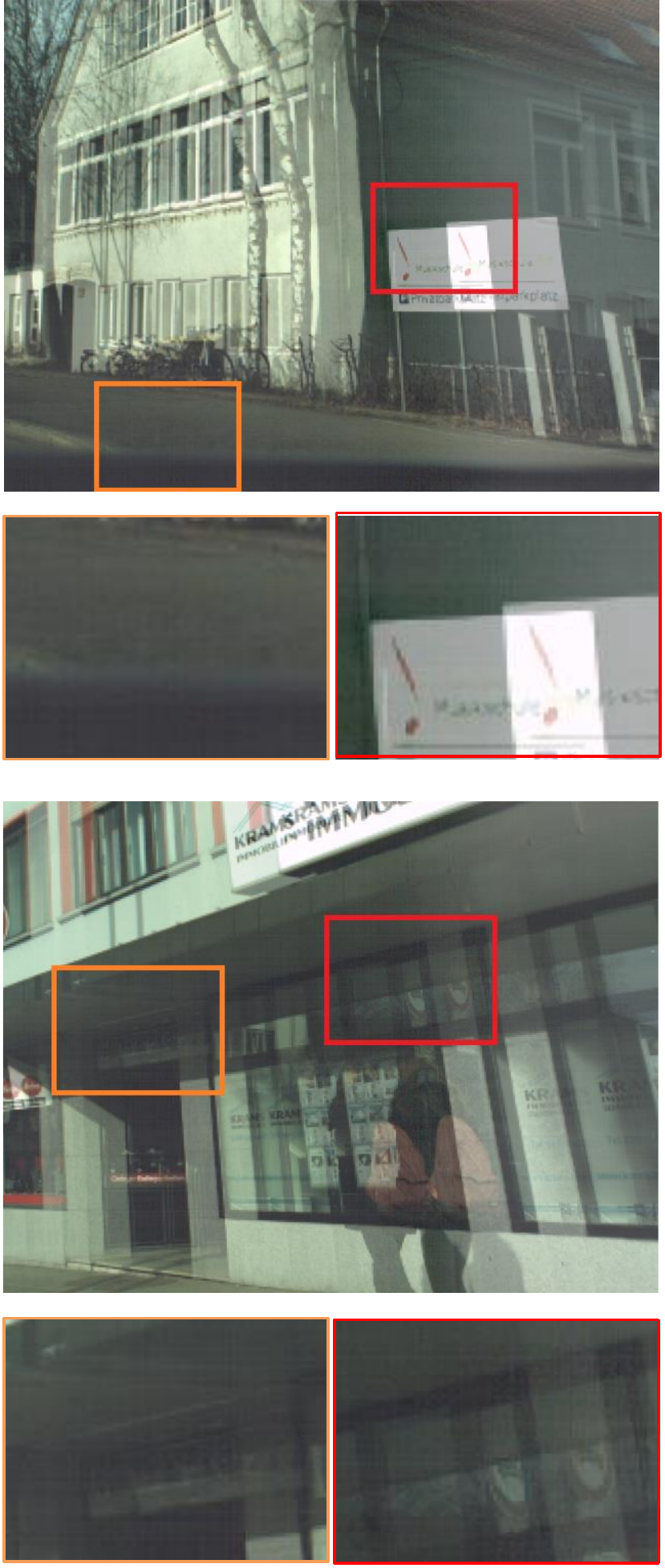}
		&
		\includegraphics[width=0.78\itemwidth]{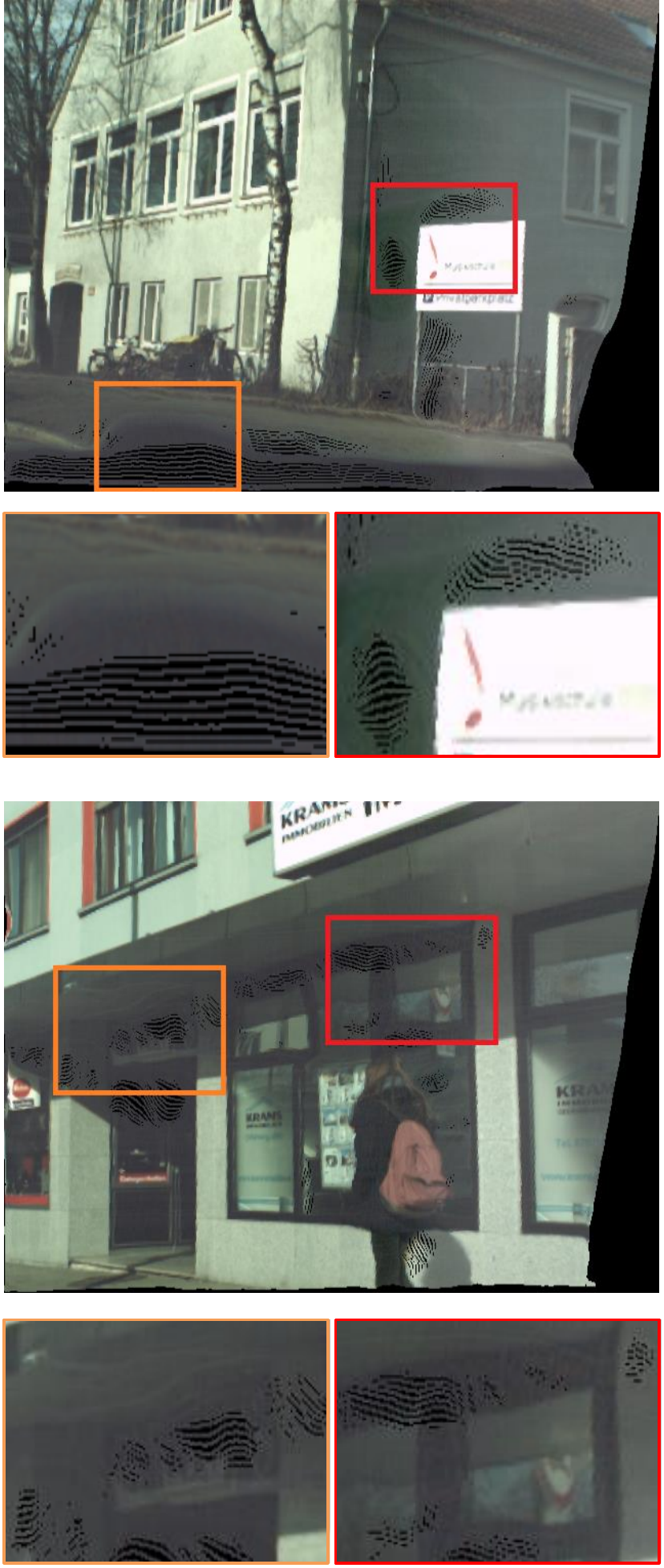}
		&
		\includegraphics[width=0.78\itemwidth]{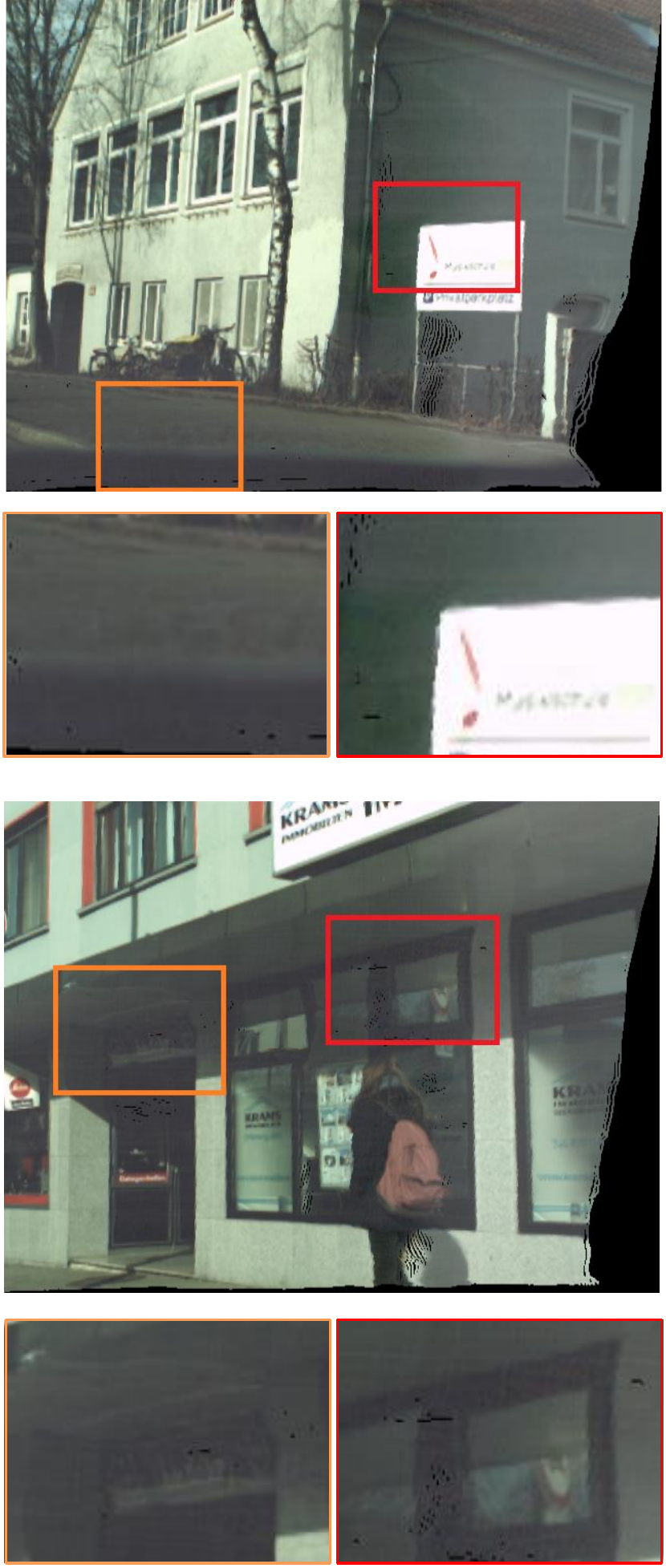}
		&
		\includegraphics[width=0.78\itemwidth]{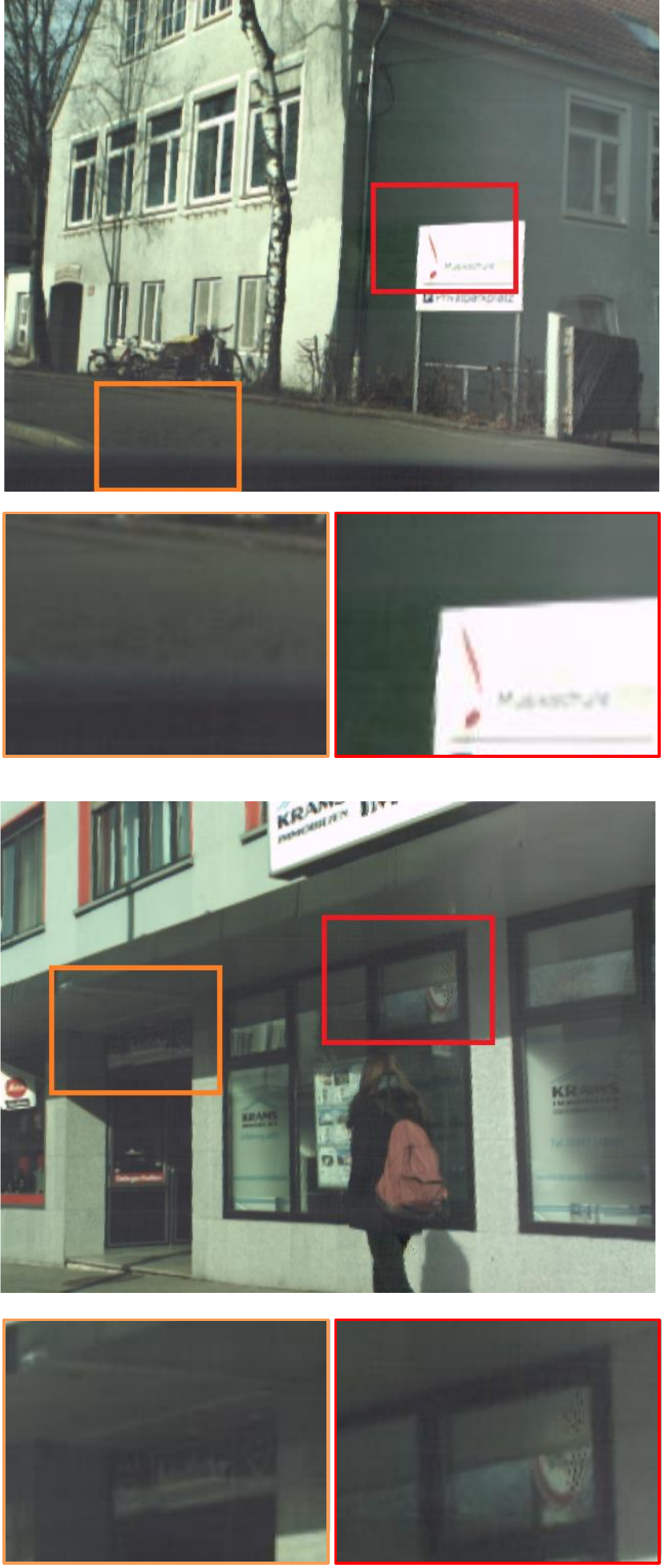}
		&
		\includegraphics[width=0.78\itemwidth]{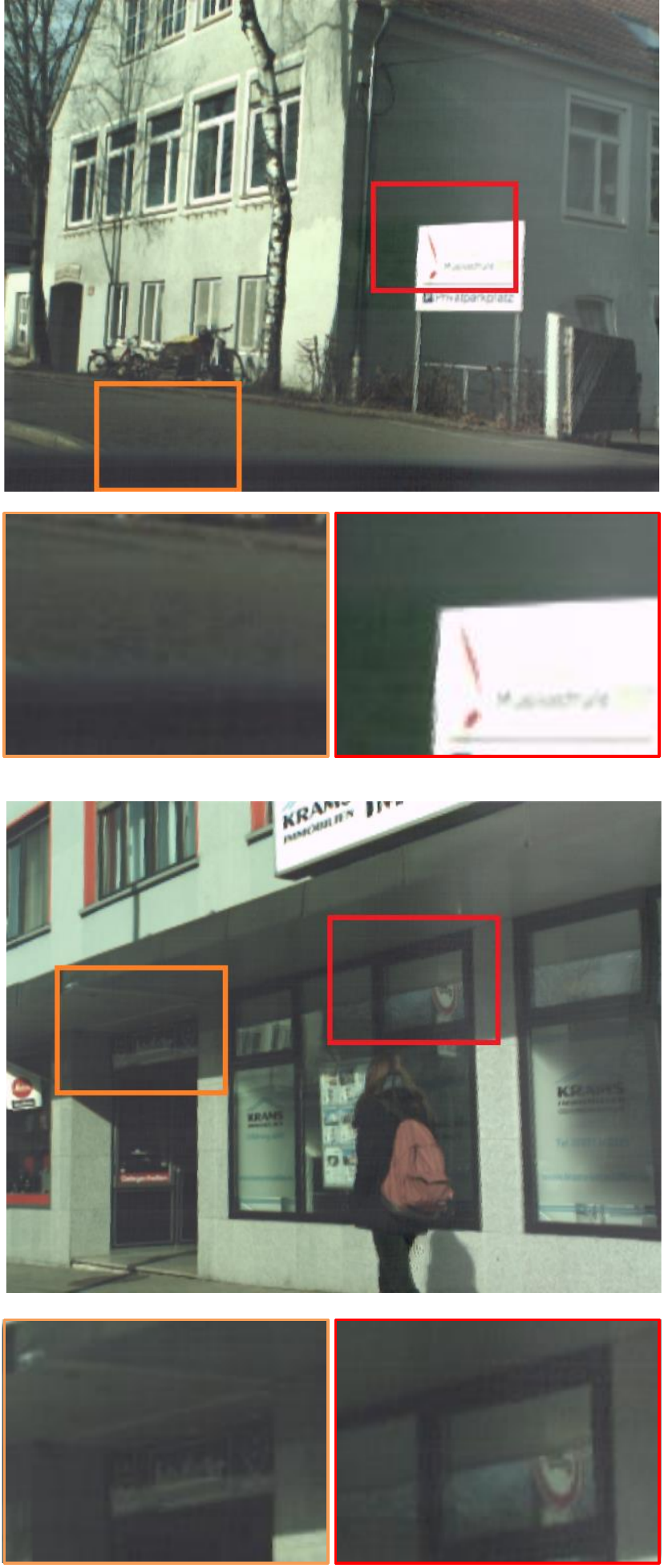}
		&
		\includegraphics[width=0.78\itemwidth]{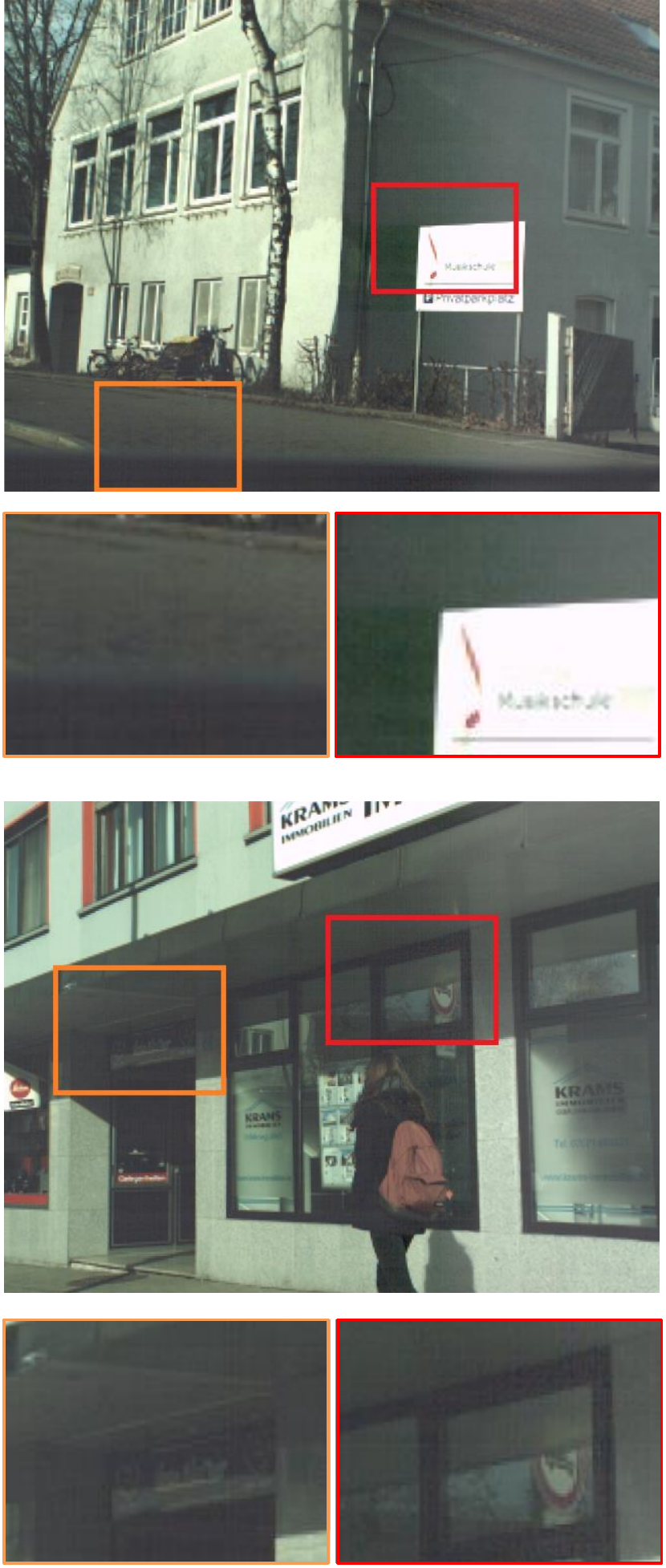}
		\vspace{-0.066cm} \\
		\scriptsize Input RS~(Overlayed)
		&
		\scriptsize RSSR*
		&
		\scriptsize RSSR~\cite{fan2021rssr}
		&
		\scriptsize CVR* (Ours)
		&
		\scriptsize CVR (Ours)
		&
		\scriptsize Ground-truth
		\\
	\end{tabular}
	\vspace{-0.38cm}
	\caption{Example results of the effectiveness of our ABMF model. Since the ABMF model ignores the depth variations, the ABMF-based RSSR* may encounter misaligned errors and unsmooth artifacts at motion boundaries, while the (NBMF-based) RSSR \cite{fan2021rssr} can alleviate these problems to some extent, but still not as well as it could be. Combined with the GS frame refinement module, ABMF provides concise and tractable benefits for GS video recovery, while NBMF can yield better initialization to generate higher fidelity GS images.}
	\label{fig:abmf_analysis_imgs}
	\vspace{-3.6mm}
\end{figure*}

Here, we show the rationality of our ABMF model proposed in Subsection 3.1 of the main paper, \ie, we need to verify that ${h\pm|\pi_v|} \approx h$ (equivalently, $\frac{|\pi_v|} {h} \approx 0$).
To this end, we define $\frac{|\pi_v|} {h}$ as the vertical pixel displacement ratio.
From \cite{fan2021rssr}, one can get $\frac{|\pi_v|} {h} = \left|{{\mathbf f}_v}/ {(h+{\mathbf f}_v)}\right|$, where ${\mathbf f}_v$ denotes the inter-RS-frame vertical optical flow and $h$ is the number of image rows.
We employ the state-of-the-art optical flow estimation pipeline RAFT \cite{teed2020raft} to obtain the optical flow map between two adjacent RS frames. Then, we calculate the average value and standard deviation of $\frac{|\pi_v|} {h}$ in each image of the test sets of Carla-RS and Fastec-RS datasets. We plot their respective statistics in Fig.~\ref{fig:abmf_analysis}. 
One can observe that the latent inter-GS-frame vertical optical flow value is usually much smaller than the number of image rows, \ie the proposed ABMF model is concise and reasonable. 
Note that $\frac{\gamma|\pi_v|} {h} \le \frac{|\pi_v|} {h}$, which indicates that our ABMF model is also valid under the general formulation of Appendix~\ref{sec:general_BMF}. 
Furthermore, as illustrated in Fig.~\ref{fig:abmf_analysis_imgs}, the ABMF-based RSSR* tends to have misaligned errors and unsmooth artifacts at local boundaries (\eg depth variation, slight blurring, monotonous texture, \etc) due to the isotropic approximation of ABMF. Fortunately, the experimental results demonstrate that our \emph{CVR*} (\ie combining ABMF with the GS frame refinement module) enhances local details and improves image quality in a coarse-to-fine manner, which can serve as an effective and efficient baseline for RS-based video reconstruction. 
Note that our \emph{CVR} can further improve the fidelity and authenticity of the recovered GS images because a better initial estimate is provided by using the network-based bilateral motion field (NBMF).

\begin{figure*}[htbp]
	\centering
	\includegraphics[width=0.96\textwidth]{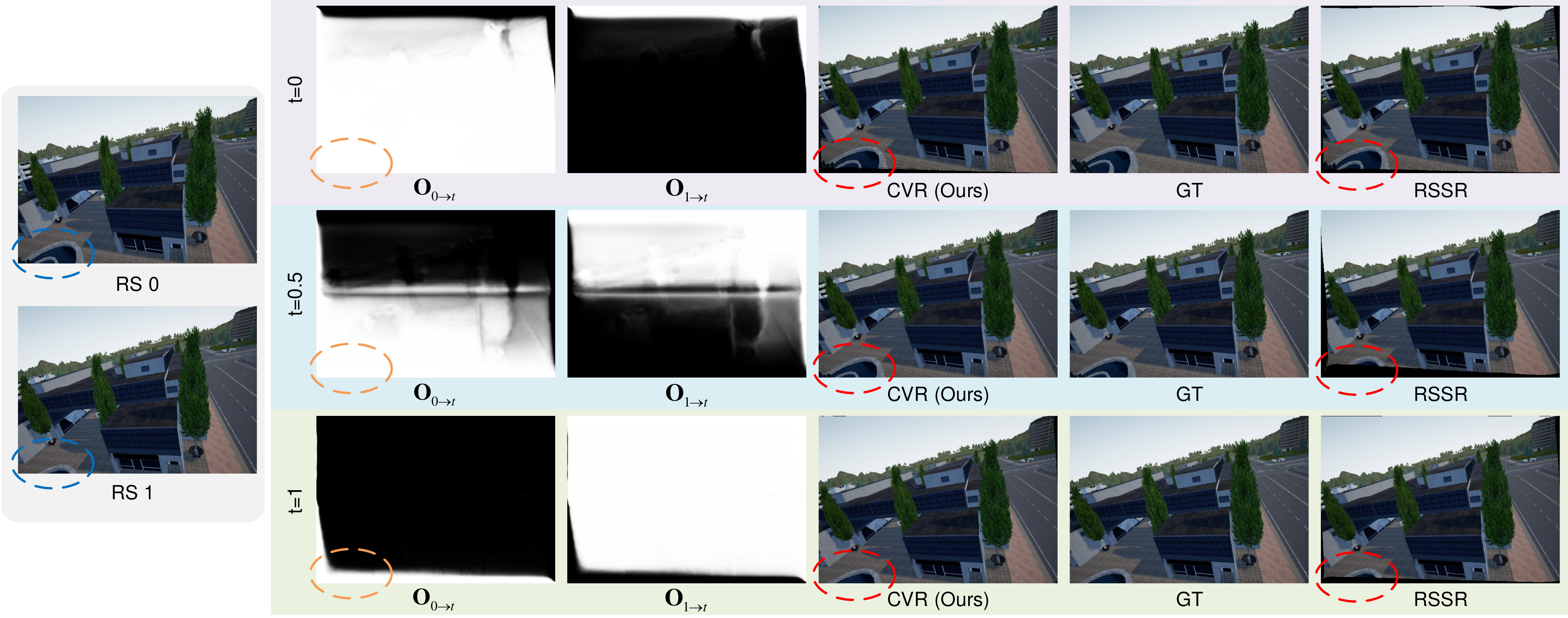} 
	\vspace{-3.2mm} 
	\caption{Example results of the effectiveness of our occlusion reasoning. We show GS frame recovery at times 0, 0.5 and 1, respectively. The brighter the color in the bilateral occlusion mask, the higher the credibility. Our method can adaptively and efficiently reason about complex occlusions and temporal abstractions, leading to visually more satisfactory GS reconstruction results than RSSR \cite{fan2021rssr}.\label{fig:occlusion_analysis}}
	\vspace{-3.4mm}
\end{figure*}

\begin{figure*}[htbp]
	\vspace{-1.0cm}
	\centering
	\setlength{\tabcolsep}{0.015cm}
	\setlength{\itemwidth}{3.543cm}
	\hspace*{-\tabcolsep}\begin{tabular}{cccccc}
		\includegraphics[width=0.78\itemwidth]{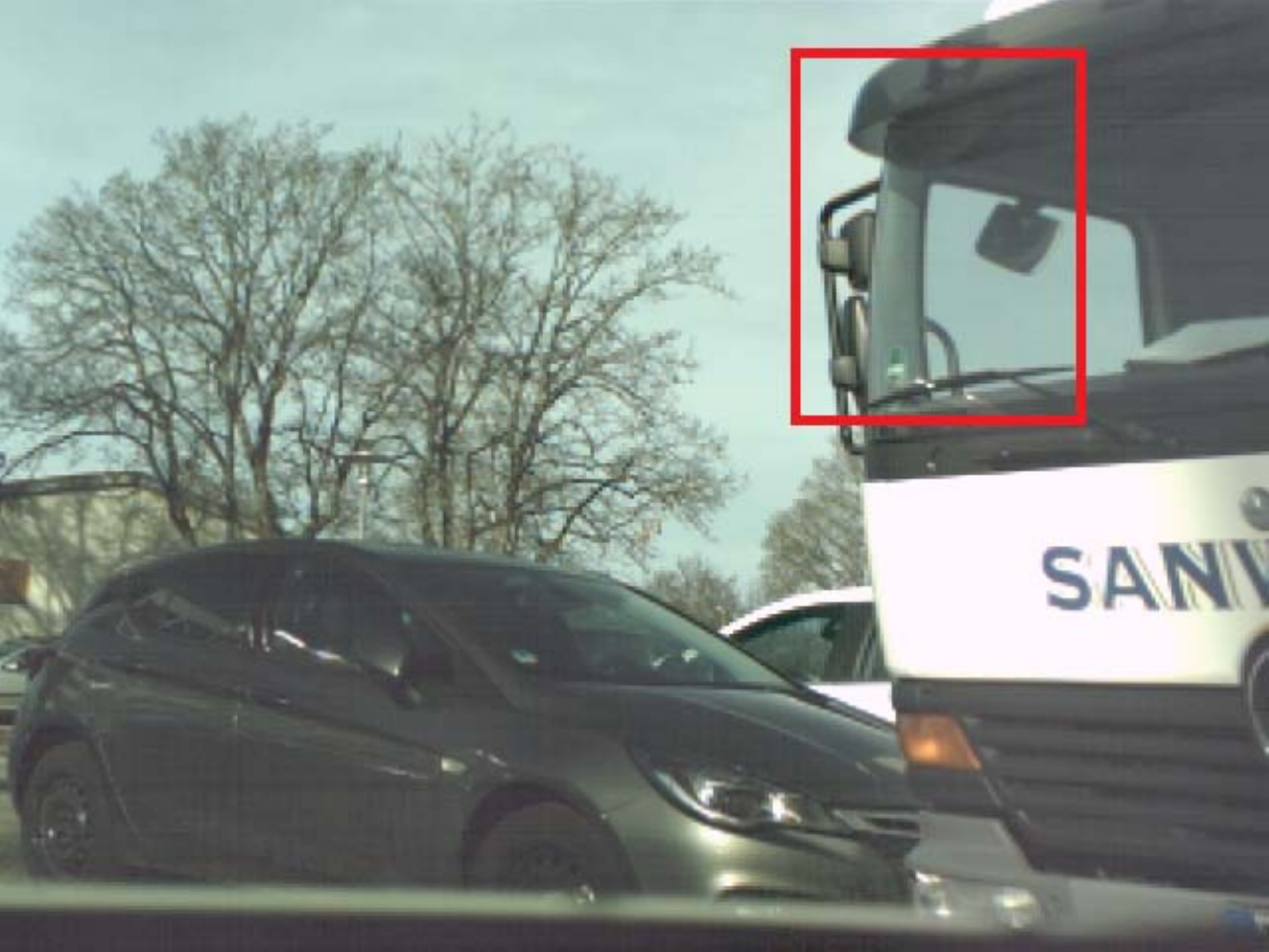}
		&
		\includegraphics[width=0.78\itemwidth]{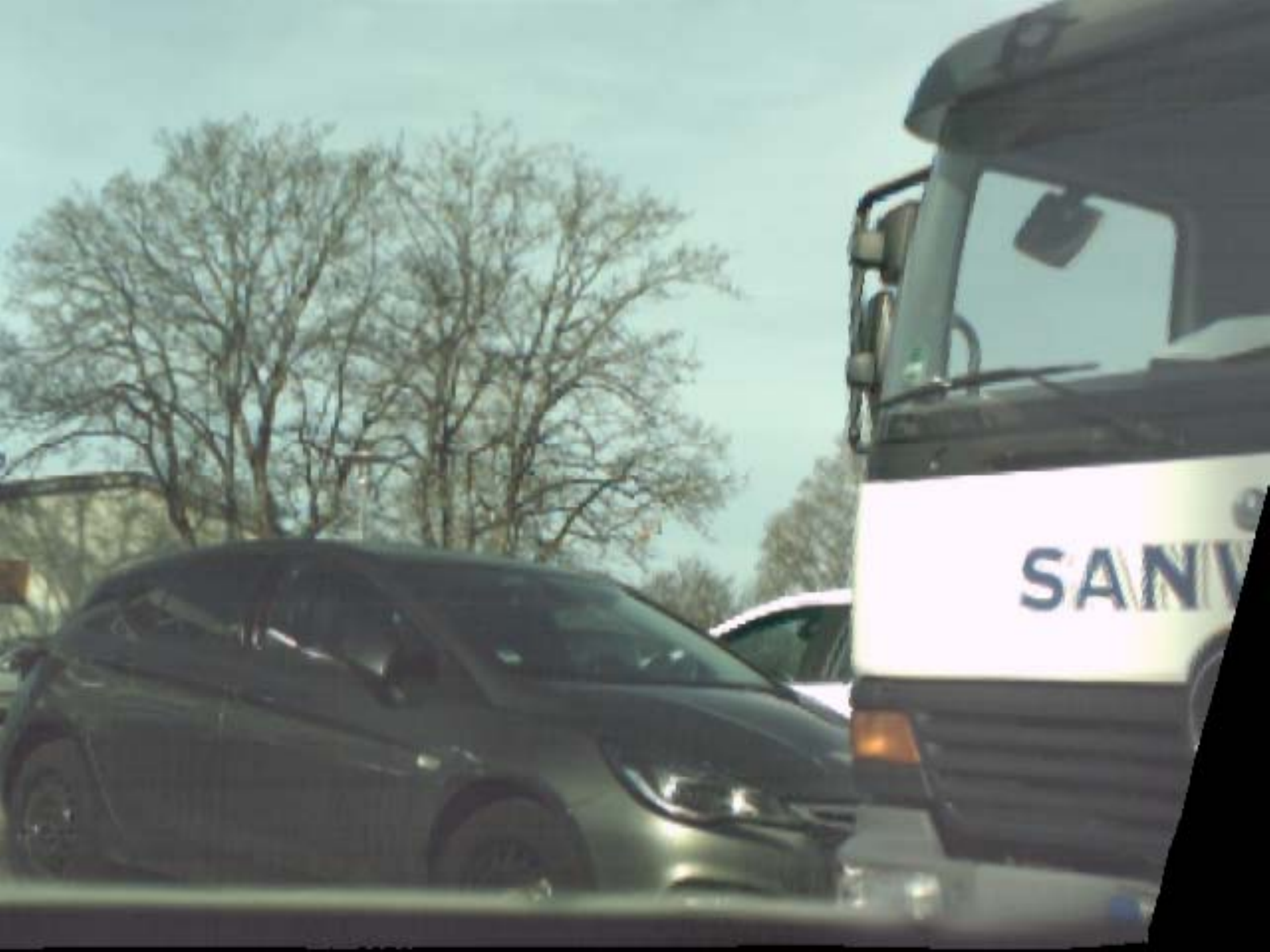}
		&
		\includegraphics[width=0.78\itemwidth]{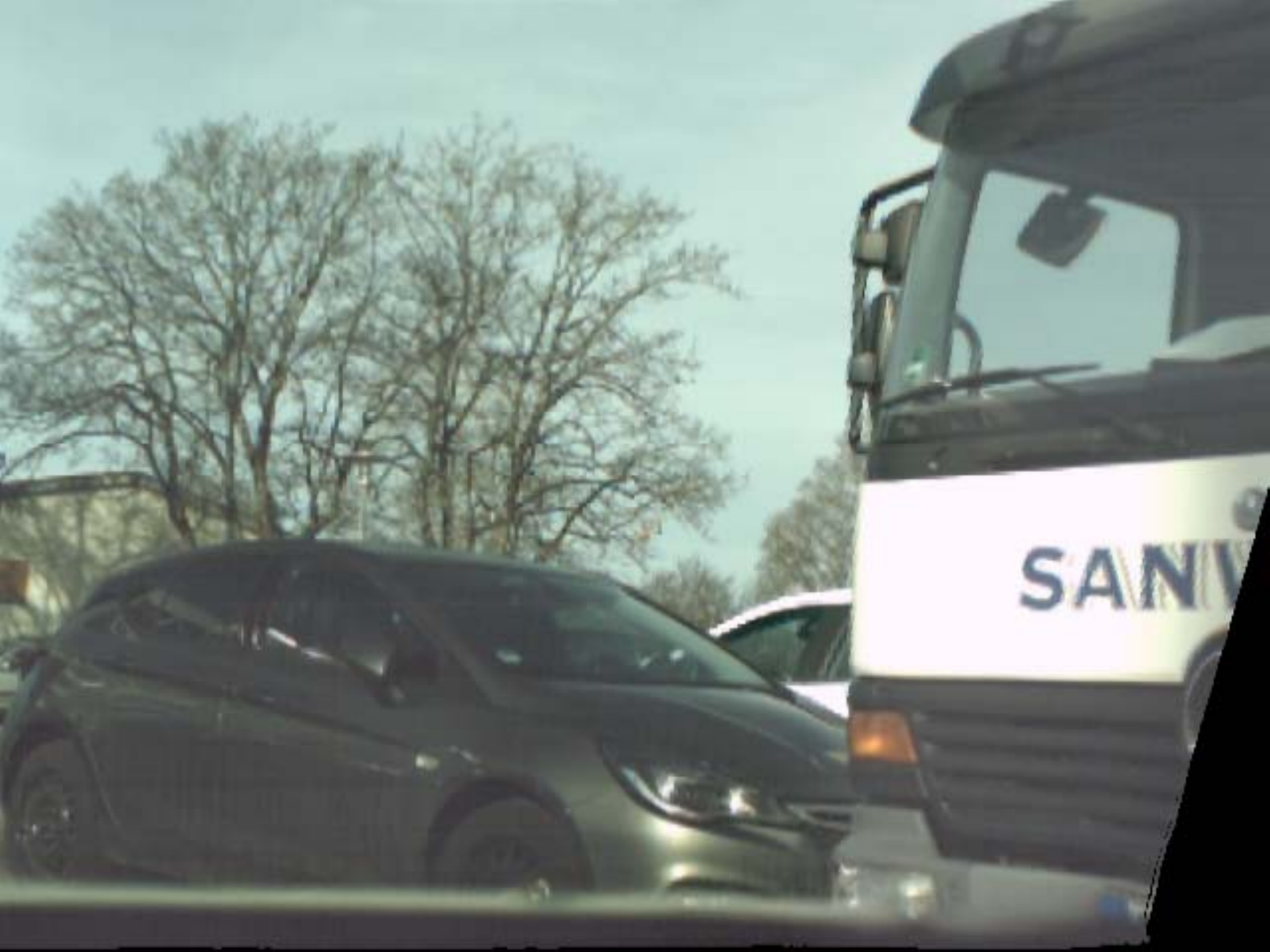}
		&
		\includegraphics[width=0.78\itemwidth]{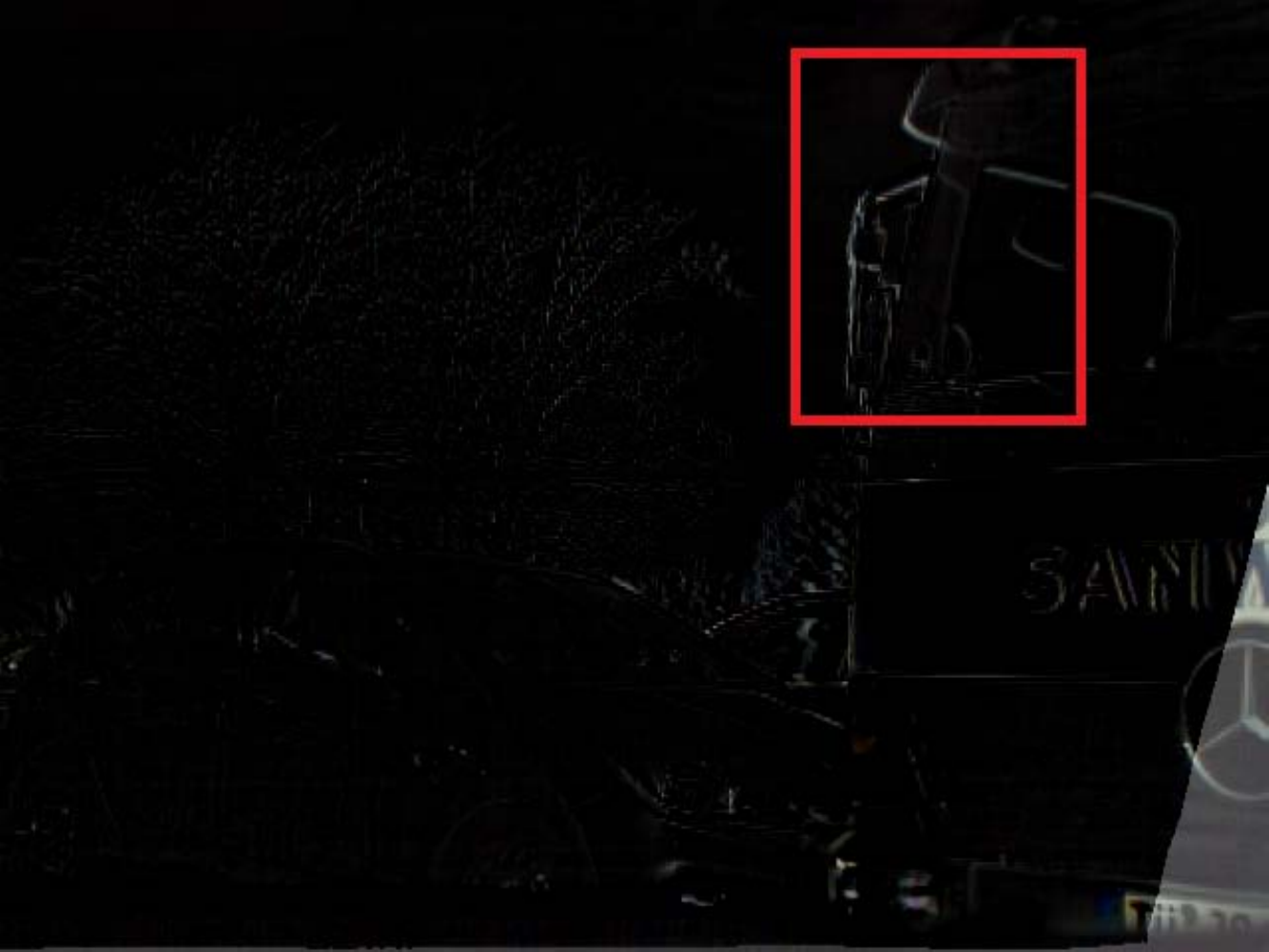}
		&
		\includegraphics[width=0.78\itemwidth]{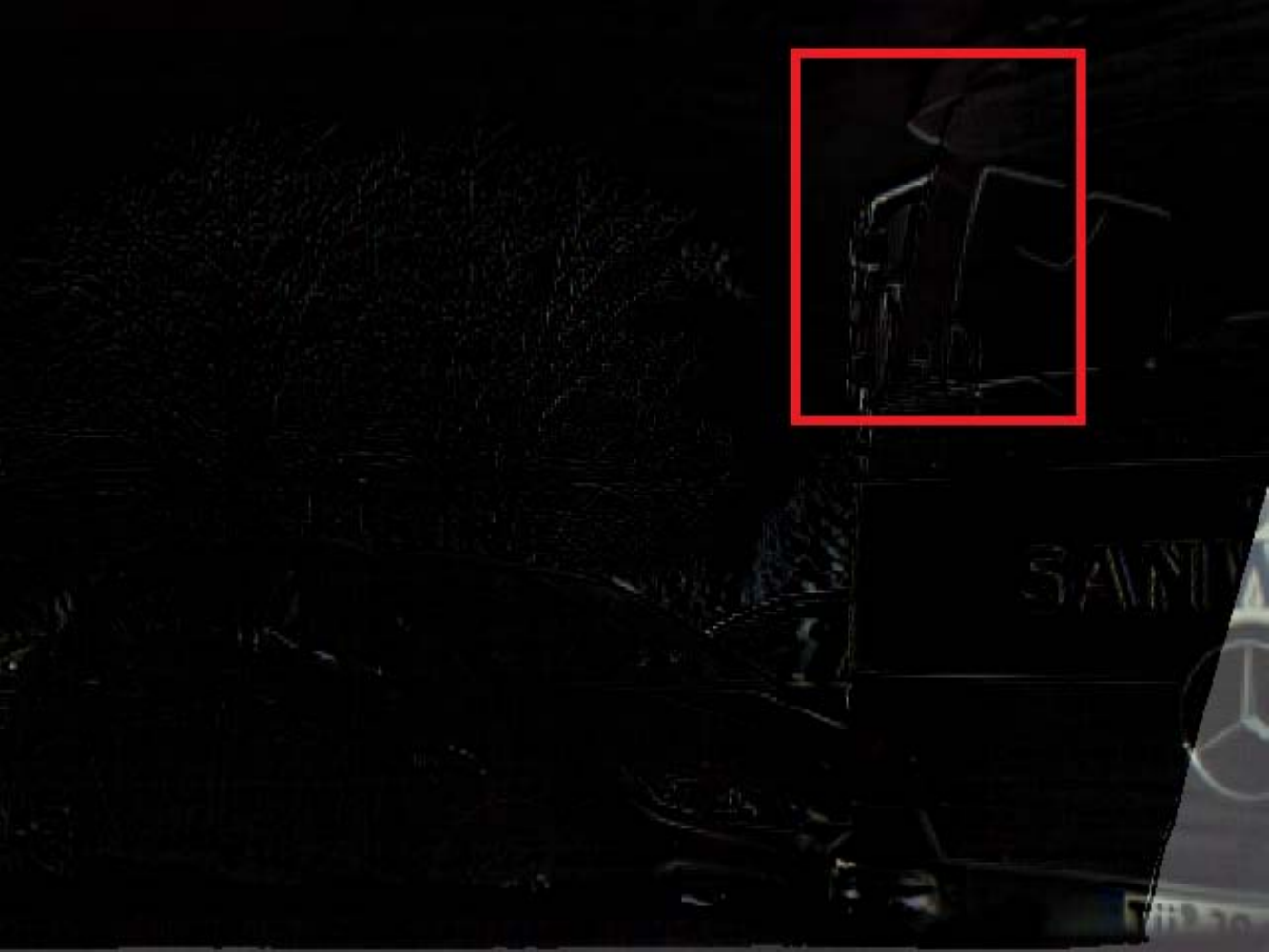}
		&
		\includegraphics[width=0.78\itemwidth]{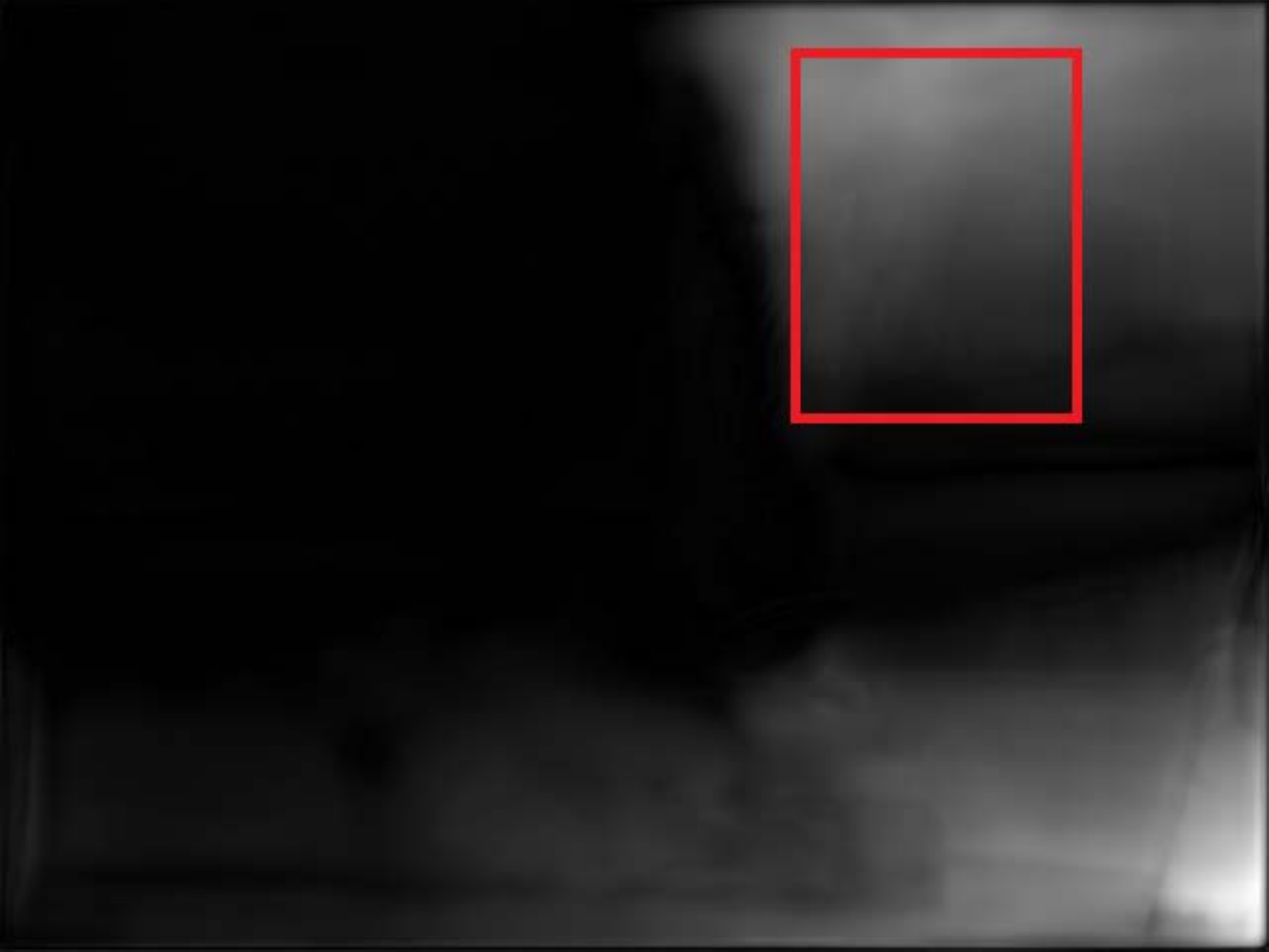}
		\vspace{-0.05cm} \\
		\includegraphics[width=0.78\itemwidth]{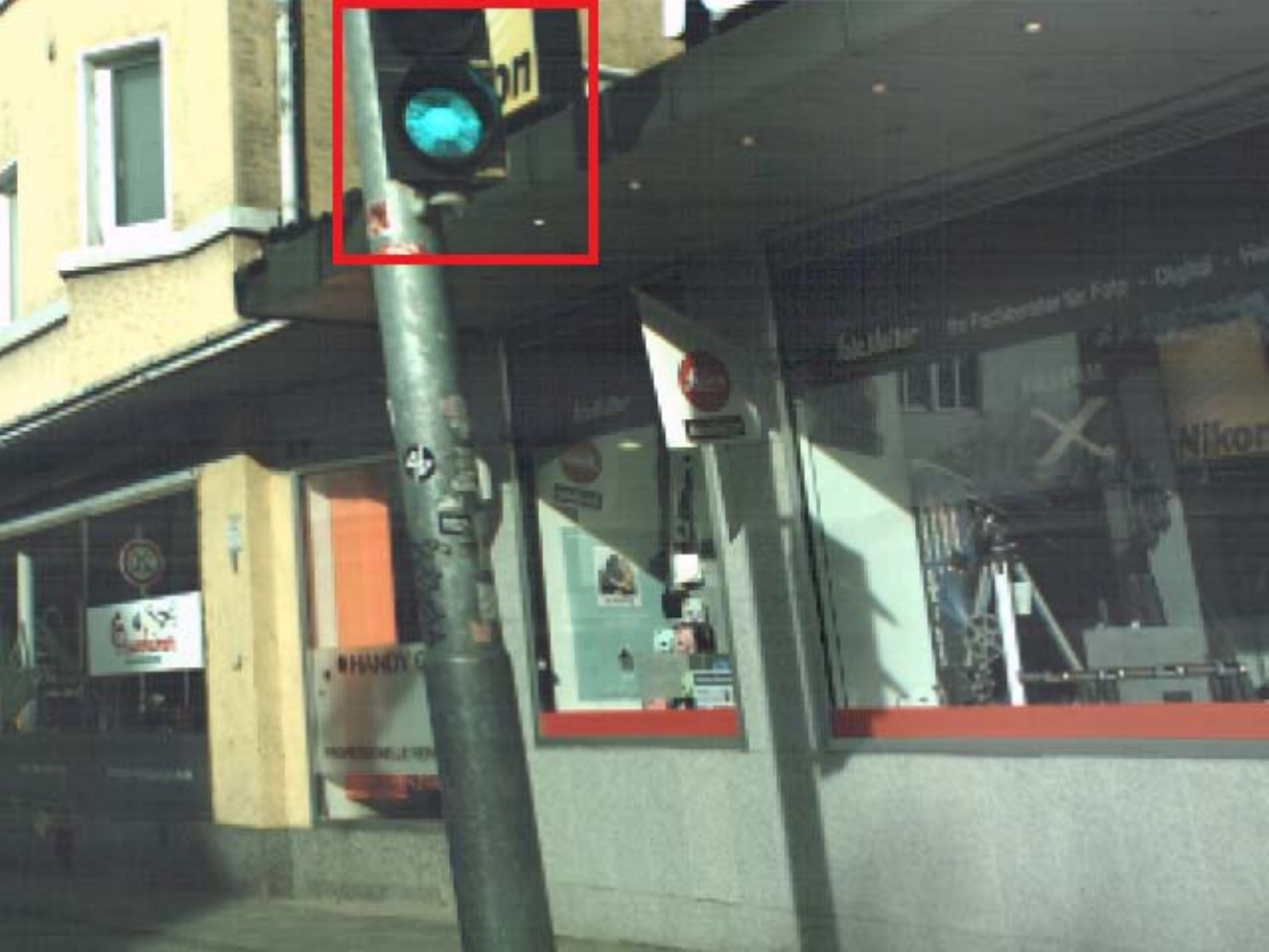}
		&
		\includegraphics[width=0.78\itemwidth]{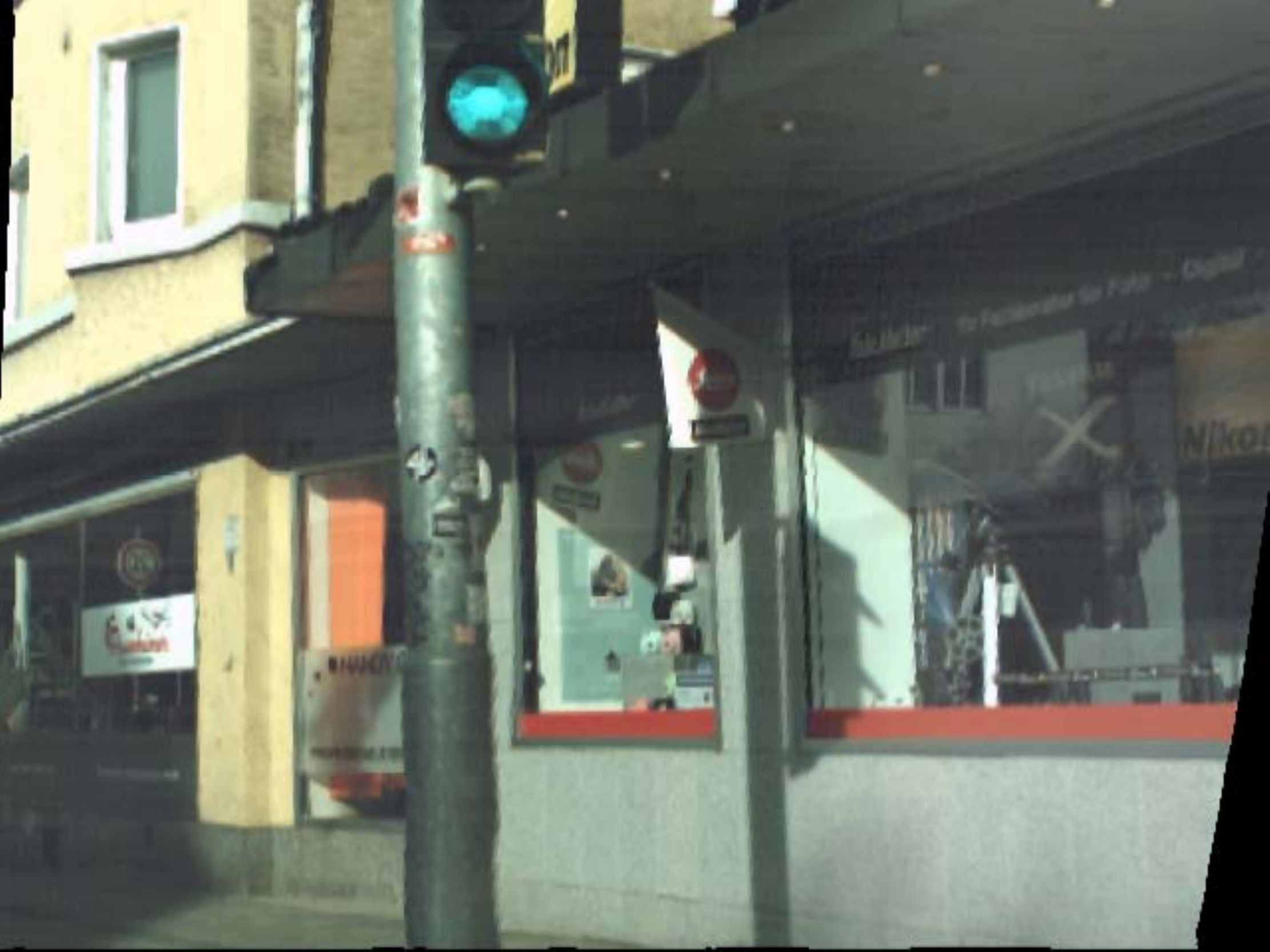}
		&
		\includegraphics[width=0.78\itemwidth]{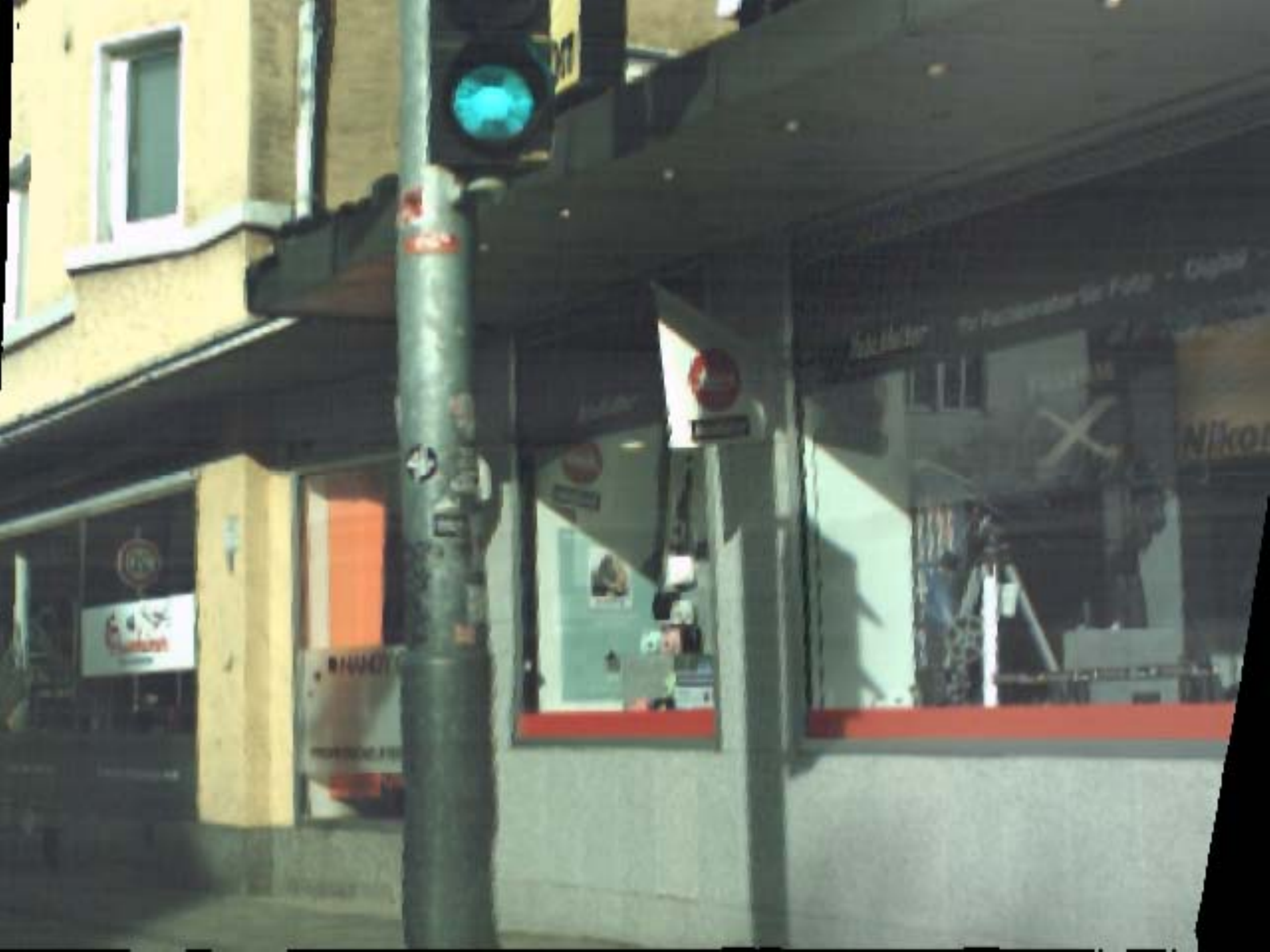}
		&
		\includegraphics[width=0.78\itemwidth]{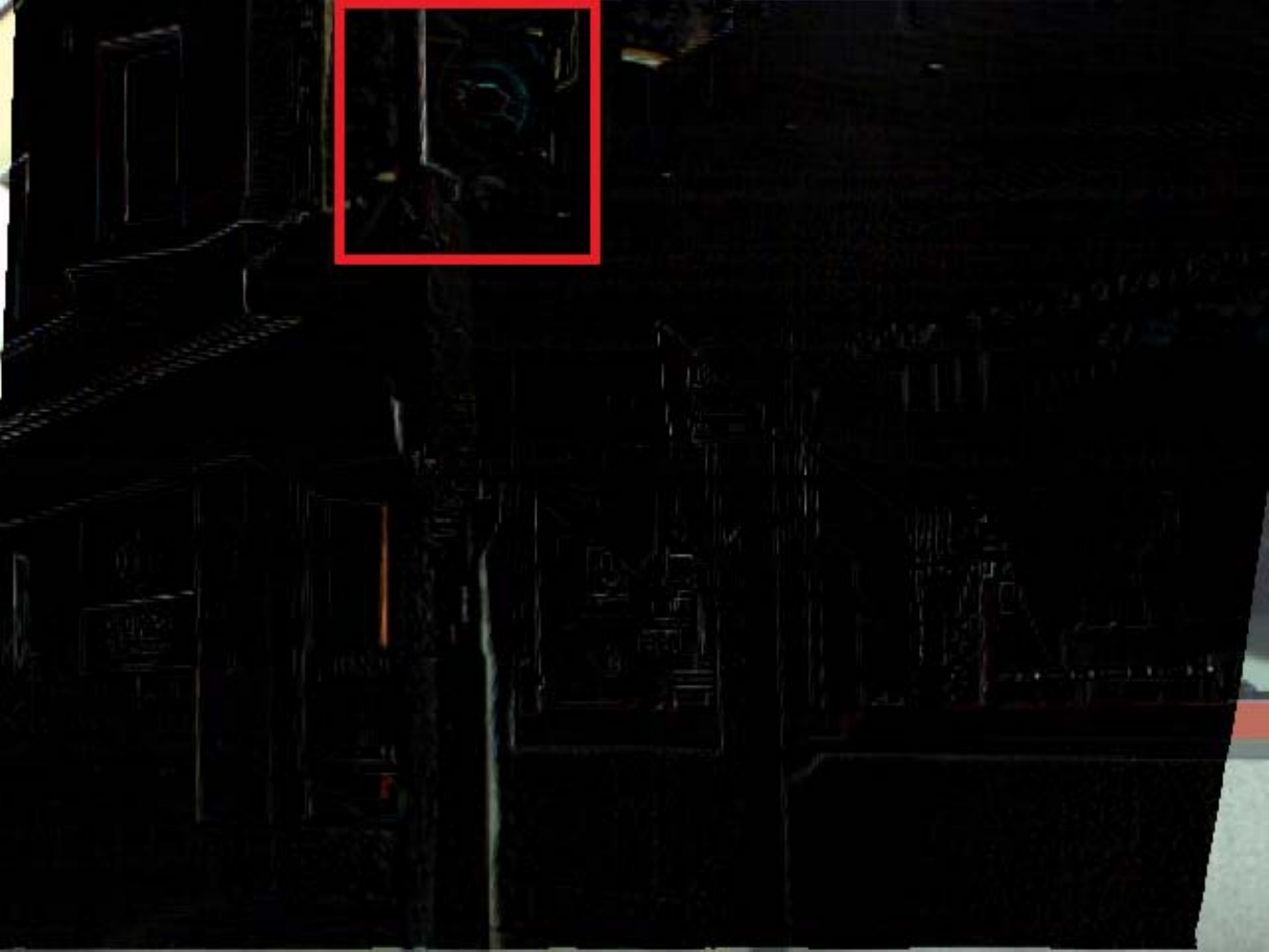}
		&
		\includegraphics[width=0.78\itemwidth]{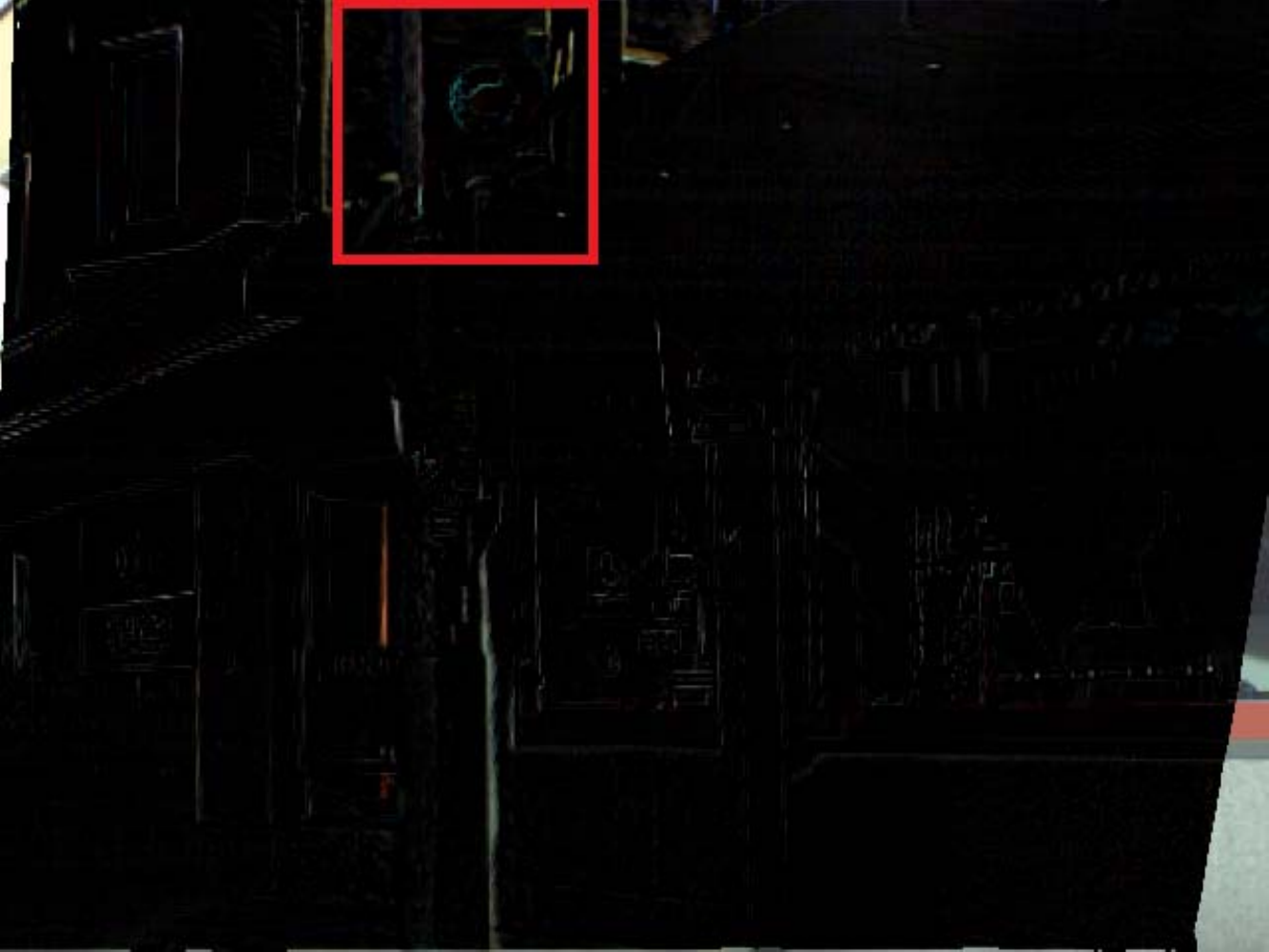}
		&
		\includegraphics[width=0.78\itemwidth]{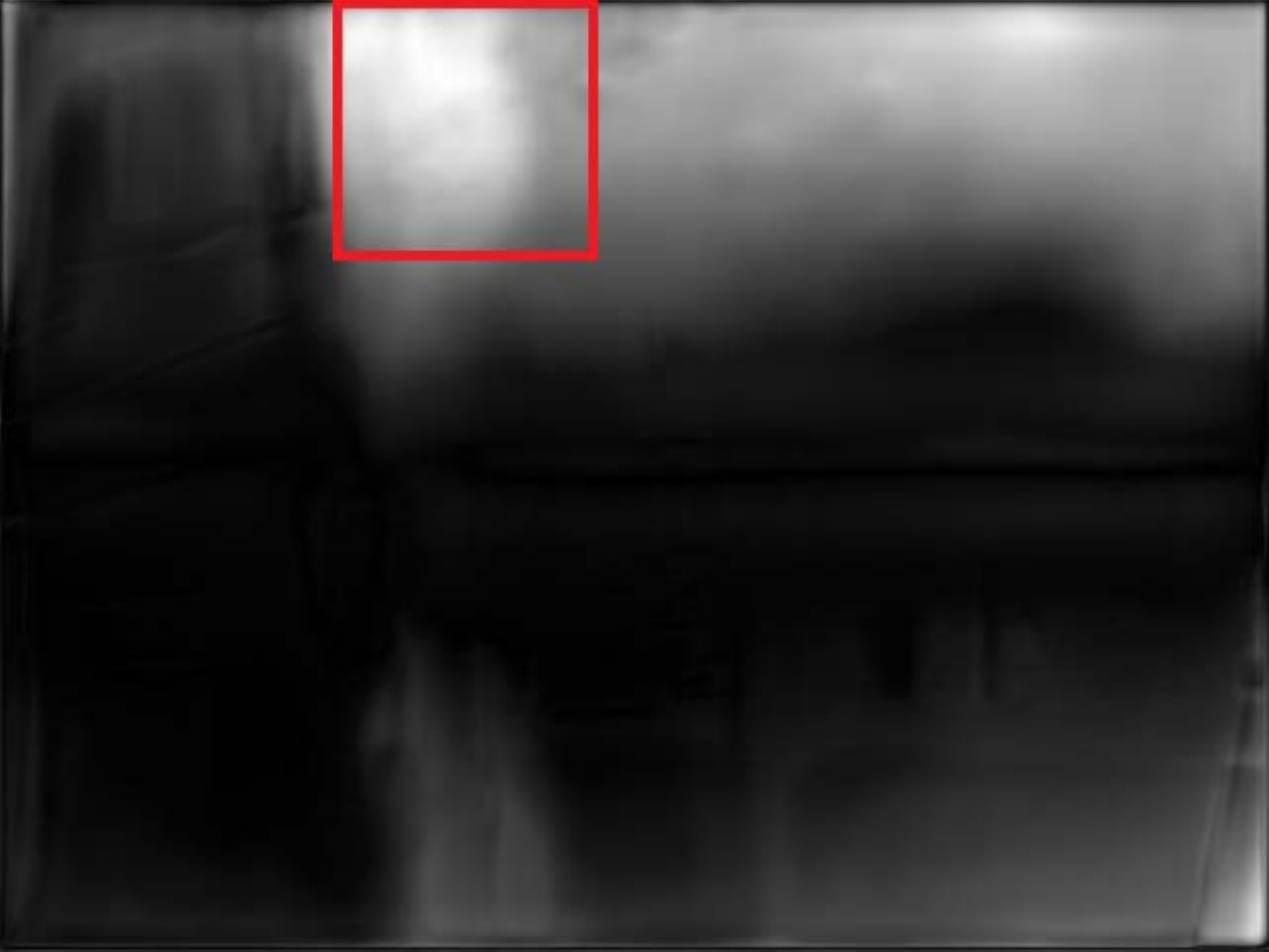}
		\vspace{-0.05cm} \\
		\scriptsize Input RS: ${\mathbf I}_{1}^r$
		&
		\scriptsize Initial GS candidate: ${\mathbf I}_{1\to1}^g$
		&
		\scriptsize Refined GS candidate: $\hat{\mathbf I}_{1\to1}^g$
		&
		\scriptsize $\left\|{\mathbf I}_{1\to1}^g - {\mathbf I}_{1}^{gt}\right\|_{2}$
		&
		\scriptsize $\left\|\hat{\mathbf I}_{1\to1}^g - {\mathbf I}_{1}^{gt}\right\|_{2}$
		&
		\scriptsize $\left\|\Delta{\mathbf U}_{1\to1}\right\|_{2}$
		\\
	\end{tabular}\vspace{-0.341cm}
	\caption{Example results of the effectiveness of our motion enhancement. The second to fifth columns show the initial intermediate GS frame candidates, the refined intermediate GS frame candidates, and their absolute differences with corresponding ground-truth, respectively. The sixth column indicates the mean of the BMF residual map (the brighter a pixel, the bigger the motion enhancement). Our \emph{CVR} effectively enhances ambiguous motion boundaries for more accurate contextual alignment.}
	\label{fig:enhancement_analysis}
\end{figure*}

\subsection{Further analysis on occlusion reasoning} \label{sec:occlusion_analyses}
As can be seen from Fig.~\ref{fig:occlusion_analysis}, the severe occlusion exists in the pool at the lower-left corner of the RS images (\cf blue circle). 
Since RSSR \cite{fan2021rssr} only uses the contents of a single RS image to synthesize the corresponding GS image, a mass of occluded black holes inevitably appear (\cf red circles). 
In contrast, we mitigate this struggle by effectively aggregating contextual information through occlusion inference.
Interestingly, in the examples of time $t=0.5$ shown in Fig.~\ref{fig:occlusion_analysis}, the estimated bilateral occlusion masks can vividly reflect the human intuitive observation discussed in \cite{fan2021sunet}, \ie the first and second rolling shutter images contribute greatly to the lower and upper parts of the latent GS image at time $t=0.5$, respectively.
Meanwhile, the GS image corresponding to time $t=0$ or $t=1$ will be more convinced by the RS image that is closer in time, which reasonably follows the RS imaging mechanism.
In a nutshell, our method can restore high-quality GS images with richer details, enhancing the visual experience. Note that our method can also model temporal abstractions in an end-to-end manner, which allows adaptively generating time-aware occlusion masks to obtain GS images at arbitrary times.

\subsection{Further analysis on motion enhancement} \label{sec:enhancement_analyses}
We further investigate the effectiveness of our motion enhancement layer in Fig.~\ref{fig:enhancement_analysis}, taking the correction to time $t=1$ as an example. As illustrated by the red boxes, the motion enhancement scheme facilitates the quality of the bilateral motion field. As a result, the local image details (\eg object-specific motion boundaries, small errors, \etc) are refined so as to encourage subsequent contextual aggregation. Combined with the proposed contextual consistency constraint, it can promote high-fidelity GS frame synthesis with the assistance of bilateral occlusion masks.

\begin{figure*}[!t]
	\centering
	\includegraphics[width=0.95\textwidth]{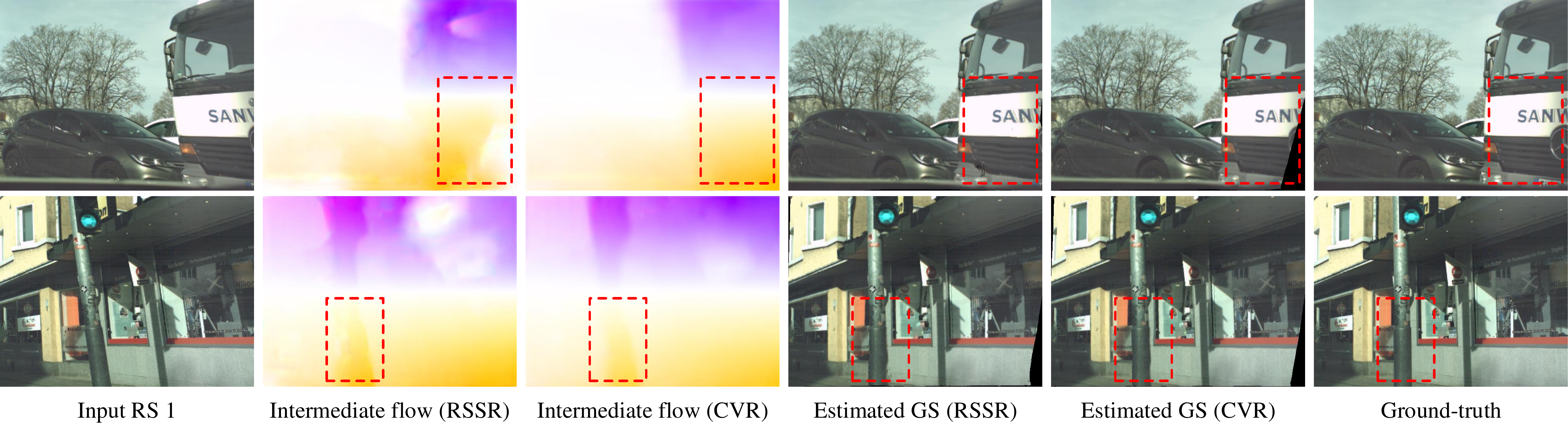}
	\vspace{-3.0mm} 
	\caption{Visual results of the intermediate flow estimation. The estimated GS images in the fourth and fifth columns are obtained by warping the input RS frames according to the intermediate flows in the second and third columns, respectively. Our \emph{CVR} estimates the intermediate flow with clearer motion boundaries than RSSR \cite{fan2021rssr} and thus generates more accurate and sharper GS content.}
	\label{fig:flow_visualization}
	\vspace{-1.2mm}
\end{figure*}

\begin{table*}[!t]
	\footnotesize
	\caption{\small Quantitative comparisons on recovering GS images at time step $t=1$. The numbers in {\color{red}\textbf{red}} and {\color{blue}\underline{blue}} represent the best and second-best performance. In addition to the SOTA quantification performance for GS image recovery at time $t=0.5$, our method also obtains almost consistent best metrics at time $t=1$. Note that not only these, high-quality GS video frames corresponding to any time $t\in[0,1]$ can be accurately estimated by our method.}\label{tab:middle_comparison}
	\vspace{-1.7mm}
	\centering
	\setlength{\tabcolsep}{3mm}{
		\begin{tabular}{lccccccccc}
			\hline
			\multirow{2}{*}{Method} &\multicolumn{3}{c}{PSNR$\uparrow$ (dB)}      &   & \multicolumn{2}{c}{SSIM$\uparrow$} &   & \multicolumn{2}{c}{LPIPS$\downarrow$}     \\ \cline{2-4} \cline{6-7} \cline{9-10}
			& CRM            & CR             & FR            & & CR            & FR   & & CR            & FR         \\ \hline
			DeepUnrollNet \cite{liu2020deep}        & 27.86    & {27.54}     & \color{red}\textbf{27.02}   & & 0.829  & {0.828}    & & 0.0555    & \color{red}\textbf{0.0791}    \\
			RSCD \cite{zhong2021rscd} & -    & -     & 24.84   & & -     & 0.778  & & -     & 0.1070        \\
			RSSR \cite{fan2021rssr} & \color{blue}\underline{29.36} & 26.57 & 24.89 & & {0.900} & 0.824 & & {0.0553}     & 0.1109\\ \hline
			\emph{CVR*} (Ours) & {28.28} & \color{blue}\underline{28.19} & {26.58} & & \color{blue}\underline{0.912} & \color{blue}\underline{0.833} & & \color{blue}\underline{0.0444}     & 0.1014 \\ 
			\emph{CVR} (Ours) & \color{red}\textbf{29.41} & \color{red}\textbf{29.19} & \color{blue}\underline{26.67} & & \color{red}\textbf{0.915} & \color{red}\textbf{0.838} & & \color{red}\textbf{0.0403}     & \color{blue}\underline{0.1011}\\ \hline
		\end{tabular}
		\vspace{-0.5mm}
		\normalsize
		\begin{tablenotes}
			\raggedleft
			\item{
				\small{*: \emph{applying our proposed approximated bilateral motion field (ABMF) model.}}
			}
		\end{tablenotes}
	}
	\vspace{-2.0mm}
\end{table*}

\section{Additional Experimental Results} \label{sec:additional_results}
In this section, we present more qualitative and quantitative experimental results on effect removal, intermediate flow, and generalization, \etc.
Furthermore, a video demo is included to show the dynamic results of reconstructing slow-motion GS video from two consecutive RS frames.

\subsection{RS effect removal} \label{sec:RS_removal}
\begin{figure*}[!t]
	\centering
	\includegraphics[width=0.8\textwidth]{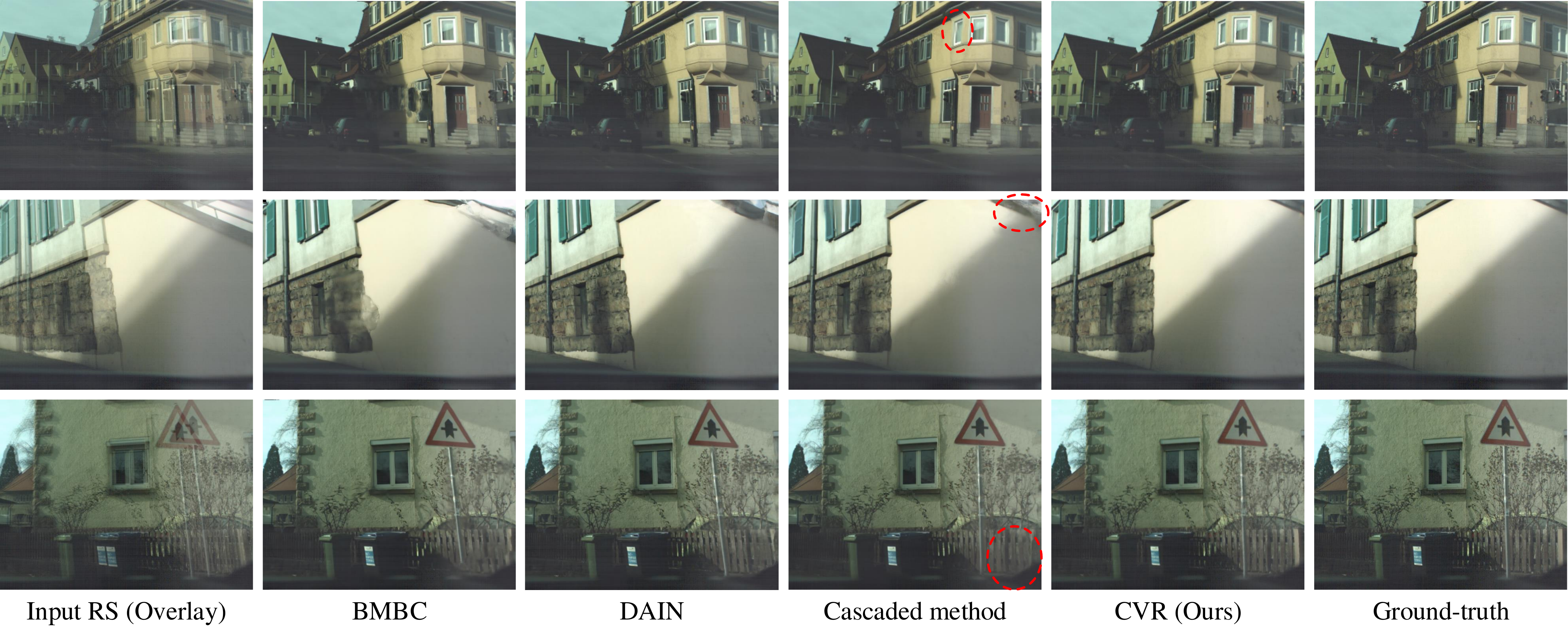}
	\vspace{-2.0mm} 
	\caption{Visual examples against off-the-shelf VFI approaches (\ie. BMBC~\cite{park2020bmbc}, DAIN~\cite{bao2019depth}, and Cascaded method). Although the cascaded method can compensate the drawback of the VFI method that cannot remove RS artifacts, it is also prone to local errors due to error accumulation, as shown by the red circles.
		\label{fig:Vs_vfi_cascate}}
	\vspace{-4.0mm}
\end{figure*}

\begin{figure*}[!t]
	\centering
	\includegraphics[width=0.754\textwidth]{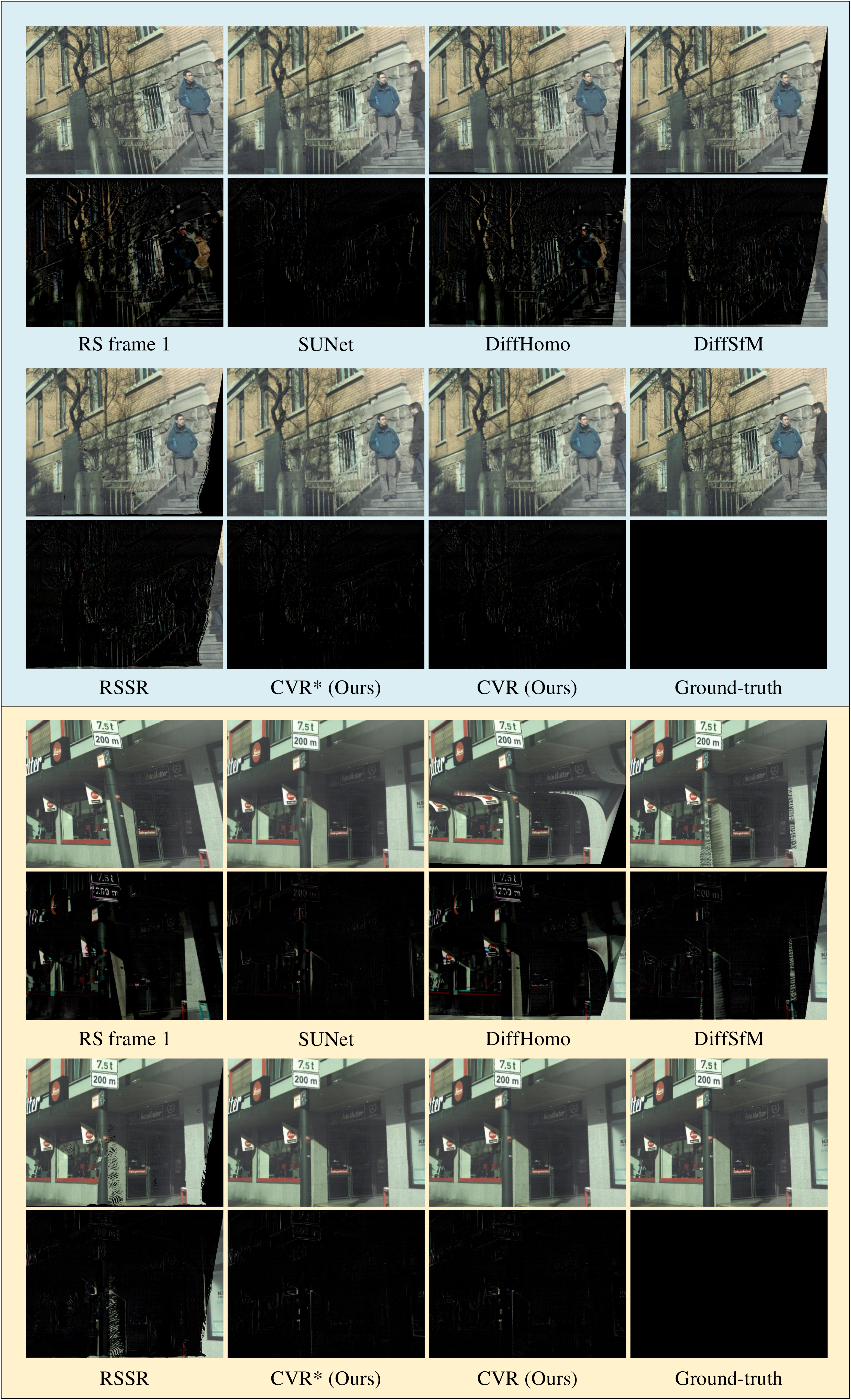}
	\vspace{-1.5mm} 
	\caption{Rolling shutter effect removal examples against competing approaches (\ie. SUNet \cite{fan2021sunet}, DiffHomo \cite{zhuang2020homography}, DiffSfM \cite{zhuang2017rolling}, and RSSR \cite{fan2021rssr}). Even columns: Absolute difference between the corrected global shutter image and the corresponding ground-truth. 
		\label{fig:Vs_other_methods}}
	\vspace{-2.0mm}
\end{figure*}

First, we visualize the intermediate flow in Fig.~\ref{fig:flow_visualization} and compare it with the SOTA RS-based video reconstruction method RSSR \cite{fan2021rssr}. In contrast to RSSR, our pipeline generates intermediate flows with clearer motion boundaries for more accurate frame interpolation due to motion interpretation and occlusion reasoning.
Then, we report more RS effect removal results in Fig.~\ref{fig:Vs_vfi_cascate} and Fig.~\ref{fig:Vs_other_methods} by comparing with the off-the-shelf video frame interpolation (VFI) and RS correction algorithms.
Finally, in Table~\ref{tab:middle_comparison}, we give quantitative comparison results of GS image recovery at time $t=1$. In addition to the superior RS effect removal performance at time $t = 0.5$, our pipeline also significantly surpasses RSSR at time $t=1$.
In summary, these experimental results consistently demonstrate that the proposed method has superior RS effect removal capabilities, successfully restoring higher fidelity global shutter video frames with fewer artifacts and richer details.

\subsection{Generalization on other real data} \label{sec:generalization}
\begin{figure*}[!t]
	\centering
	\includegraphics[width=0.973\textwidth]{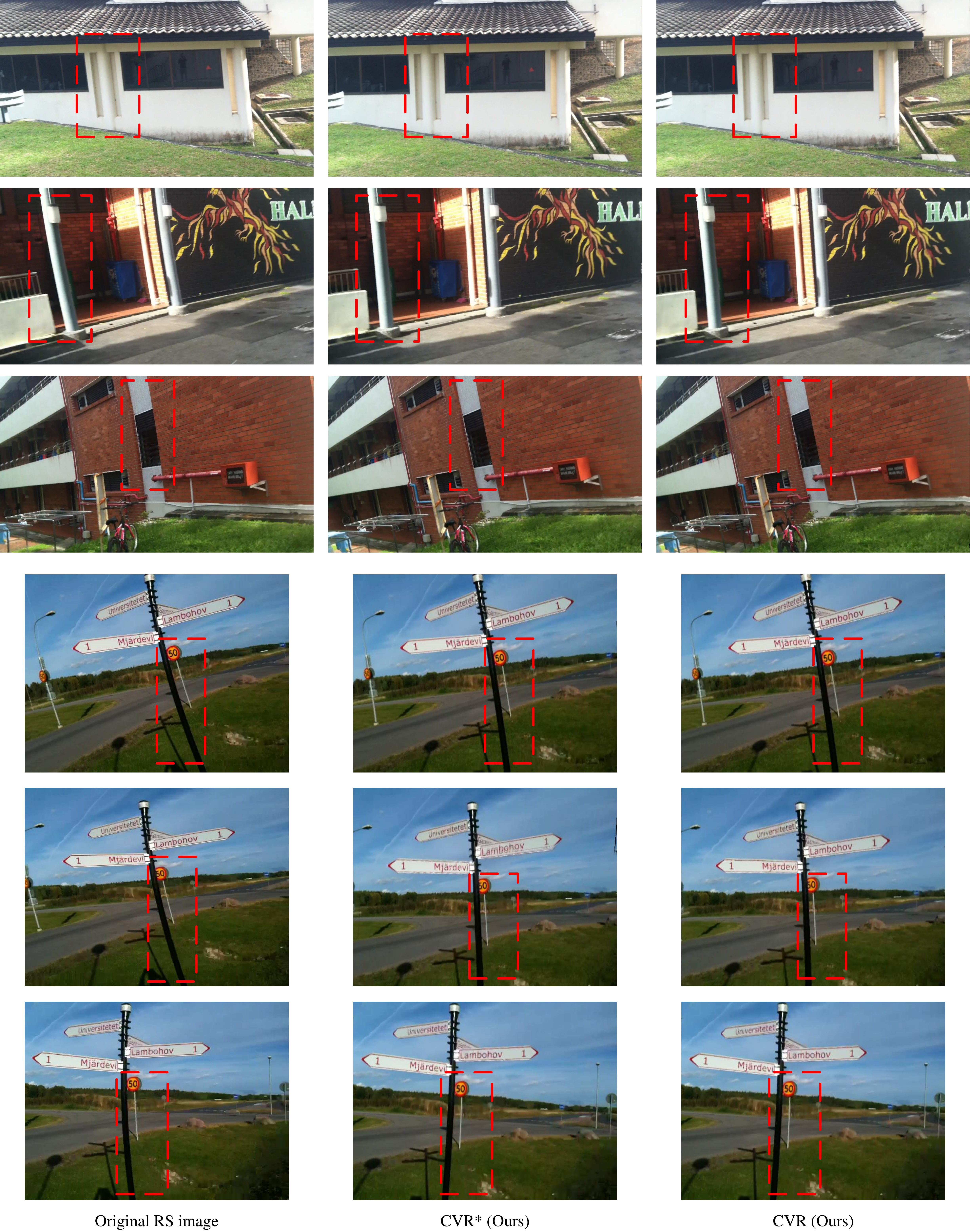}
	\vspace{-1.5mm} 
	\caption{Generalization results on real rolling shutter data with noticeable rolling shutter artifacts. The data in the first three rows are from \cite{zhuang2017rolling}, and the last three rows are from \cite{forssen2010rectifying}. Consistent and high-quality correction results are obtained by our \emph{CVR} and \emph{CVR*}.
		\label{fig:real_exp}}
	\vspace{-2.0mm}
\end{figure*}

To evaluate the generalization performance of the proposed method on real rolling shutter images, we utilize the data provided by \cite{zhuang2017rolling} and \cite{forssen2010rectifying}, in which the hand-held cameras move quickly in the real world to capture real RS image sequences. As shown in Fig.~\ref{fig:real_exp}, our \emph{CVR} and \emph{CVR*} can effectively and robustly remove the RS effect to obtain consistent distortion-free images, which validates the excellent generalization performance of our method in practice.

\subsection{GS video reconstruction demo} \label{sec:video_demo}
We attach a supplementary video {\color{magenta}$\mathsf{demo\_video.mp4}$} to dynamically demonstrate the GS video reconstruction results.
In the video, we show the $10 \times$ temporal upsampling results, \ie evenly interpolating 11 intermediate GS frames corresponding to time steps 0, 0.1, 0.2, ..., 0.9, 1.
In essence, our method is capable of generating GS videos with arbitrary frame rates.
Note that except for times 0, 0.5, and 1, our method has not been fed with GS images of other time instances during training.
More qualitative results on RS correction datasets \cite{liu2020deep} and real RS data \cite{zhuang2017rolling,forssen2010rectifying} can be seen in the supplementary video.
With these examples, we can conclude that our method not only achieves the state-of-the-art RS effect removal performance that is significantly better than competing methods, but also has the superior ability to recover high-quality and high-framerate GS videos.

\section{Details of Loss Function} \label{sec:details_loss}
Assuming that $T$ GS images at time instances $\{t_i\}_1^T, t_i \in [0,1]$ are to be recovered to supervise the training of our model and that ${\mathbf I}_{t_i}^{gt}$ is the corresponding ground-truth (GT) GS image, our loss function $\mathcal{L}$ is a linear combination of the reconstruction loss $\mathcal{L}_r$, perceptual loss $\mathcal{L}_p$, contextual consistency loss $\mathcal{L}_c$, and total variation loss $\mathcal{L}_{tv}$, \ie
\begin{equation}\label{eq:16}
\mathcal{L} = \lambda_r\mathcal{L}_r + \mathcal{L}_p + \lambda_c\mathcal{L}_c + \lambda_{tv}\mathcal{L}_{tv},
\end{equation}
where $\lambda_r$, $\lambda_c$ and $\lambda_{tv}$ are hyper-parameters.
The pixel intensities of images are normalized.

The \emph{reconstruction loss $\mathcal{L}_r$} models the pixel-wise $\mathcal{L}_1$ loss betweeen the final GS frame prediction and the corresponding ground-truth, given by
\begin{equation}\label{eq:17}
\mathcal{L}_{r} = \frac{1}{T} \sum_{i=1}^T\left\|\hat{\mathbf I}_{t_i}^g-{\mathbf I}_{t_i}^{gt}\right\|_{1}.
\end{equation}

The \emph{perceptual loss $\mathcal{L}_p$} contributes to produce fine details and improves the perceptual quality of the final intermediate GS frame \cite{johnson2016perceptual} by
\begin{equation}\label{eq:18}
\mathcal{L}_{p} = \frac{1}{T} \sum_{i=1}^T\left\| \phi \left( \hat{\mathbf I}_{t_i}^g \right) - \phi \left( {\mathbf I}_{t_i}^{gt} \right) \right\|_{1},
\end{equation}
where $\phi$ is the conv4\_3 features of the pre-trained VGG16 network \cite{simonyan2014very}, as widely used in \cite{liu2020deep,jiang2018super,niklaus2018context}.

The \emph{contextual consistency loss $\mathcal{L}_c$} encourages the alignment of the refined intermediate GS frame candidates and their ground-truth frame at time $t_i$. This can also facilitate the final enhanced BMF to reason about the underlying occlusions and the object-specific motion boundaries, which are crucial for the final GS frame synthesis. Specifically, we define $\mathcal{L}_c$ as:
\begin{equation}\label{eq:19}
\mathcal{L}_{c} = \frac{1}{2T} \sum_{i=1}^T\left(\left\|\hat{\mathbf I}_{0 \to t_i}^g-{\mathbf I}_{t_i}^{gt}\right\|_{1} + \left\|\hat{\mathbf I}_{1 \to t_i}^g-{\mathbf I}_{t_i}^{gt}\right\|_{1} \right).
\end{equation}

The \emph{total variation loss $\mathcal{L}_{tv}$} enforces piecewise smoothness in the final enhanced BMF \cite{liu2017video,fan2021sunet}, \ie
\begin{equation}\label{eq:20}
\mathcal{L}_{tv} = \frac{1}{2T} \sum_{i=1}^T\left(\left\|\nabla \hat{\mathbf U}_{0 \to t_i}\right\|_{2} + \left\|\nabla \hat{\mathbf U}_{1 \to t_i}\right\|_{2} \right).
\end{equation}

\begin{table}[!t]
	\footnotesize
	\caption{Ablation results for \emph{CVR*} architecture on $\mathcal{M}_A$ and $\mathcal{G}$.} \label{tab:ablation_cvr*}
	\vspace{-2.5mm}
	\centering
	\begin{tabular}{lcccccc}
		\hline
		\multirow{2}{*}{Settings} & \multicolumn{3}{c}{PSNR$\uparrow$ (dB)}                      &   & \multicolumn{2}{c}{SSIM$\uparrow$}      \\ \cline{2-4} \cline{6-7}
		& CRM       & CR         & FR        & & CR       & FR     \\ \hline
		RAFT-based              & 30.40     & 29.91      & 27.67     & & 0.914     & 0.835     \\ 
		Freeze $\mathcal{M}_A$    & 31.69     & 31.53      & 28.51     & & \textbf{0.927}     & 0.843     \\ \hline
		${\mathbf T}\cdot\Delta{\mathbf U}$ & 31.15     & 30.95      & 28.01     & & 0.916     & 0.831  \\
		w/o $\Delta{\mathbf U}$ & 31.61     & 31.41      & 27.99     & & 0.925     & 0.831     \\ 
		w/o ${\mathbf O}$       & 30.96     & 30.80      & 23.89     & & 0.913     & 0.804     \\ \hline
		full model & \textbf{31.82} & \textbf{31.60} & \textbf{28.62} & & \textbf{0.927} & \textbf{0.845} \\ \hline
	\end{tabular}
	\vspace{-3.0mm}
\end{table}

\vspace{-0.5mm}
\section{Ablations of the Proposed CVR*}
Additionally, we report the impact of different network architecture designs on our \emph{CVR*} in Table~\ref{tab:ablation_cvr*} by referring to the GS images at time $t=0.5$.
Ablation results that are almost consistent with \emph{CVR} in the main paper can be obtained, which fully demonstrates the validity of the network architecture we used.

\vspace{-1.0mm}
\section{Failure Cases} \label{sec:failure cases}
\begin{figure}[htbp]
	\centering
	\includegraphics[width=0.38\textwidth]{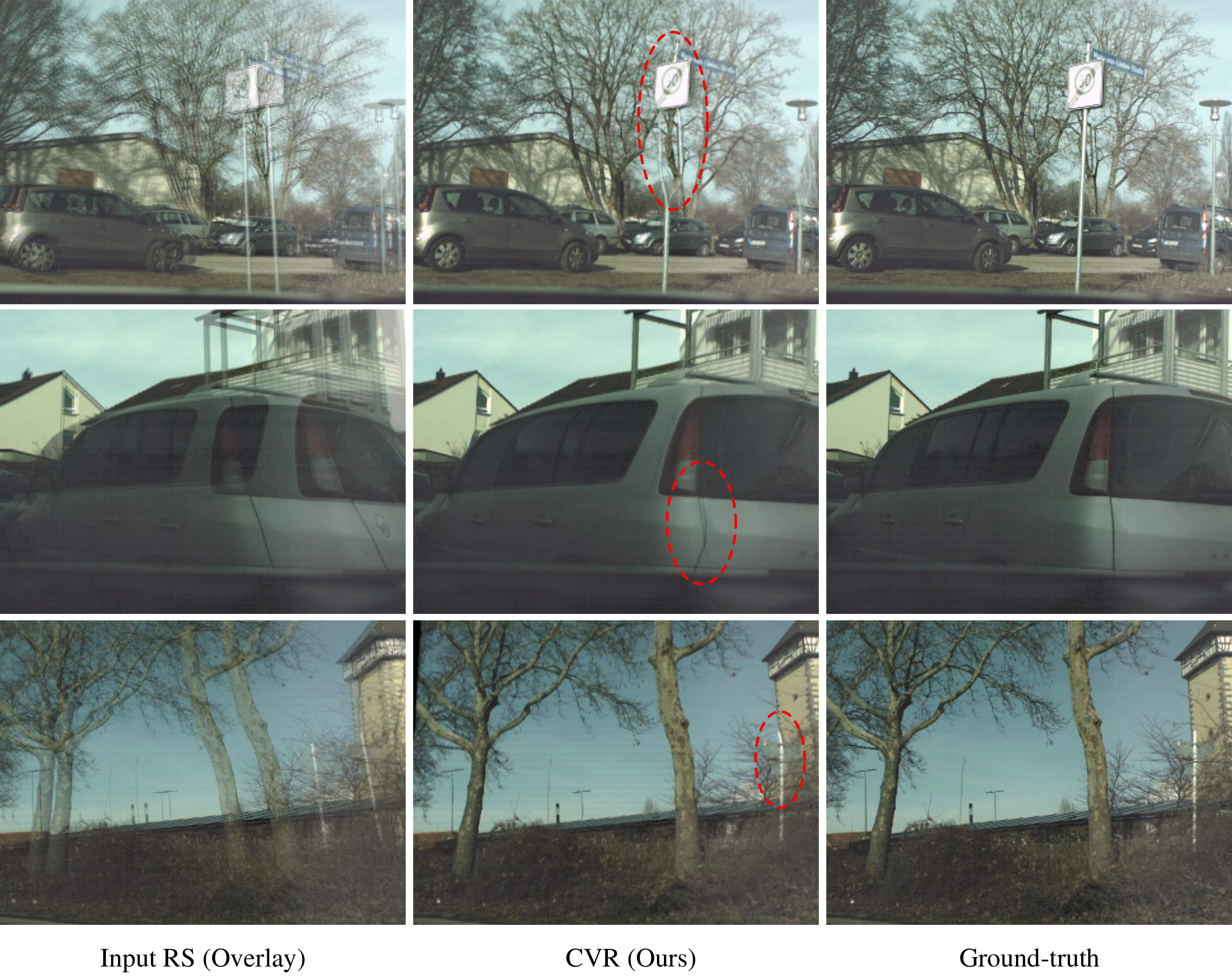}
	\vspace{-1.0mm} 
	\caption{Failure cases in challenging areas. The white pillars and car tails lack texture and are thus prone to aliasing artifacts.
		\label{fig:failure_cases}}
\end{figure}

\vspace{-1.0mm}
We have discussed that our pipeline may have blending and ghosting artifacts in image regions such as low/weak/repetitive textures. We reckon this is because our method exploits image-based warping, and thus potential gross errors of the estimated BMF in these challenging regions can easily lead to contextual misalignment. In fact, this is a common challenge for the current RS correction method based on image warping, \eg \cite{zhuang2017rolling,zhuang2020homography,zhuang2019learning,fan2021rssr}. 
We show visual results of the failure cases in Fig.~\ref{fig:failure_cases}. 
Similar to the training process of VFI methods 
\cite{paliwal2020deep,xu2019quadratic,jiang2018super}, 
it will likely be helpful to use more GT GS images at different time steps to supervise the training of our network.
In the future, we also plan to improve the BMF estimation or design feature-based aggregation schemes to ameliorate this weakness.

\end{appendices}

\end{document}